\documentclass[letterpaper]{article} 
\RequirePackage{etex} 
\usepackage{hyperref} 
\usepackage{xcolor}
\hypersetup{
  colorlinks, 
  linkcolor={red!60!black},
  citecolor={blue!80!black},
  urlcolor={blue!80!black}
}

\usepackage{aaai23}
\usepackage{times}  
\usepackage{helvet}  
\usepackage{courier}  
\usepackage{graphicx} 
\usepackage{natbib}  
\usepackage{caption} 
\usepackage{mydefs}
\usepackage{newfloat}
\usepackage{listings}
\DeclareCaptionStyle{ruled}{labelfont=normalfont,labelsep=colon,strut=off} 
\floatstyle{ruled}
\newfloat{listing}{tb}{lst}{}
\floatname{listing}{Listing}
\title{Versatile Single-Loop Method for Gradient Estimator:\\ First and Second Order Optimality, and its Application to Federated Learning}
\author {
    Kazusato Oko\textsuperscript{\rm 1,2},
    Shunta Akiyama\textsuperscript{\rm 1},
    Tomoya Murata \textsuperscript{\rm 1,3},
    Taiji Suzuki \textsuperscript{\rm 1,2}
}
\affiliations {
    \textsuperscript{\rm 1} Graduate School of Information Science and Technology, The University of Tokyo, Tokyo, Japan\\
    \textsuperscript{\rm 2} RIKEN Center for Advanced Intelligence Project, Tokyo, Japan\\
    \textsuperscript{\rm 3} NTT DATA Mathematical Systems Inc., Tokyo, Japan\\
    oko-kazusato@g.ecc.u-tokyo.ac.jp, shunta\_akiyama@mist.u-tokyo.ac.jp,
    murata@msi.co.jp,
    taiji@mist.u-tokyo.ac.jp
}

\usepackage{bibentry}

\begin{document}

\maketitle

\begin{abstract}
While variance reduction methods have shown great success in solving large scale optimization problems, many of them suffer from accumulated errors and, therefore, should periodically require the full gradient computation.
In this paper, we present a single-loop algorithm named SLEDGE (Single-Loop mEthoD for Gradient Estimator) for finite-sum nonconvex optimization, which does not require periodic refresh of the gradient estimator but achieves nearly optimal gradient complexity.
Unlike existing methods, SLEDGE has the advantage of versatility; (i) second-order optimality, (ii) exponential convergence in the PL region, and (iii) smaller complexity under less heterogeneity of data.

We build an efficient federated learning algorithm by exploiting these favorable properties.
We show the first and second-order optimality of the output and also provide analysis under PL conditions.
When the local budget is sufficiently large and clients are less (Hessian-)~heterogeneous, the algorithm requires fewer communication rounds then existing methods such as FedAvg, SCAFFOLD, and Mime.
The superiority of our method is verified in numerical experiments.
\end{abstract}

\section{Introduction}
We solve a stochastic minimization problem of the following form without periodic full gradient computation:
\begin{align}
\label{eq:Intro-FiniteSum}
    \min_{x\in \R^d} \left\{f(x):=\frac1n\sum_{i=1}^n f_i(x)\right\},
\end{align}
where each $f_i$ is a smooth function and can be \textit{nonconvex}.
Nowadays, the problem \eqref{eq:Intro-FiniteSum} appears in a wide range of machine learning optimization with very large $n$ \citep{bottou2018optimization}.
We aim to efficiently find a solution $x$ that is an $\eps$-first-order stationary point (i.e., $\|\nabla f(x)\|\leq \eps$) and furthermore an $(\eps,\delta)$-second-order stationary point (SOSP; i.e., $\|\nabla f(x)\|\leq \eps$ and $\lambda_{\rm min}(\nabla^2 f(x))\geq -\delta$). 

\subsection{Variance reduction}
Variance reduction is a technique in minibatch sampling to construct a gradient estimator with a smaller variance than vanilla SGD by utilizing gradients at previously obtained anchor points 
\citep{roux2012stochastic,johnson2013accelerating,defazio2014saga}.
One of the difficulties in obtaining an appropriate gradient estimator, especially in nonconvex settings, is that recursive update of a gradient estimator with minibatch gradients easily accumulates the error and eventually buries the correct descent directions.
To address this issue, there have been two major approaches.
The first approach is to explicitly store previously calculated gradients as in SAGA \citep{defazio2014saga}. 
However, SAGA's convergence rate is $O(\frac{n^\frac23}{\eps^2})$ \citep{reddi2016fast}, which is still sub-optimal from the lower bound 
\cite{fang2018spider,li2021page}.
The second one is to use double-loop algorithms that periodically compute the full gradient or a gradient with a large minibatch to refresh a gradient estimator.
These algorithms include SARAH, SPIDER, and NestedSVRG
\citep{nguyen2017stochastic,fang2018spider,zhou2020stochastic}, which have the optimal rate of $O(\frac{\sqrt{n}}{\eps^2})$.
On the other hand, this approach has an issue that the step of gradient-refreshing slows down
practical computational speed 
and becomes a bottleneck in the application to federated learning since this leads to periodic synchronization of the whole client.


\begin{table*}[t]\centering
\begin{threeparttable}
  \caption{Stochastic gradient complexity of optimization algorithms for a nonconvex finite-sum problem \eqref{eq:Intro-FiniteSum}.}
  \label{table:Intro-FiniteSum}
  \begin{tabularx}{\linewidth}{p{48mm}p{15mm}p{41mm}p{25.7mm}XX}  \bhline{0.8pt}
  \multirow{2}{*}{Algorithms} & \multicolumn{3}{c}{Stochastic gradient complexity}& Periodic \\\cline{2-4}
   & Nonconvex & SOSP & PL condition & full gradient  \\  \bhline{0.8pt}
  (Noisy) SGD {\small\citep{ghadimi2013stochastic,ge2015escaping,karimi2016linear}} & $\frac{\Delta\sigma_c^2}{\eps^4}$ & ${\rm poly}(\eps^{-1},\delta^{-1},d,\sigma,\Delta)$ & $\frac{\sigma^2}{\mu^2\eps}\log \eps^{-1}$ & Every iteration\\\hline
  SPIDER$\text{-SFO}^{+}$ {\small\citep{fang2018spider}}& $n+\frac{\Delta\sqrt{n}}{\eps^2}$
  & $n+\Delta(\frac{\sqrt{n}}{\eps^2}+\frac{1}{\eps\delta^3}+\frac{1}{\delta^5})$ & None & Required \\\hline
  SARAH {\small\citep{nguyen2017stochastic}}  and its variants {\small \citep{li2019ssrgd,nguyen2021inexact}}
  & $n+\frac{\Delta\sqrt{n}}{\eps^2}$ & $n+\Delta(\frac{\sqrt{n}}{\eps^2}+\frac{\sqrt{n}}{\delta^4}+\frac{n}{\delta^3})$ & $n+\frac{\sigma^2}{\eps^2}\log \eps^{-1}$ & Required \\\hline
  ZeroSARAH & $\frac{(\Delta+\sigma_c^2)\sqrt{n}}{\eps^2}$ & None & None &
  Never required \\\cline{2-5}
  {\small\citep{li2021zerosarah}} & $n+\frac{\Delta\sqrt{n}}{\eps^2}$ & None & None &
  Only at $x^0$ \\\hline
  PAGE {\small\citep{li2021page}} & $n+\frac{\Delta\sqrt{n}}{\eps^2}$ & None & $(n+\frac{L\sqrt{n}}{\mu})\log \eps^{-1}$ &
  Required  \\\hline
  \rowcolor[rgb]{0.85, 1.0, 1.0} SLEDGE (Option I) {\small(ours)} & $\frac{(\Delta+\sigma_c^2)\sqrt{n}}{\eps^2}$ & $(\Delta+\sigma_c^2)(\sqrt{n}+\frac{\zeta^2}{\delta^2})(\frac{1}{\eps^2}+\frac{1}{\delta^2})$ & $(n+\frac{\zeta\sqrt{n}}{\mu})\log \eps^{-1}$ & Never required \\ \hline
 \rowcolor[rgb]{0.85, 1.0, 1.0} SLEDGE (Option II) {\small(ours)} &  $n+\frac{\zeta\Delta\sqrt{n}}{\eps^2}$ & $n+\Delta(\frac{\sqrt{n}}{\eps^2}+\frac{\sqrt{n}}{\delta^4}+\frac{\zeta^2}{\eps^2\delta^2}+\frac{\zeta^2}{\delta^6})$ & $(n+\frac{\zeta\sqrt{n}}{\mu})\log \eps^{-1}$ & Only at $x^0$ \\ 
  \bhline{0.8pt}
  \end{tabularx}
\footnotesize
{\bf Note:} 
Here $\Delta = f(x^0)-\inf f(x)$, $\sigma_c$ is the variance between $f_i(x)$, $\mu$ is the parameter for PL condition, and $\zeta$ is the Hessian-heterogeneity.

In nonconvex and SOSP problems, polylogarithmic terms are omitted.
Since $\zeta \leq 2L$, SLEDGE with Option I has at most the same complexity to ZeroSARAH, and SLEDGE with Option II does to SPIDER, SARAH, PAGE, and the lower bound, up to log factors.

In PL, polylogarithmic dependency on other than $\eps^{-1}$ and doubly-logarithmic terms are omitted, thus SLEDGE has exponential convergence.
\end{threeparttable}
\vspace{-2mm}
\end{table*}

Recent studies have attempted to develop methods that solve this trade-off, or namely that do not require periodic computation of gradients with a large minibatch size to achieve near-optimal rates \citep{cutkosky2019momentum,liu2020optimal,kovalev2020don,li2021zerosarah,nguyen2021inexact,beznosikov2021random,tran2022hybrid}.
Among them, \citet{li2021zerosarah} introduced ZeroSARAH as a single-loop algorithm with optimal gradient complexity for nonconvex optimization.
Here, we say an algorithm is single-loop when it does not require periodic full or large minibatch gradients.

However, these recent single-loop methods have an issue in their versatility.
First, while it is usual to extend an optimization algorithm to ensure second-order optimality \citep{ge2015escaping,jin2017escape,vlatakis2019efficiently}, and variance reduction methods also have been applied to this \citep{allen2018neon2,fang2018spider,li2019ssrgd},
no single-loop algorithm cannot find SOSPs.
Since first-order stationary points can include a local maximum or a saddle point in nonconvex optimization, escaping them and finding SOSPs are necessary to guarantee the quality of the solution.


Next, most of the existing single-loop methods have focused on removing full gradient computation in some specific setting. 
Thus none of them achieve both optimal complexity in nonconvex settings and exponential convergence in strongly-convex settings. 
Here we are interested in the Polyak-{\L}ojasiewicz (PL) condition as a generalization of strong convexity to nonconvex settings \citep{polyak1963gradient}.
One of the recent lines of research is to loosen the conventional assumption of strong convexity and to show exponential convergence under the PL condition \citep{karimi2016linear,li2021page}.
For example, PAGE \cite{li2021page} achieves exponential convergence under the PL condition and the optimal rate in general nonconvex settings, but it should compute the full gradient at a certain probability.

Thus, one of our goals in this paper is to develop a single-loop variance reduction method as versatile as dominant algorithms such as SARAH, satisfying nearly optimal complexity for general nonconvex settings, second-order optimality, and exponential convergence under the PL condition.

\subsection{Federated learining}

Developing a single-loop variance reduction method essentially contributes to communication-efficient federated learning.
Federated learning is a paradigm of distributed learning in which optimization is performed by exchanging only model parameters of local clients without sending data out from them \citep{konevcny2016federated,shokri2015privacy}.
Here, we consider the following objective
\begin{align}
\label{eq:Intro-Federated}
    \min_{x\in \R^d} \left\{f(x):=\frac{1}{P}\sum_{i=1}^P\mathbb{E}_j[f_{i,j}(x)]\right\},
\end{align}
where $f_{i,j}$ is smooth.
Clients and data are indexed by $i$ and $j$. 
In such settings, synchronization and communication between all clients $f_i$ are often demanding.
Thus, many federated learning algorithms use
a variance reduction technique to allow sampling of clients by developing an estimator of $\nabla f(x^t)$ based on past gradients of the small number of clients sampled at each step \citep{karimireddy2020scaffold,karimireddy2021breaking}.

Such a technique has recently been combined with the local update, where we update parameters several times inside each client using only its data and then aggregate them in the server.
This can reduce the number of communication rounds and
communication complexity (the number of gradients communicated), especially when clients have less heterogeneity \citep{mcmahan2017communication,woodworth2020minibatch}.

However, here is a point that should not be overlooked.
Ignoring intra-client variance and allowing periodic full client synchronization, it is not difficult to combine local update with SARAH to achieve $\tilde{O}(\frac{1}{K\eps^2}+\frac{\zeta}{\eps^2})$ communication rounds, which approaches $\tilde{\Theta}(1)$ if (Hessian-)~heterogeneity of clients $\zeta$ (\Cref{assumption:Heterogeneity})
vanishes and the number of local steps $K$ gets large.
On the other hand, according to \cref{table:Intro-FL}, many existing methods including FedAvg, SCAFFOLD, and MimeMVR have additional terms other than $\frac{1}{K\eps^2}+\frac{\zeta}{\eps^2}$, meaning that they failed in fully exploiting local updates and small heterogeneity.
This is because these algorithms are based on variance reduction methods that do not require periodic full gradient but have sub-optimal gradient complexities. 
The only exception is BVR-L-SGD, but they require full client synchronization at every communication round.
We note that although the importance of handling client sampling errors has been mentioned by \citet{li2021zerosarah,jhunjhunwala2022fedvarp}, the rates of these methods are slower than $O\left(\frac{L}{\eps^2}\right)$.

Therefore, it is straightforward to develop a single-loop variance reduction algorithm and,
combining the local update,
to apply that for a federated learning algorithm allowing sampling of clients and requiring only $\tilde{O}(\frac{1}{K\eps^2}+\frac{\zeta}{\eps^2})$ communication rounds. 

\begin{table*}[t]\centering
  \begin{threeparttable}
  \caption{Comparison of communication rounds and complexity for a non-convex federated learning problem \eqref{eq:Intro-Federated}.}
  \label{table:Intro-FL}
  \begin{tabularx}{\linewidth}{p{75mm}Xp{45mm}p{22.5mm}}   \bhline{0.8pt}
  Algorithms  & Communication rounds & Client sampling (other then $x^0$)  \\  \bhline{0.8pt}
 FedAvg (nonconvex) {\small\citep{karimireddy2020scaffold}} & $\frac{\sigma^2_c}{p\eps^4}+\frac{\sigma_c}{\eps^3}+\frac{1}{\eps^2}$ & $\checkmark$\\ \hline
 SCAFFOLD (nonconvex) {\small\citep{karimireddy2020scaffold}} & $\frac{1}{\eps^2}(\frac{P}{p})^\frac23$ & $\checkmark$\\ \hline
 MimeMVR (nonconvex) {\small\citep{karimireddy2021breaking}} & $\frac{\zeta'\sigma_c}{\sqrt{p}\eps^3}+\frac{\sigma^2_c}{p\eps^2}+\frac{1}{K\eps^2}+\frac{\zeta'}{\eps^2}$  & $\checkmark$\\ \hline
 BVR-L-SGD (nonconvex) {\small\citep{murata2021bias}} &  $\frac{1}{K\eps^2}+\frac{\zeta}{\eps^2}$  & $\times$\\ \hline
 \rowcolor[rgb]{0.85, 1.0, 1.0} $\text{FLEDGE}$ (nonconvex) {\small(ours)} & $\frac{1}{K\eps^2}+\frac{\zeta\sqrt{P}}{p\eps^2}+\frac{\zeta}{\eps^2}$ & $\checkmark$\\ \hline
 BVR-L-PSGD (SOSP) {\small\citep{murata2022escaping}} &  $(\frac{1}{K}+\zeta)(\frac{1}{\eps^{2}}+\frac{1}{\delta^{4}})$  & $\times$\\ \hline
 \rowcolor[rgb]{0.85, 1.0, 1.0} $\text{FLEDGE}$ (SOSP) {\small(ours)}  & $(\frac{1}{K}+\zeta)(\frac{1}{\eps^{2}}+\frac{1}{\delta^{4}})$ & $\checkmark$ \text{\hspace{5mm}\small$(\text{requiring }p\gtrsim \sqrt{P}+\frac{\zeta^2}{\delta^2}
 )$}\\ \hline
 MimeSGD (PL) {\small\citep{karimireddy2021breaking}} & $\frac{\sigma^2_c}{\mu p\eps^2}+\frac{L}{\mu}$ & $\checkmark$\\ \hline
 \rowcolor[rgb]{0.85, 1.0, 1.0} $\text{FLEDGE}$ (PL) {\small(ours)} & $\frac{L}{\mu K}+\frac{\zeta\sqrt{P}}{\mu p}+\frac{\zeta}{\mu}+\frac{P}{p}$ & $\checkmark$\\ \bhline{0.8pt}
  \end{tabularx}
  {\footnotesize 
{\bf Note:} 
  $P$ is the number of clients, 
  $\mu$ is the parameter for PL condition, 
  $\sigma_c$ is the variance between clients, which can be as large as $P$.
  $\zeta$ is the Hessian-heterogeneity between clients and
  $\zeta'$ in MimeMVR is the Hessian-heterogeneity \textit{between all data} (i.e., $\|\nabla^2f_{i,j}(x)-\nabla^2f(x)\|\leq \zeta'$).  $\zeta'$ contains not only the inter-client Hessian-heterogeneity but also the intra-client Hessian-heterogeneity. Thus, $\zeta\leq \zeta' $ always holds and moreover it is possible that $\zeta \ll \zeta'$.
  Importantly, $\zeta \to 0$ does not necessarily means $\zeta'\to 0$.
  
  We here choose Option II for FLEDGE, where full participation of clients is conducted only once at $x^0$. 
  Option I allows client sampling even at $x^0$, at the cost of additional terms, as we detail in \Cref{section:Appendix:FLEDGE}.
  
  }
  \end{threeparttable}
\end{table*}

\subsection{Our contributions}
We first propose a novel, completely single-loop variance reduction method called SLEDGE (Single-Loop mEthoD for Gradient Estimator) 
that does not require periodic computation of full gradients (\cref{theorem:First-Main}) to achieve nearly optimal gradient complexity.
Unlike ZeroSARAH, our algorithm possesses the following advantages.
\begin{itemize}
    \item[(i)]Just adding small noises, SLEDGE efficiently escapes saddle points (\cref{theorem:Second-Main}).
    This is the first single-loop algorithm that does not require any conditional branch, except for Noisy SGD with a large minibatch.
    \item[(ii)]When SLEDGE enters the PL region, it automatically switches to exponential convergence (\cref{theorem:PL-Main}).
    PAGE \citep{li2021page} also has this property, but it requires periodic full gradients.
    \item[(iii)]
    In anticipation of application to federated learning, we analyze SLEDGE under Hessian-heterogeneity of $\zeta$, and show the improved complexity of $O(n+\frac{\zeta\sqrt{n}}{\eps^2})$ (Option II), matching the lower bound we give in  \cref{proposition:LowerBound}.
\end{itemize}
One shortcoming of the algorithm is the memory cost to save past gradients for each $i$.
However, this cost has been popular for avoiding full gradients \cite{defazio2014saga,roux2012stochastic,li2021zerosarah} and in federated learning literature \cite{karimireddy2020scaffold,murata2021bias}.

Next, we extend SLEDGE to an efficient federated algorithm named FLEDGE. 
Owing to (iii), the required number of the communication rounds in nonconvex settings is $\tilde{O}(\frac{1}{K\eps^{2}}+\frac{\zeta\sqrt{P}}{p\eps^{2}}+\frac{\zeta}{\eps^{2}})$, where $K$ is the number of local steps and $p$ is the number of sampled clients at each step. 
Notably, FLEDGE is the first algorithm such that the number of required communication rounds goes to $\tilde{\Theta}(1)$ when $K\to \infty$ and $\zeta \to 0$ while allowing sampling of clients.
For SOSPs and under the PL condition, it also surpasses existing algorithms.
See \cref{table:Intro-FL}.

Now we detail the superiority of FLEDGE.
Setting $p\geq \sqrt{P}$, this rate is always better than or at least equivalent to that of FedAvg, SCAFFOLD, and MimeMVR in all ranges of $\zeta$ and $K$.
We emphasize that MimeMVR used the stronger assumptions on the intra-client Hessian-heterogeneity and the first-order variance, see the note of \cref{table:Intro-FL}.
Moreover, although BVR-L-SGD uses the same communication rounds, it does not allow client sampling. 
Thus our algorithm requires lower communication complexity.
Furthermore, when $p= \sqrt{P}$ and $K\geq \zeta^{-1}$, the communication complexity of FLEDGE is $\tilde{O}(P+\frac{\zeta\sqrt{P}}{\eps^2})$ (Option II), which is nearly optimal in a sense that it matches the lower bound of gradient complexity in \cref{proposition:LowerBound}.
This lower bound means that the server must receive information of $\Omega(P+\frac{\zeta \sqrt{P}}{\eps^2})$ gradients to produce $\eps$-first-order solutions.

Moreover, FLEDGE inherits other advantages from SLEDGE as well.
Due to (i), it can find $(\eps,\delta)$-SOSPs using the same communication rounds as that of BVR-L-PSGD, which is the only federated learning algorithm with a second-order guarantee up to this time.
Again, when $K\to \infty$ and $\zeta \to 0$, the number of rounds goes to $\tilde{\Theta}(1)$, and our algorithm allows sampling of clients at each step while BVR-L-PSGD does not, yielding strictly fewer communication complexity.
In a classical setting of $\delta = O(\sqrt{\rho\eps})$, FLEDGE finds SOSPs without hurting optimal communication costs to find only first-order critical points.

In addition, under the PL condition, FLEDGE
is also the first algorithm that yields exponential convergence without suffering from the condition number
 when $\zeta$ is small, by exploiting the advantage of (ii).
The required communication round is $\tilde{O}(\frac{L}{\mu K}+\frac{\zeta}{\mu})$, taking $p$ sufficiently large.
As $\zeta$ gets small and the local budget gets large, it goes $\tilde{\Theta}(1)$.
Even in the strongly-convex case and without client sampling, all existing algorithms require $\tilde{\Theta}(\frac{L}{\mu})$ rounds. 

We emphasize that all these merits of FLEDGE come from the single-loop nature of SLEDGE.

\section{Preliminaries}\label{section:preliminaries}
 
Here we formally describe the problem settings.
First, gradient Lipschitzness and boundedness are assumed as usual.
\begin{assumption}[Gradient Lipschitzness]\label{assumption:GradLipschitz}
For all $i\in [n]$, $f_i$ is $L$-gradient Lipschitz, i.e., $\|\nabla f_i(x) - \nabla f_i(y)\|\leq L\|x-y\|,\forall x,y\in \R^d$.
For $f_{i,j}$, we also assume the same.
\end{assumption}
\begin{assumption}[Existence of global infimum]\label{assumption:GlobalInfimum}
$f$ has the global infimum $f^*=\inf_{x\in \R^d}f(x)$ and $\Delta:=f(x^0)-f^*$.
\end{assumption}
Below, (i) inter-client gradient boundedness is assumed for SLEDGE with Option I to remove full gradient even at $x^0$, as in ZeroSARAH \citep{li2021zerosarah}.
(ii) Intra-client gradient boundedness is assumed for FLEDGE.
\begin{assumption}[Boundedness of Gradient]
\label{assumption:BoundedGradient}
{\rm(i)}
For all $i$, $\|\nabla f_i(x^0)-\nabla f(x^0)\|^2\leq \sigma_c^2$.
{\rm(ii)} 
For all $i,j$ and $x$, $\|\nabla f_{i,j}(x)-\nabla f_{i}(x)\|^2\leq \sigma^2$.
\end{assumption}
Hessian Lipschiteness is usual to give second-order optimality \citep{ge2019stabilized,li2019ssrgd}.
\begin{assumption}[Hessian Lipschitzness]\label{assumption:HessianLipschitz}
$\{f_i\}_{i=1}^n$ is $\rho$-Hessian Lipschitz, i.e., $\|\nabla^2 f_i(x) - \nabla^2 f_i(y)\|\leq \rho\|x-y\|,\ \forall i\in [n]\text{ and }x,y\in \R^d$.
\end{assumption}
For federated learning, we solely assume inter-client Hessian-heterogeneity to show the efficiency of the proposed method in a less heterogeneous setting.
It has previously appeared in Mime \citep{karimireddy2021breaking} (but intra-client Hessian-heterogeneity was assumed at the same time) and BVR-L-SGD \citep{murata2021bias}.
\begin{assumption}[Hessian-heterogeneity]\label{assumption:Heterogeneity}
$\{f_i\}_{i=1}^n$ is Hessian-heterogeneous with $\zeta$, i.e., for any $i,j\in [n]$ and $x\in \R^d$, $\|\nabla^2 f_i(x) - \nabla^2 f_j(x)\|\leq \zeta$.
\end{assumption}
Finally, we explain the PL condition \cite{polyak1963gradient}. 
It is easy to see $\mu$-strongly convex function satisfies this with $\mu$.
\begin{assumption}[PL Condition]\label{assumption:PL}
$f$ satisfies PL condition, i.e., 
$\|\nabla f(x)\|^2\geq 2\mu \left(f(x)-f^*\right)$
for any $x\in \mathbb{R}^d$.
\end{assumption}

\section{Proposed Method: SLEDGE}\label{section:SLEDGE}
In this section, we concretely describe the proposed algorithm SLEDGE and provide its convergence guarantee.
In the following pseudocode, $B(0,r)$ denotes the uniform distribution on the Euclidean ball in $\R^d$ with radius $r$.
\begin{algorithm}[ht]
    \caption{\AlgMain$(x^0, \eta, b, T, r)$\label{algorithm:main}}
    \begin{algorithmic}[1]
        \State{{\bf Option I: }Randomly sample minibatch $I^{0}$ with size $b$}
        \State{\hspace{15mm}$y^0_i\leftarrow \frac{1}{b}\sum_{j\in I^0}\nabla f_j(x^0)\ (i=1,\cdots,n)$}
        \State{{\bf Option II: }$y^0_i\leftarrow \nabla f_i(x^0) \ (i=1,\cdots,n)$}
            \For{$t=1$ to $T$}
            \State{Ramdomly sample minibatch $I^{t}$ with size $b$}
            \State{$\displaystyle x^t \leftarrow x^{t-1} - \frac{\eta}{n} \sum_{i=1}^n y^{t-1}_i + \xi^t \quad (\xi^t \sim B(0,r))$}
            \State{$y^t_{i} \leftarrow \begin{cases}\nabla f_i (x^t) \quad {\rm for\ } i\in I^t \\ \frac{1}{b} \sum_{j\in I^t} (\nabla f_j(x^t) - \nabla f_j (x^{t-1}))+ y^{t-1}_{i}\\
            \hspace{45mm} {\rm for\ } i\notin I^t\end{cases}$}
            {\small (The update can be computed in $O(b)$ time. See ``efficient implementation" paragraph.)}
            \EndFor
    \end{algorithmic}
\end{algorithm}

\vspace{-4.5mm}

\subsection{Algorithm Description}
SLEDGE is designed so that it inherits the best points of SAGA \citep{defazio2014saga,reddi2016fast} and SARAH \citep{nguyen2017sarah,nguyen2017stochastic}, in that it does not require full gradient as SAGA and can construct an estimator with small variance as SARAH.

According to SAGA's update rule, the discrepancy of the gradient estimator from the true gradient at a step $t$ can be decomposed as
$$
\sum_{i=1}^n\frac{\nabla f_i(x^t) - \nabla f_i(x^{T(t,i)})}{n}
-\sum_{i\in I^t}\frac{\nabla f_i(x^t) - \nabla f_i(x^{T(t,i)})}{b}
,
$$
where $I^t$ is the randomly chosen minibatch with size $b$ and $T(t,i)$ is the step when $f_i$ is last sampled.
Note that SAGA stores $\nabla f_i(x^{T(t,i)})$ for each $i$.
Thus, the first term is a change from the referable gradient of $\frac1n\sum_{i=1}^n \nabla f_i(x^{T(t,i)})$, and the second term is an approximation of the first term using a minibatch with size $b$.
Then, the variance of the gradient estimator is roughly bounded by $\frac{1}{b}\|x^t-x^{T(t)}\|^2 \leq \frac{t-T(t)}{b}\sum_{s=T(t)+1}^t\|x^{s}-x^{s-1}\|^2$, with $T(t)=\min_i T(t,i)$.

On the other hand, 
the difference between SARAH's gradient estimator, which computes the full gradient periodically, and the true gradient can be written as
$$
  \sum_{\substack{s=T(t)+1}}^t\Bigl(\nabla f(x^s) -\nabla f(x^{s-1}) - \sum_{i\in I^s}\frac{\nabla f_i(x^s) - \nabla f_i(x^{s-1})}{b}\Bigr)    ,
$$
where $T(t)$ is the time of the last full gradient evaluation.
We can interpret this scheme as it decomposes $\nabla f(x^t)-\nabla f(x^{T(t)})$ into the sum of $\nabla f(x^s) -\nabla f(x^{s-1})$, and each term is approximated by an independent minibatch with size $b$.
Then, the variance is bounded by $\frac{1}{b}\sum_{s=T(t)+1}^t\|x^{s}-x^{s-1}\|^2$, meaning that
SARAH's estimator is better than that of SAGA by the $t-T(t)$ factor, which can be as large as $\frac{n}{b}$.

Based on the above discussion, we first decompose SAGA's approximation target $\sum_{i=1}^n\frac{\nabla f_i(x^t) - \nabla f_i(x^{T(t,i)})}{n}$ into the sum of $\nabla f_i(x^s) - \nabla f_i(x^{s-1})$, each of which is approximated in SARAH's manner. 
Namely, the decomposed form is written as follows:
\begin{align}
    \frac1n \sum_{s=T(t)+1}^n \sum_{i\in \tilde{I}_s^t}^n (f_i(x^s) - f_i(x^{s-1})).
\end{align}
Here $\tilde{I}_s^t = [n]\setminus \bigcup_{\tau=s}^t I^t$,
so that $\tilde{I}_s^t$ is the set of indexes not sampled between $s$ and $t$.
Then, we approximate $\sum_{i\in \tilde{I}_s^t}^n (f_i(x^s) - f_i(x^{s-1}))$ with $\frac{|\tilde{I}_s^t|}{b}\sum_{i\in I^s}^n $ $(f_i(x^s) - f_i(x^{s-1}))$.
This procedure yields \cref{algorithm:main}, and the following error bound on the SLEDGE estimator.
\begin{lemma}[Informal]\label{lemma:First-VarianceBound}
Let $\nu\in (0,1)$, $T_1 = \tilde{O}(\frac{n}{b})$, and $C=\tilde{O}(1)$.
We have that, ignoring the initialization error, with probability $1-\nu$ for all $t=1,\cdots,T$,
\begin{align}
    \left\|\nabla f(x^t)-\frac1n\sum_{i=1}^n y^t_i\right\|^2 \leq 
    \frac{C\zeta^2}{b}\sum_{\substack{s =1\lor (t-T_1+1)}}^{t}\|x^s -x^{s -1}\|^2.
\end{align}
\end{lemma}
Here, $T_1$ is defined so that $T_1 \geq t - T(t)$ holds with high probability.
This lemma tells us that our gradient estimator has comparable quality to SARAH without computing full gradient.
Moreover, this lemma explicitly states that the variance of SLEDGE estimator is quadratically bounded with $\zeta$, meaning that we require fewer gradients when $\zeta \ll L$, which is later exploited for federated learning application.

While the development is intuitively straightforward, we have
the technical difficulty to evaluate the error, that $|\tilde{I}_s^t|$ depends not only on $I^s$ but also on $I^{s+1},\dots,I^t$.
In other words, unlike SARAH, the discrepancy cannot be decomposed into completely independent terms about $I^s$, 
which prevents us from using a usual expectation bound.
To address this, we prepared vector Bernstein inequality without replacement (\cref{proposition:BernsteinNoReplacement}) to give a high probability bound on the discrepancy.



\paragraph{Efficient implementation} 
Note that we can update 
$\frac1n\sum_{i=1}^n y_i^t$ in $O(bd)$ time and using $O(nd)$ memory.
Indeed, first introduce an auxiliary variable $v^t$ which is inductively defined by $v^t = \frac{1}{b}\sum_{i\in I^t}(\nabla f_i(x^t)-\nabla f_i(x^{t-1})) + v^{t-1}$ with $v^0=0$.
For each $i$, define $v^t_i$ with $v_i^t=0$ and update it as $v^t_i = v^t$ iff $i\in I^t$.
We also define $w_i^t$ with $w_i^0=y_i^0$ and update it as $w^t_i = y^t_i = \nabla f_i(x^t)$ iff $i\in I^t$.
Now we can see that
$\frac1n\sum_{i=1}^n y_i^t = \frac{1}{b}\sum_{i\in I^t}(\nabla f_i(x^t)-\frac{n-b}{n}\nabla f_i(x^{t-1})) + \frac1n\sum_{i=1}^n y_i^{t-1} - \frac1n\sum_{i\in I^t} (w_i^{t-1}+v_i^t-v_i^{t-1})$.
Therefore, $\frac1n\sum_{i=1}^n y_i^t$ can be updated by only $O(bd)$ computation with $O(nd)$ memory.

\subsection{Convergence guarantee}

Based on 
Lemma 1, we present the following convergence guarantee for the problem \eqref{eq:Intro-FiniteSum}.

\begin{theorem}\label{theorem:First-Main}
Under \cref{assumption:GradLipschitz,assumption:GlobalInfimum,assumption:Heterogeneity}, and \ref{assumption:BoundedGradient}-(i) for Option I, if we choose
$\eta = \tilde{\Theta} (\frac{1}{L}\land
\frac{b}{\zeta\sqrt{n}}) \text{ and } r \leq \frac{\eta\eps}{2}$,
\AlgNumMain finds $\eps$-first-order stationary points using
\begin{align}
    \tilde{O}\left(\frac{\Delta \left(\zeta \sqrt{n}\lor Lb\right)+ \frac{n}{b}\sigma_c^2}{\eps^2} \right)
    \quad (\text{\rm Option I}),\\
    \tilde{O}\left(n+\frac{\Delta \left(\zeta \sqrt{n}\lor Lb\right) }{\eps^2}\right)\quad (\text{\rm Option II})  
\end{align}
stochastic gradients with probability at least $1-\nu$.

\end{theorem}
Remember that $\zeta$ is at most $2L$.
Thus, SLEDGE with Option I can achieves nearly optimal convergence rate completely without full gradient.
Our analysis covers all range of $b$, while ZeroSARAH only is only verified for $b=\sqrt{n}$.
Option II uses full gradient at the initial point, which is necessary to avoid dependency on $\sigma_c$. 
The rate is equivalent to existing algorithms such as SPIDER and SARAH, and optimal even considering the dependency on $\zeta$, up to $\log$ factors.

The formal lower bound is stated as follows, which is not difficult to derive, using the results by  \citet{carmon2020lower,fang2018spider,li2021page}. See \Cref{section:lowerbound} for details.
\begin{proposition}\label{proposition:LowerBound}
    Assume \cref{assumption:GradLipschitz,assumption:GlobalInfimum,assumption:Heterogeneity}.
    Then, any linear-span first-order algorithm requires
    \begin{align}
        \Omega\left(n+\frac{\Delta(\zeta\sqrt{n} +L)}{\eps^2}\right)
    \end{align}
    stochastic gradients to find $\eps$-first-order stationary points of the problem \eqref{eq:Intro-FiniteSum}.
\end{proposition}

\paragraph{Finding Second-order Stationary Points}
The proposed method goes beyond finding first-order stationary points.
One of the notable features of SLEDGE is that it can find second-order stationary points.
To our knowledge, all existing algorithms require sub-routine for negative curvature extraction (e.g., SPIDER-$\text{SFO}^{+}+$Neon2 \citep{fang2018spider}) or periodic large minibatch as large as $n$ or $O(\frac{\sigma_c^2}{\eps^2})$ (e.g., SSRGD \citep{li2019ssrgd}).
On the other hand, SLEDGE requires only adding small noise at each update to escape saddle points, and therefore SLEDGE is the first completely single-loop algorithm that can find SOSPs.
Indeed, we have the following theorem.

\begin{theorem}\label{theorem:Second-Main}
Assume \cref{assumption:GradLipschitz,assumption:GlobalInfimum,assumption:Heterogeneity,assumption:HessianLipschitz}, \ref{assumption:BoundedGradient}-(i) for Option I.
Let $b\gtrsim \sqrt{n}+\frac{\zeta^2}{\delta^2}$, $\eta = \tilde{\Theta}(\frac1L)$, $r\lesssim\tilde{O}\left(\frac{\eps}{L}\right)$, and $\nu\in (0,1)$.
Then,  \AlgNumMain\ finds $(\eps,\delta)$-SOSPs using
\begin{align}
    \tilde{O}\left(\left(L\Delta + \sigma_c^2\right)\left(\frac{1}{\eps^2}+\frac{\rho^2}{\delta^4}\right)b\right)
    \quad &(\text{\rm Option I}),\\ 
    \tilde{O}\left(n+L\Delta \left(\frac{1}{\eps^2}+\frac{\rho^2}{\delta^4}\right)b\right)
    \quad &(\text{\rm Option II})
\end{align}
stochastic gradients, with probability at least $1-\nu$.
\end{theorem}
Note that our bound requires minibatch size of $b\gtrsim \sqrt{n}+\frac{\zeta^2}{\delta^2}$, but this minibatch size is common in many existing algorithms.
In fact, SSRGD assumes $b\geq \sqrt{n}$ or $b\geq \frac{\sigma_c}{\eps}$ and Stabilized SVRG assumes $b\geq n^\frac23$.
Considering that $\delta=O(\sqrt{\rho\eps})$ is often assumed, our minibatch size is as moderate as existing algorithms.
The necessity of this assumption comes from that
if $b$ is too small, the sampling error hides the right direction of negative curvature.

Since $\zeta \leq 2L$, we can see that the complexity is comparable to existing algorithms, such as SPIDER-$\text{SFO}^{+}$ and SSRGD.
In addition to the $\frac{\sqrt{n}}{\eps^2}+\frac{\sqrt{n}}{\delta^4}$ term, different algorithms have different additional terms due to the technical reasons.
Ours is $\frac{\zeta^2}{\eps^2\delta^2}+\frac{\zeta^2}{\delta^6}$, which is smaller than that of SPIDER-$\text{SFO}^{+}$ when $n\gtrsim \delta^{-4}$ and that of SSRGD when $n\gtrsim \min \{\delta^{-3},\delta\eps^{-2}\}$.
\cref{remark:ReduceDeltaDependency} in the appendix introduces a small trick to reduce this term to $\frac{\zeta^2}{\eps^2\delta}+\frac{\zeta^2}{\delta^5}$ or $\frac{n\delta}{\eps^2}+\frac{n}{\delta^3}$.

Finally, we briefly explain the proof outline.
As in \citet{jin2017escape,ge2019stabilized,li2019ssrgd}, we consider the two coupled sequences with slightly different initial points.
We can show that, when negative curvature exists, these two sequence separate each other exponentially.
In other words, one of these sequences goes further from the initial point.
This in turn means that if we perturb the algorithm a little around saddle points, then it can escape saddle points with high probability. 
Although this proof technique is classical, we confront the following difficulties. First, while other algorithms refresh their gradient estimators around saddle points, our single-loop algorithm does not. Thus, we have to address the error up to that point, and it is not trivial whether our gradient estimator can sense the right direction of the negative eigenvalue despite the accumulated sampling error. Second, our estimator is more correlated due to the $|\tilde{I}_s^t|$ term, thus requiring more delicate analysis.

\paragraph{Exponential convergence under PL condition}
We can further show that when SLEDGE enters the PL region, it automatically switches into an exponential convergence phase.
\begin{theorem}\label{theorem:PL-Main}

Assume Assumptions \ref{assumption:GradLipschitz}, \ref{assumption:GlobalInfimum}, \ref{assumption:BoundedGradient}-(i), \ref{assumption:Heterogeneity}, and \ref{assumption:PL}.
If 
$\eta =  \tilde{\Theta}(\frac{1}{L}\land \frac{b}{\zeta\sqrt{n}}\land\frac{b}{\mu n}), \text{ and } r \leq \eta\sqrt{\frac{\eps\mu}{3}}
$,
\AlgNumMain with Option I finds an $\eps$-solution with $f(x^t)-f^*\leq \eps$ using
\begin{align}
    \tilde{O}\left( \left(\frac{Lb}{\mu}\lor \frac{\zeta \sqrt{n}}{\mu}\lor n\right)\log \frac{\Delta+\sigma_c}{\eps}\right) 
\end{align}
stochastic gradients with probability at least $1-\nu$.
$\tilde{O}$ hides at most $\log^{5.5} (n+\mu^{-1}+\nu^{-1})$ and ${\rm polyloglog}$ 
factors. 
\end{theorem}
Option II does not require dependency on $\log\sigma_c$.
Compared to PAGE, while PAGE computes full gradient probabilistically, ours completely removed the requirement of full gradient, even at the initial point.


\section{Application to Federated Learning}\label{section:FLEDGE}
We further propose the federated learning extension called FLEDGE, which inherits the advantages of SLEDGE.
The proposed algorithm uses SLEDGE for update of global parameters, while adopting a SARAH-type gradient estimator for local steps. 

Here we only present FLEDGE with Option II, that uses full participation of the clients only at $x^0$.
Other algorithms including MimeMVR do similarly, but we also provide Option I in the appendix, which completely removes the requirement of full participation.

\begin{algorithm}[ht]\label{alg:dist}
    \caption{\AlgDist$(x^0, \eta, p, b, T, K ,r)$}
    \begin{algorithmic}[1]
        \State{{\bf for }$i\in I^0 = I$ in parallel {\bf do}}
        \State{\hspace{3mm}Randomly select minibatch $J_i^0$ with size $Kb$}
        \State{\hspace{3mm}$y^0_i\leftarrow \frac{1}{bK}\sum_{j\in J_i^0} \nabla f_{i,j}(x^0)$}
        \State{{\bf for }$t=1$ to $T$ {\bf do}}
        \State{\hspace{3mm}Randomly sample one client $i_t$}
        \State{\hspace{3mm}Send $\frac{1}{P} \sum_{i=1}^P y^{t-1}_i$ and $x^{t-1}$ from the server to $i_t$}
        \State{\hspace{3mm}$x^{t,0}\leftarrow x^{t-1},\ z^{t,0}\leftarrow 0$}
        \State{\hspace{3mm}{\bf for }$k=1$ to $K$ {\bf do}}
        \State{\hspace{6mm}$\xi^{t,k}\sim B(0,r)$}
        \State{\hspace{6mm}$x^{t,k} \leftarrow x^{t,k-1} - \eta(\frac{1}{P} \sum_{i=1}^P y^{t-1}_i + z^{t,k-1})+\xi^{t,k}$}
        \State{\hspace{6mm}Randomly select minibatch $J^{t,k}_{i_t}$ with size $b$}
        \State{\hspace{6mm}$z^{t,k} \leftarrow z^{t,k-1} +  \sum_{j\in J^{t,k}_{i_t}}\frac{\nabla f_{i_t,j}(x^{t,k}) - \nabla f_{i_t,j} (x^{t,k-1})}{b}$}
            \State{\hspace{3mm}$x^{t}\leftarrow x^{t,K}$}
            \State{\hspace{3mm}Randomly select $p$ clients $I^{t}$}
            \State{\hspace{3mm}Send $x^t$ from $i_t$ to $I^{t}$}
            \State{\hspace{3mm}{\bf for }$i\in I^t$ in parallel {\bf do}}
            \State{\hspace{6mm}Randomly select minibatch $J^{t}_i$ with size $Kb$}
            \State{\hspace{6mm}$y^t_{i} \leftarrow \frac{1}{bK}\sum_{j\in J^{t}_i}\nabla f_{i,j}(x^t)$}
            \State{\hspace{6mm}$\Delta y^t_{i} \leftarrow \frac{1}{bK}\sum_{j\in J^{t}_{i}}(\nabla f_{i,j}(x^t)-\nabla f_{i,j}(x^{t-1}))$}
       \State{\hspace{3mm}Send $\{(y^{t}_i,\Delta y^{t}_i)\}_{i\in I^t}$ from $I^t$ to the server}
    \State{\begin{varwidth}[t]{\linewidth}
      \hspace{3mm}$y^t_{i} \leftarrow y^{t-1}_{i} + \frac{1}{p}\sum_{i\in I^t}\Delta y^t_i\ (\text{for }i\notin I^{t})$\par
\hspace{3mm}\small (Practically, we only update $\frac{1}{P} \sum_{i=1}^P y^{t}_i$ in $O(p)$ time.)
      \end{varwidth}}
    \end{algorithmic}
\end{algorithm}

\subsection{Convergence guarantee}
\AlgNumDist finds first-order stationary points for the problem \eqref{eq:Intro-Federated}.
The assertion is formalized as follows.
\begin{theorem}\label{theorem:Dist-First-Main}

Under 
Assumptions \ref{assumption:GradLipschitz}, \ref{assumption:GlobalInfimum}, \ref{assumption:BoundedGradient}-(ii), and \ref{assumption:Heterogeneity}, 
let 
$\eta =\tilde{\Theta}(\frac{1}{L}\land \frac{p\sqrt{b}}{\zeta\sqrt{PK}}\land \frac{1}{\zeta K}\land \frac{\sqrt{b}}{L\sqrt{K}})$, $r\leq \frac{\eta\eps}{2\sqrt{2}}$, 
$b \gtrsim\frac{\sigma^2}{PK\eps^2}$, and $\nu\in (0,1)$.
FLEDGE finds $\eps$-first-order stationary points using

\begin{align}
   \tilde{O}\left(
    1+\left(\frac{L}{K}\lor\frac{\zeta\sqrt{P}}{p}\lor \zeta \lor \frac{L}{\sqrt{Kb}}\right)\frac{\Delta}{\eps^2}
    \right)
\end{align}
communication rounds, with probability at least $1-\nu$.
\end{theorem}
Setting $p\geq\sqrt{P}$ and $b\geq K$, FLEDGE requires only $\tilde{O}(\frac{\zeta}{K\eps^2}+\frac{\zeta }{\eps^2})$ communication rounds, which approaches to $\tilde{\Theta}(1)$ by letting $K\to \infty$ and $\zeta\to 0$.
Thus, FLEDGE is the first algorithm with this property and allowing sampling of clients.
As a result, our algorithm requires only $\tilde{O}(P+\frac{\zeta \sqrt{P}}{\eps^2})$ communication complexity, which is superior to any existing algorithm.
Remembering \cref{proposition:LowerBound}, this rate is optimal in a sense that the server must receive information of $\Omega(P+\frac{\zeta \sqrt{P}}{\eps^2})$ gradients to output $\eps$-first-order solutions.

\paragraph{Finding Second-order Stationary Points}
By adding a small noise, FLEDGE can also find SOSPs.
\begin{theorem}\label{theorem:D-Second-Main}
%

Under 
Assumptions \ref{assumption:GradLipschitz}, \ref{assumption:GlobalInfimum}, \ref{assumption:BoundedGradient}-(ii), \ref{assumption:HessianLipschitz}, and \ref{assumption:Heterogeneity}, 
and $\delta <\zeta$,
let $p\gtrsim \sqrt{P}+\frac{\zeta^2}{\delta^2}+\frac{L^2}{Kb\delta^2}, b\geq K$, $\eta = \tilde{\Theta}(\frac{1}{L})$, $r\lesssim \frac{\eps}{L}$, $b \gtrsim\frac{\sigma^2}{PK\eps^2}$ and $\nu\in (0,1)$.
Then,  \AlgDist\ finds $(\eps,\delta)$-second-order stationary points using
\begin{align}
    \tilde{O}\left(1+ \Delta\left(\frac{L}{K}+\zeta\right)\left(\frac{1}{\eps^2}+\frac{\rho^2}{\delta^4}\right)\right)
\end{align}
communication rounds, with probability at least $1-\nu$.
\end{theorem}
The proposed algorithm is the first algorithm that can find SOSPs while allowing sampling of the clients.
It uses the same number of communication rounds as that of BVR-L-PSGD, but requires the improved communication complexity when $
\frac{\zeta^2}{\delta^2}\lesssim P$.
Note that here we have an additional assumption of $\delta < \zeta$, but \cref{lemma:D-Small-Stuck-Region-Relax} in the appendix will remove this with small modification of the algorithm.

\begin{figure*}[htbp]
\begin{tabular}{cc}
    \begin{minipage}[t]{0.35\hsize}
        \begin{center}
        \vspace{-0.63cm}
        \includegraphics[width=65mm]{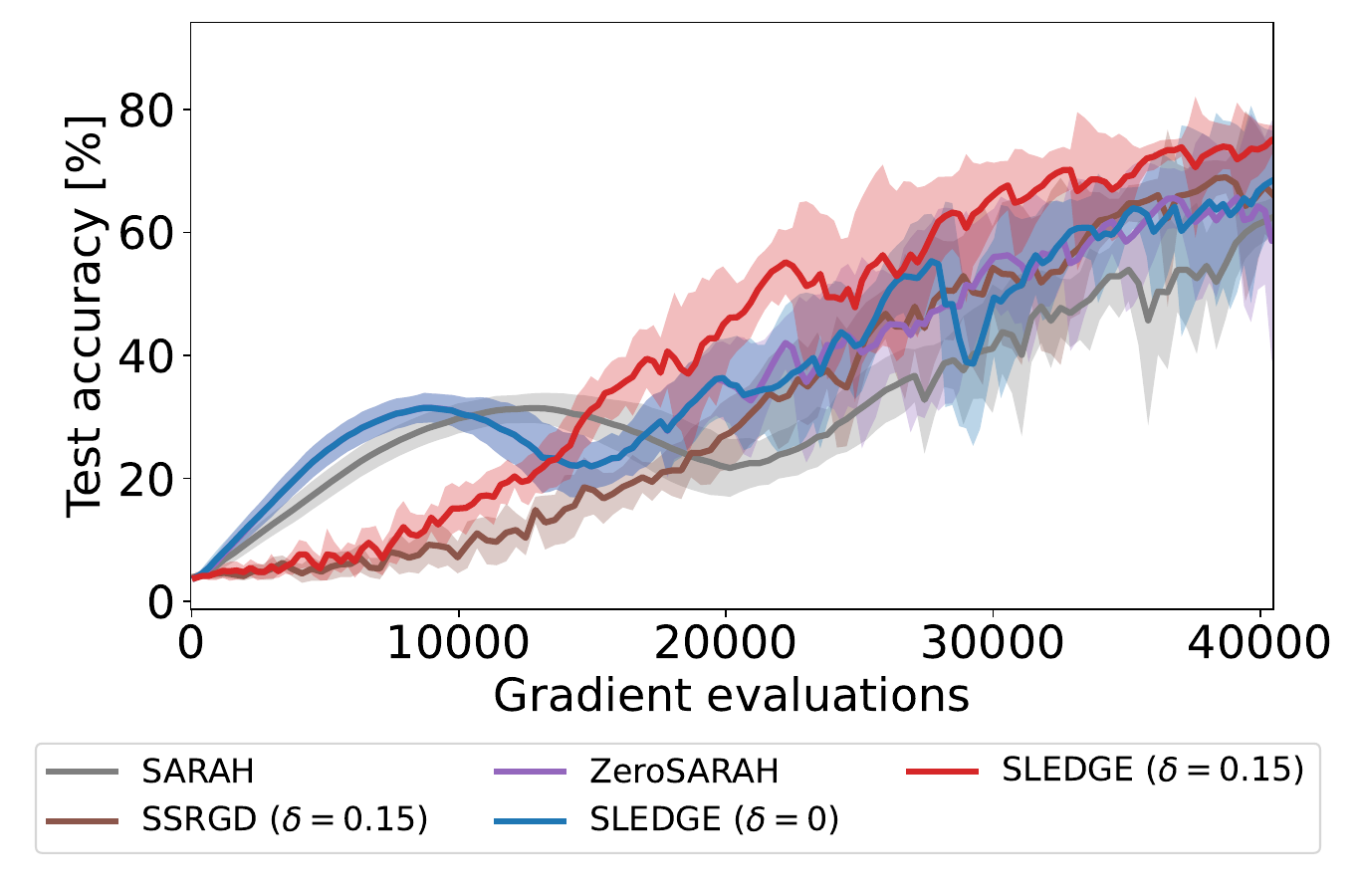}
        \end{center}
        \vspace{-0.55cm}
        \caption{Comparison of the test accuracy\\ (finite-sum setting)}
        \label{fig:SLEDGE_train_acc}
    \end{minipage} 
    &
    \begin{minipage}[t]{0.65\hsize}
        \begin{center}
        \vspace{-0.55cm}
        \includegraphics[width=105mm]{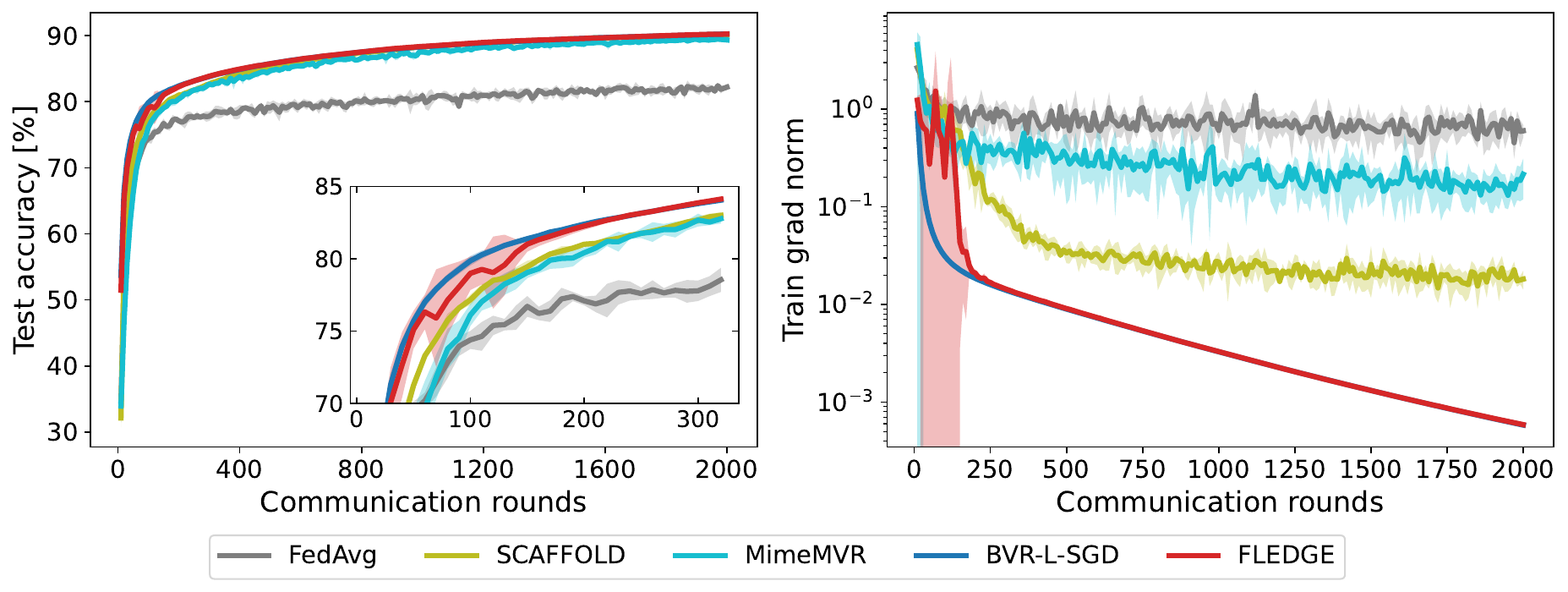}
        \end{center}
        \vspace{-0.41cm}
        \caption{Comparison of the test accuracy (left) and the gradient norm (right)\\ (federated learning)}
        \label{fig:SLEDGE_test_acc}
    \end{minipage} 
\end{tabular}
\vspace{-3mm}
\end{figure*}


\paragraph{Exponential convergence under PL Condition}
Furthermore, FLEDGE also automatically switches to exponential convergence when entering the PL region.

\begin{theorem}\label{theorem:D-PL}


Under 
Assumptions \ref{assumption:GradLipschitz}, \ref{assumption:GlobalInfimum}, \ref{assumption:BoundedGradient}-(ii), and \ref{assumption:HessianLipschitz} to \ref{assumption:PL}, 
letting
$\eta = \tilde{\Theta}(
    \frac{1}{L}\land
    \frac{p\sqrt{b}}{\zeta \sqrt{PK}}\land
    \frac{p}{\mu PK}\land
    \frac{1}{\zeta K}\land
    \frac{\sqrt{b}}{L\sqrt{K}} 
    ),b \gtrsim\frac{\sigma^2}{PK\eps^2}\text{ and } r \lesssim \eps\sqrt{\eta}\text{, and }\nu \in (0,1)
$, \AlgDist\ finds $\eps$-first-order stationary points using
\begin{align}
     \tilde{O}\left(1+ \left(
    \frac{L}{\mu K}\land
    \frac{\zeta \sqrt{P}}{\mu p}\land
    \frac{ P}{p}\land
    \frac{\zeta }{\mu}\land
    \frac{L}{\mu\sqrt{Kb}}
    \right)\log \frac{\Delta}{\eps}\right)
\end{align}
communication rounds with probability $1-\nu$.
Here $\tilde{O}$ hides $\log^{6.5} (P+K+\mu^{-1}+\nu^{-1})$ and ${\rm polyloglog}$ factors.
\end{theorem}
By setting $K,b$ and $p$ sufficiently large, the algorithm requires $\tilde{O}(\frac{\zeta}{\mu}\log \frac{\Delta}{\eps})$ communication rounds. 
Thus, even if $\mu$ is small, the required communication rounds goes to $\tilde{O}(1)$ when $\zeta\to 0$.
On the other hand, MimeSGD depends on condition number $\frac{L}{\mu}$ even if $\zeta$ is small, and other algorithms do so even under the strong convexity.


\section{Numerical Experiment}\label{section:experiment}
We conducted numerical experiments to show the effectiveness of SLEDGE and FLEDGE.
Detailed explanation and additional experiments can be found in \Cref{section:AppendixExperiments}.

\subsection{Escaping saddle points with SLEDGE}
For the finite-sum problem \eqref{eq:Intro-FiniteSum}, we consider a classification of the capital letters using EMNIST dataset \citep{cohen2017emnist}.
We prepared each $f_i$ by sampling $100$ data from one class, employing a four-layer neural network as the training model, and then defining the average of the cross-entropy loss over the data as $f_i$.
We repeated this five times for each class, and thus $n=130$.
We set $b=12$, the inner-loop length of SARAH and SSRGD to $10$, and $\lambda =\frac{b}{n} \fallingdotseq 0.092$ for ZeroSARAH. 
Then, we compared SLEDGE with SARAH, SSRGD, and ZeroSARAH, in terms of the test accuracy.
For SSRGD and SLEDGE, we add small perturbation of $\delta=0.09$.
We tuned the learning rate for each algorithm individually.
The experiment was repeated with ten different random seeds for each method.

\Cref{fig:SLEDGE_train_acc} shows the result.
We can observe that (i) SLEDGE and ZeroSARAH require fewer gradient evaluations than SARAH to achieve the same test accuracy, owing to avoidance of periodic full gradient.
(ii) Adding small noise helps stable convergence; 
Around $10000$-$15000$ gradient evaluations, although SLEDGE with $\delta =0$ does not necessarily yield a monotonic increase in the accuracy, SLEDGE with small noise perturbation makes the accuracy increase almost monotonically.
In summary, SLEDGE with small noise yields the fastest and most stable convergence. 


\subsection{Faster Convergence with FLEDGE}
For the federated learning problem \eqref{eq:Intro-Federated}, we again consider the classification of the capital letters, where each $f_i$ consists of $90\%$ data from one class and $10\%$ data from the other classes.
This makes each $f_i$ a little less heterogeneous.
We used two-leyer neural networks with width of the hidden layer $100$.
We compared FLEDGE with FedAvg, SCAFFOLD, MimeMVR, and BVR-L-SGD.
For each algorithm, we employed $P=104$ as the total number of clients and $p=10$ as the number of the clients used at each communication (except for BVR-L-SGD, which requires $P=p=104$).
Then, we set $b=16$ and $K=10$.
We tuned the learning rate for each algorithm individually.
The experiment was repeated with five different random seeds for each method.

\Cref{fig:SLEDGE_test_acc} (left) shows that 
FLEDGE achieves the high test accuracy with fewer communication, compared to FedAvg, SCAFFOLD, and MimeMVR.
In \Cref{fig:SLEDGE_test_acc} (right), FLEDGE achieves the small gradient norm $\|\nabla f(x^t)\|$ and the linear convergence at the neighborhood of solutions.
Moreover, we observe that FLEDGE performs similarly to BVR-L-SGD, which is almost a special case of FLEDGE with $P=p$.
This means that FLEDGE can appropriately correct the errors from sampling of the clients and is about ten times more efficient than BVR-L-SGD in terms of communication complexity by allowing sampling of the clients.
In summary, these experimental results strongly validate our theoretical guarantees about FLEDGE.

\section{Conclusion}
In this paper, we first developed a versatile single-loop gradient estimator, and showed first and even second-order optimality and faster convergence under PL condition, with explicit dependency on the Hessian-heterogeneity $\zeta$.
Then, based on this solid algorithm, we build a federated learning algorithm that allows client sampling, and extensively showed its inherited advantages.
Especially, the dependency of communication rounds and complexity on the Hessian-heterogeneity $\zeta$ improves many existing algorithms.

\section*{Acknowledgement}
The authors would like to thank Yuki Yoshida for his helpful advice on efficient implementation of the proposed methods.
KO was partially supported by IIW program of The Univ. of Tokyo. 
SA was partially supported by JSPS KAKENHI (JP22J13388).
TS was partially supported by CREST (JPMJCR2015, JPMJCR2115).


\bibliography{aaai23}


\onecolumn

\appendix
\section{Additional experiments}\label{section:AppendixExperiments}

\subsection{Details of the experiment for Figure 1}

We consider a classification of the capital letters using EMNIST By\_Class dataset \citep{cohen2017emnist}.
The original dataset consists of $814,255$ images of handwritten uppercase and lowercase letters and numbers $0$-$9$. 
Note that the number of data points in each class is not balanced. 
Since the number of images of lowercase letters is relatively small, 
we only used the images of uppercase letters for the experiment.
To balance the number of data points between each class, we took the following procedure.
We repeatedly sampled $100$ data points five times per each uppercase letter, which yields $26\times 5 = 130$ groups of sampled data.
For each group $i$, we define $f_i$ as the average of the cross-entropy loss between the output of the model and the true class, over the $100$ data points belonging to the group.
As a model, we adopted a four-layer fully-connected neural network, following \citet{murata2022escaping}.
We added $L_2$-regularizer with a regularization
parameter of $\lambda = 0.01$ to the empirical risk.

As competitors, we implemented SARAH \citep{nguyen2017sarah,nguyen2017stochastic}, SSRGD \citep{li2019ssrgd}, and ZeroSARAH \citep{li2021zerosarah}.
We set the minibatch size to $b=12\fallingdotseq \sqrt{n}=\sqrt{130}$ for all algorithms, the inner-loop length of SARAH and SSRGD to $m=\lfloor \frac{n}{b}\rfloor=10$, and $\lambda =\frac{b}{n} \fallingdotseq 0.092$ for ZeroSARAH.
Note that \citep{li2021zerosarah} adopted $\lambda = \frac{b}{2n}$, but we found that $\lambda =\frac{b}{n}$ was more stable in this setting. 
The learning rate for each method was tuned individually, from $\{1.0, 0.3, 0.1, 0.03, 0.01, 0.003, 0.001\}$, so that the test accuracy after $2000$ iterations is the highest.
For SSRGD and noisy SLEDGE, we added small noise of $r=0.15$.
We plotted the mean of the ten trials with different random seeds and the sample variance is also shown in the corresponding (lighter) color for each algorithm.

\subsection{Details of the experiment for Figure 2}\label{subsection:Appendix_A2}

We consider a classification of the capital letters using EMNIST By\_Class dataset \citep{cohen2017emnist} as well.
However, here we divided the images in such a way that $f_i$ is a little less heterogeneous, but still more heterogeneous than i.i.d. sampling, as follows.
First, we prepared the same number of data points for each class, and divided them into each client $i$ by the following procedure, setting $q=0.9$;
Then, for each class, we distributed $q\times 100\%$ of the images into four clients, and the rest into the remaining $100$ clients.
This yields that we have $4\times 26=104$ clients, each of which contains $q\times 100\%$ of the data from one class, and $(1-q)\times 100\%$ of the data from the other classes.
We call this grouping as a dataset with the heterogeneity parameter of $q$.
Then, we constructed $f_{i,j}$ with the cross-entropy loss and a two-layer fully-connedted neural network, following \citet{murata2021bias}.
$L_2$-regularizer with a scale of $\lambda = 0.01$ is added to the empirical risk.


We compared FLEDGE with FedAvg \citep{mcmahan2017communication}, SCAFFOLD \cite{karimireddy2020scaffold}, MimeMVR \cite{karimireddy2021breaking}, and BVR-L-SGD \cite{murata2021bias}.
For each algorithm, we set $p=10 \fallingdotseq \sqrt{P}=\sqrt{104}$ as the number of the clients used at each communication (except for BVR-L-SGD, which requires $p=P=104$).
Then, we set the local minibatch size as $b=16$ and the number of local update to $K=10$.
We tuned the learning rate for each algorithm individually from $\{1.0, 0.3, 0.1, 0.03, 0.01, 0.003, 0.001\}$, so that the test accuracy after $2000$ outer-loop iterations is the highest.
Here we 
set the global learning rate of SCAFFOLD to $\eta = 1$, as is done in the original paper \citep{karimireddy2020scaffold}.
MimeMVR adopted a momentum parameter of $a=0.1$ as the authors of the paper reported as the best.
We plotted the mean of the five trials with different random seeds and the sample variance is also shown in the corresponding (lighter) color for each algorithm.

\subsection{Additional experiments for SLEDGE}
\paragraph{Comparison with SARAH by changing the learning rate}
Here we provide comparison of SLEDGE with SARAH \citep{nguyen2017sarah,nguyen2017stochastic}, which is one of the most prevailing variance reduction algorithm with theoretical optimal complexity of $O(\frac{\sqrt{n}}{\eps^2})$.

As is done in the experiment for Figure 1,we prepared $f_i$ in the following way. 
We repeatedly sampled $100$ data points five times per each uppercase letter, which yields $26\times 5 = 130$ groups of sampled data.
For each group $i$, we define $f_i$ as the average of the cross-entropy loss between the output of the model and the true class over the $100$ data points belonging to the group.
As a model, we adopted a two-layer fully-connected neural network, following \citet{murata2021bias}.
We set the minibatch size to $b=12\fallingdotseq \sqrt{n}=\sqrt{130}$ for both algorithms, and 
the inner-loop length of SARAH to $m=\lfloor \frac{n}{b}\rfloor=10$.
We added $L_2$-regularizer to the empirical risk with a fixed regularization
parameter of $\lambda = 0.01$.
We compared SLEDGE with SARAH in terms of the training loss, the norm of the gradient computed by the whole training data, the test loss, and the test accuracy, under the same number of stochastic gradient accesses.
We changed the learning rate $\eta$ between $\{0.1, 0.03, 0.01, 0.003, 0.001\}$.
We plotted the mean of the five trials with different random seeds and the sample variance is also shown in the corresponding (lighter) color for each algorithm.

\begin{figure}[htbp]
    \begin{tabular}{cccc}
      \begin{minipage}[t]{0.23\hsize}\centering
        \includegraphics[width=40mm]{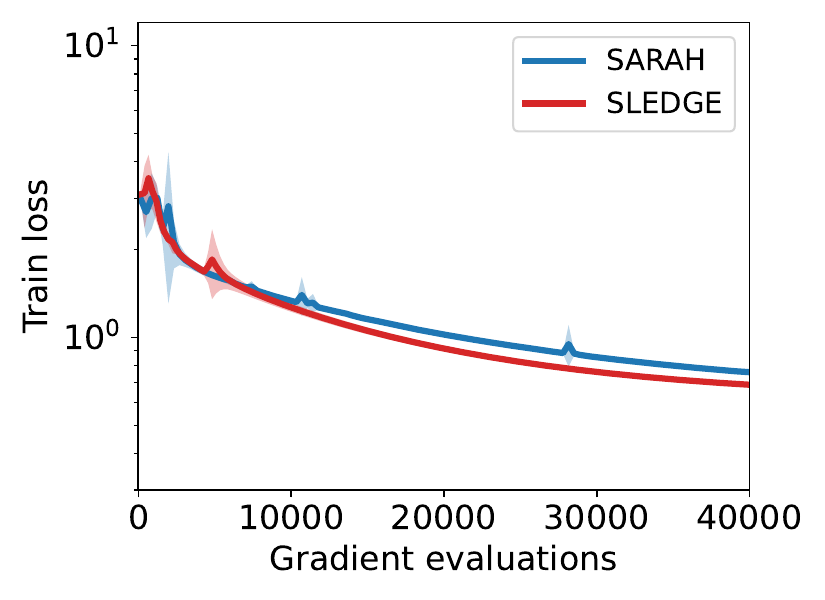}
        \vspace{-2.1mm}\subcaption{Train loss ($\eta=0.1$)}\label{fig3:1-1}\end{minipage} 
      &
      \begin{minipage}[t]{0.23\hsize}\centering
        \includegraphics[width=40mm]{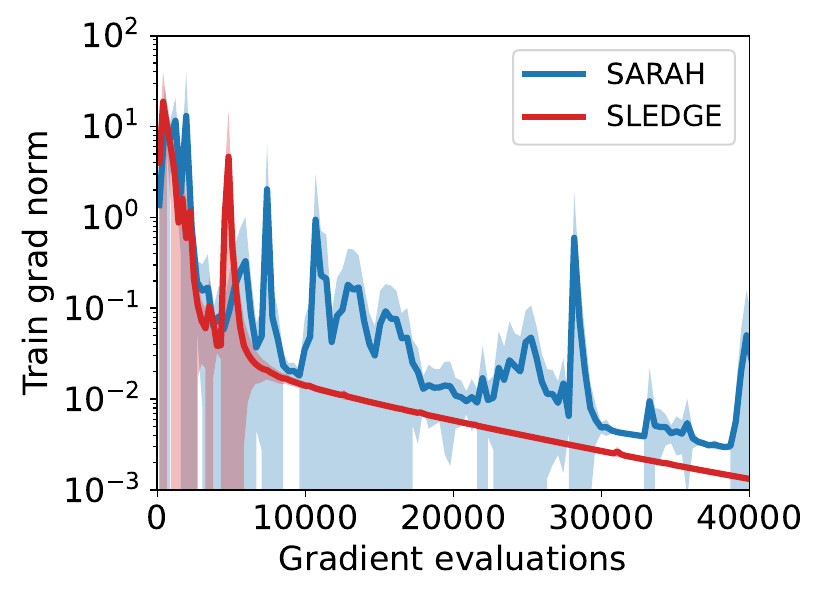}
        \vspace{-2.1mm}\subcaption{Gradient norm ($\eta=0.1$)}\label{fig3:1-2}\end{minipage} 
      &
      \begin{minipage}[t]{0.23\hsize}\centering
        \includegraphics[width=40mm]{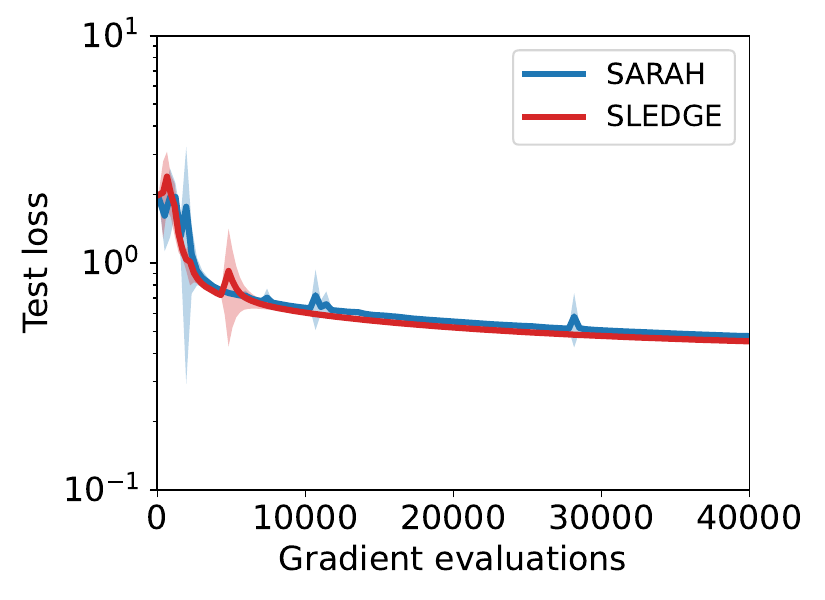}
        \vspace{-2.1mm}\subcaption{Test loss ($\eta=0.1$)}\label{fig3:1-3}\end{minipage} 
      &
      \begin{minipage}[t]{0.23\hsize}\centering
        \includegraphics[width=40mm]{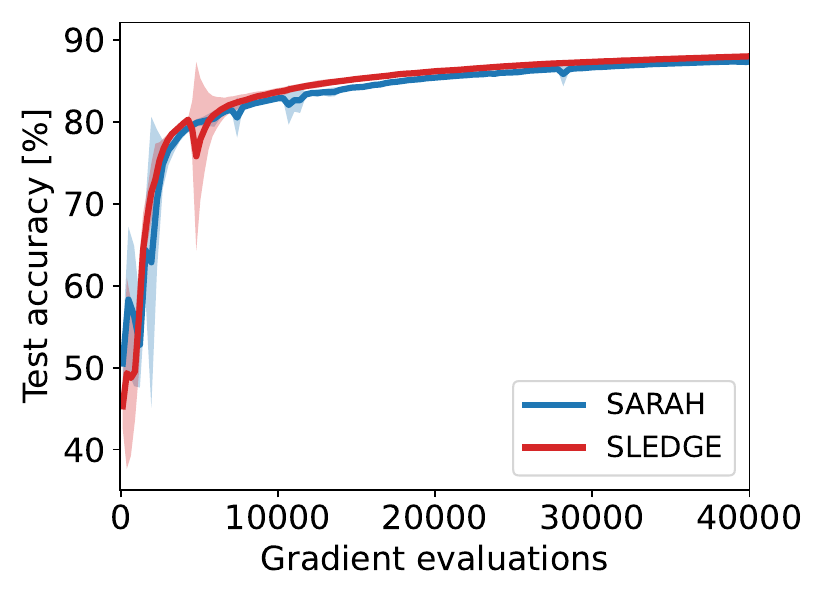}
        \vspace{-2.1mm}\subcaption{Test accuracy ($\eta=0.1$)}\label{fig3:1-4}\end{minipage} 
        \\
      \begin{minipage}[t]{0.23\hsize}\centering
        \includegraphics[width=40mm]{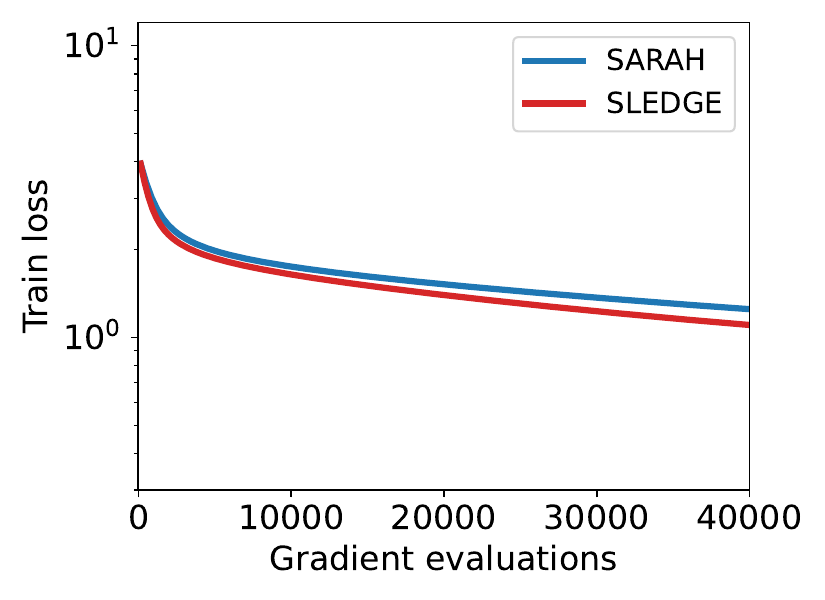}
        \vspace{-2.1mm}\subcaption{Train loss ($\eta=0.03$)}\label{fig3:2-1}\end{minipage} 
      &
      \begin{minipage}[t]{0.23\hsize}\centering
        \includegraphics[width=40mm]{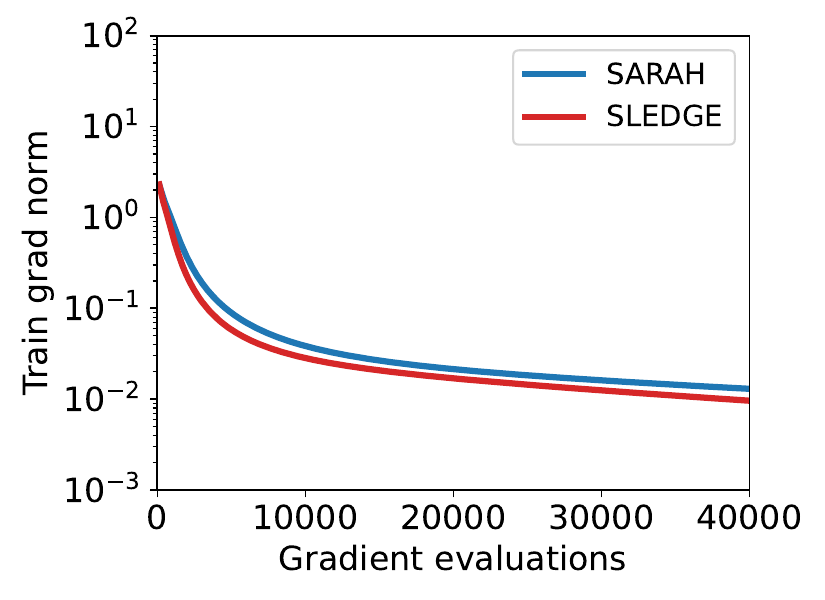}
        \vspace{-2.1mm}\subcaption{Gradient norm ($\eta=0.03$)}\label{fig3:2-2}\end{minipage} 
      &
      \begin{minipage}[t]{0.23\hsize}\centering
        \includegraphics[width=40mm]{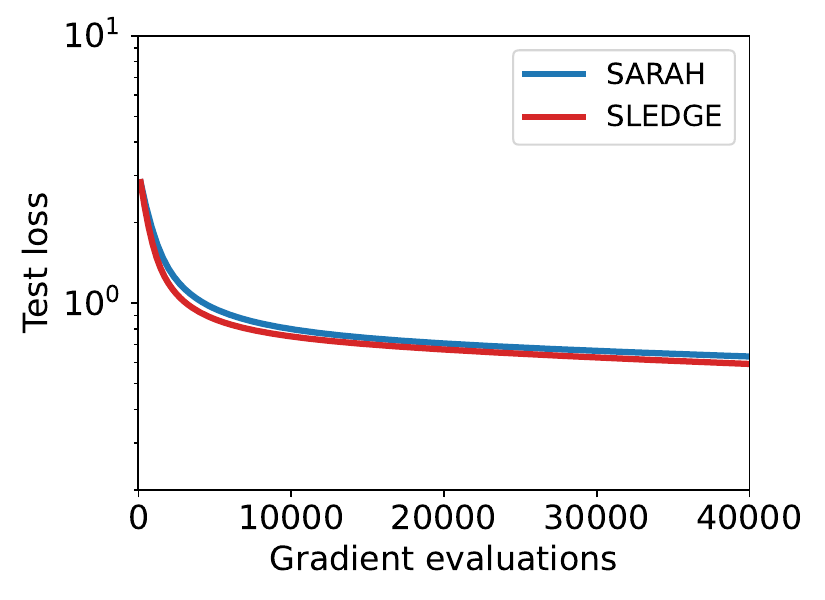}
        \vspace{-2.1mm}\subcaption{Test loss ($\eta=0.03$)}\label{fig3:2-3}\end{minipage} 
      &
     \begin{minipage}[t]{0.23\hsize}\centering
        \includegraphics[width=40mm]{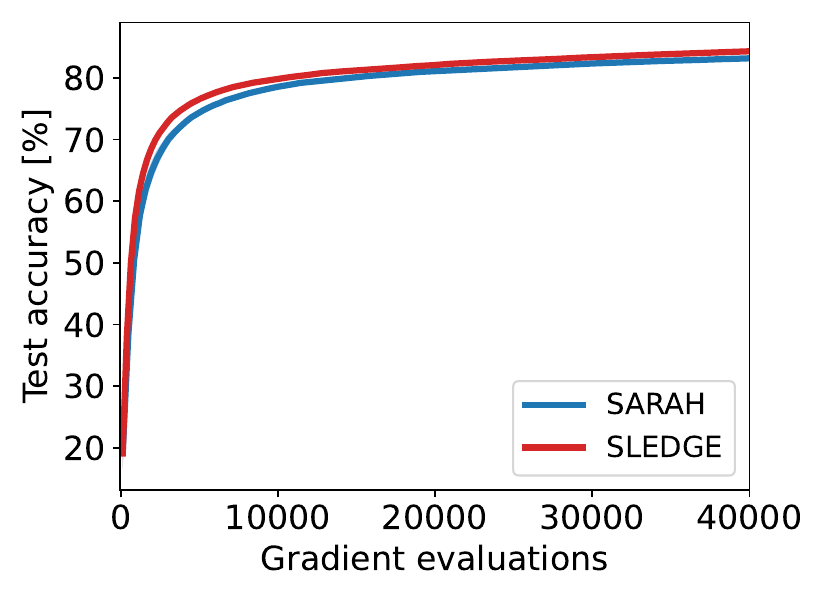}
       \vspace{-2.1mm}\subcaption{Test accuracy ($\eta=0.03$)}\label{fig3:2-4}\end{minipage} 
        \\
     \begin{minipage}[t]{0.23\hsize}\centering
        \includegraphics[width=40mm]{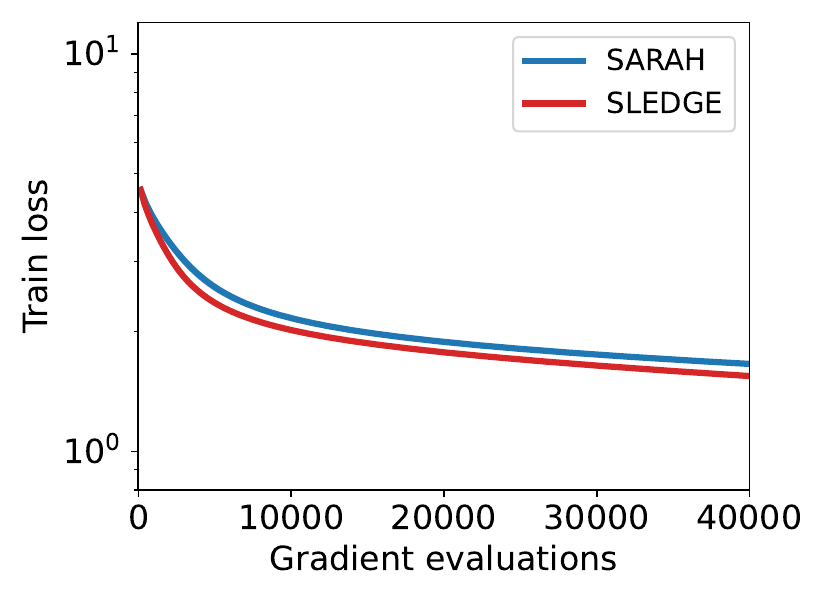}
        \vspace{-2.1mm}\subcaption{Train loss ($\eta=0.01$)}\label{fig3:3-1}\end{minipage} 
      &
     \begin{minipage}[t]{0.23\hsize}\centering
        \includegraphics[width=40mm]{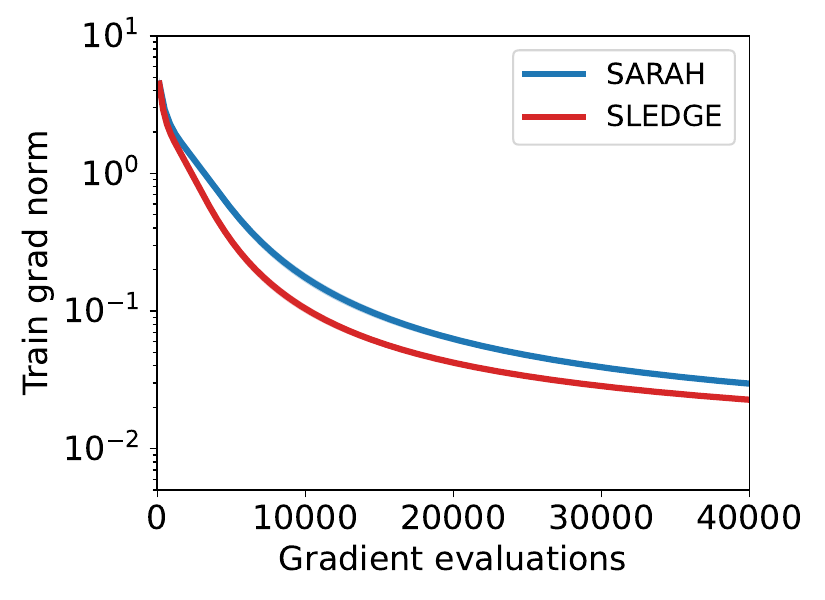}
        \vspace{-2.1mm}\subcaption{Gradient norm ($\eta=0.01$)}\label{fig3:3-2}\end{minipage} 
      &
    \begin{minipage}[t]{0.23\hsize}\centering
        \includegraphics[width=40mm]{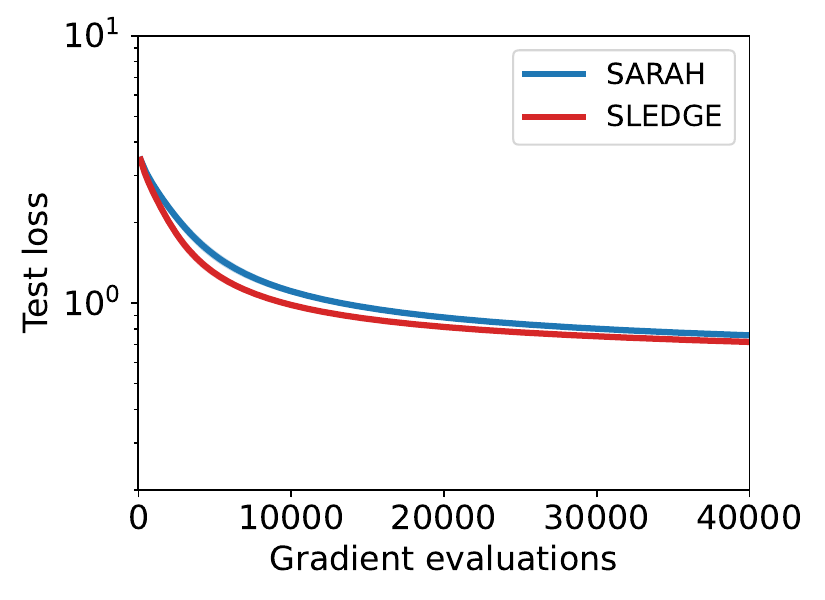}
        \vspace{-2.1mm}\subcaption{Test loss ($\eta=0.01$)}\label{fig3:3-3}\end{minipage} 
      &
    \begin{minipage}[t]{0.23\hsize}\centering
        \includegraphics[width=40mm]{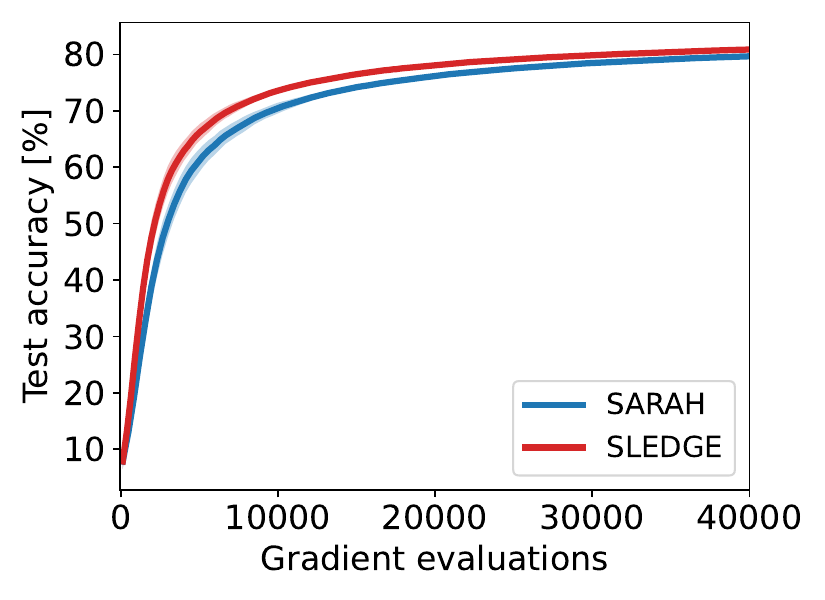}
       \vspace{-2.1mm}\subcaption{Test accuracy ($\eta=0.01$)}\label{fig3:3-4}\end{minipage} 
        \\
     \begin{minipage}[t]{0.23\hsize}\centering
        \includegraphics[width=40mm]{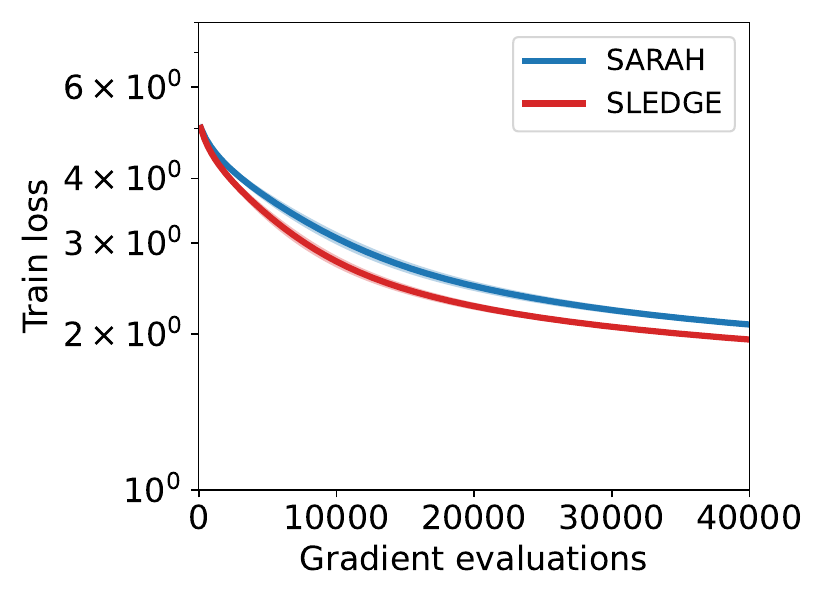}
        \vspace{-2.1mm}\subcaption{Train loss ($\eta=0.003$)}\label{fig3:4-1}\end{minipage} 
      &
     \begin{minipage}[t]{0.23\hsize}\centering
        \includegraphics[width=40mm]{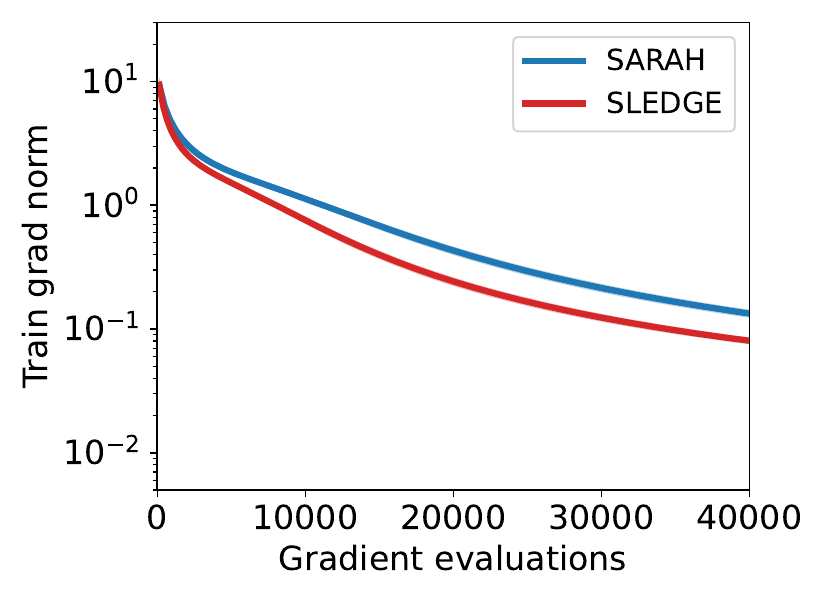}
        \vspace{-2.1mm}\subcaption{Gradient norm ($\eta=0.003$)}\label{fig3:4-2}\end{minipage} 
      &
    \begin{minipage}[t]{0.23\hsize}\centering
        \includegraphics[width=40mm]{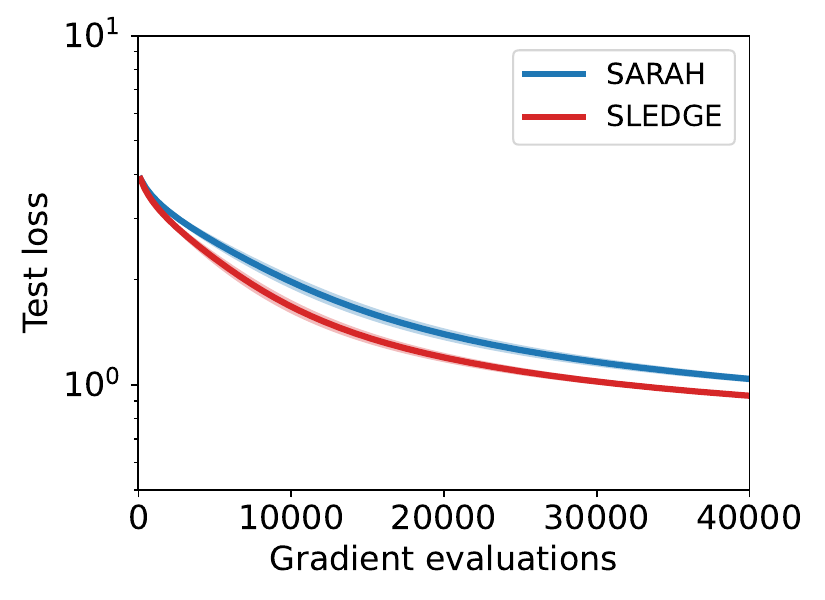}
        \vspace{-2.1mm}\subcaption{Test loss ($\eta=0.003$)}\label{fig3:4-3}\end{minipage} 
      &
    \begin{minipage}[t]{0.23\hsize}\centering
        \includegraphics[width=40mm]{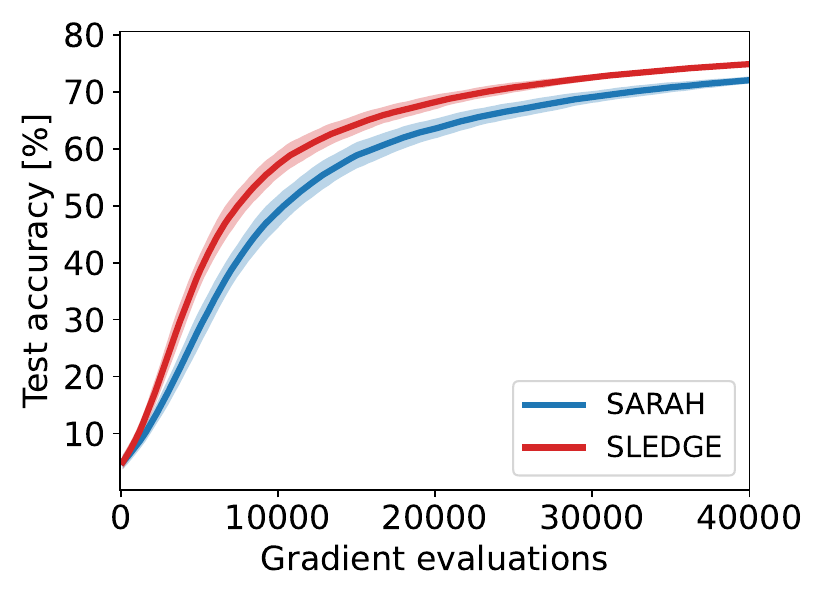}
       \vspace{-2.1mm}\subcaption{Test accuracy ($\eta=0.003$)}\label{fig3:4-4}\end{minipage} 
        \\
     \begin{minipage}[t]{0.23\hsize}\centering
        \includegraphics[width=40mm]{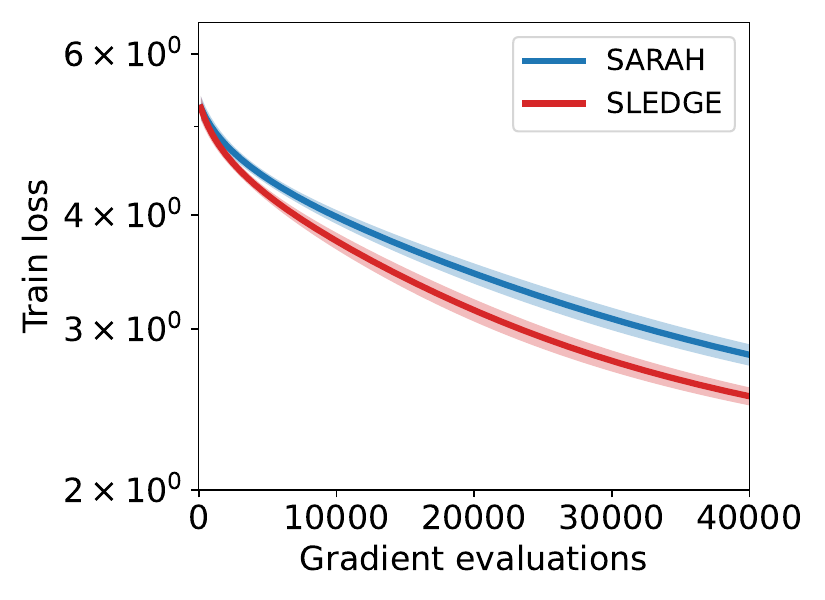}
        \vspace{-2.1mm}\subcaption{Train loss ($\eta=0.001$)}\label{fig3:5-1}\end{minipage} 
      &
     \begin{minipage}[t]{0.23\hsize}\centering
        \includegraphics[width=40mm]{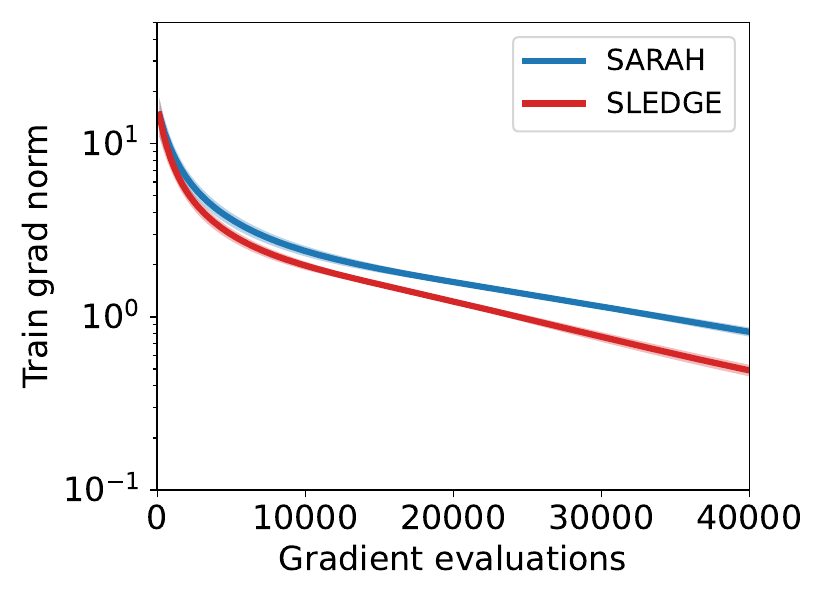}
        \vspace{-2.1mm}\subcaption{Gradient norm ($\eta=0.001$)}\label{fig3:5-2}\end{minipage} 
      &
    \begin{minipage}[t]{0.23\hsize}\centering
        \includegraphics[width=40mm]{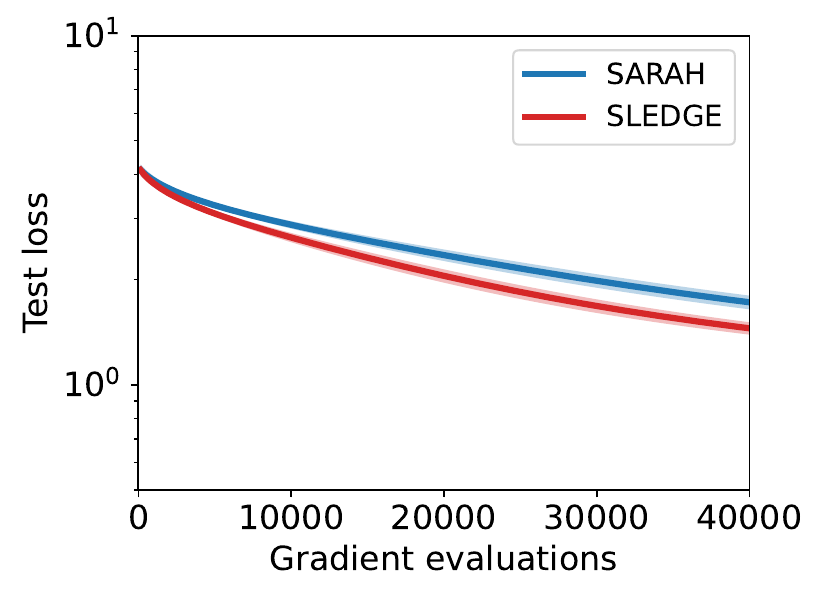}
        \vspace{-2.1mm}\subcaption{Test loss ($\eta=0.001$)}\label{fig3:5-3}\end{minipage} 
      &
    \begin{minipage}[t]{0.23\hsize}\centering
        \includegraphics[width=40mm]{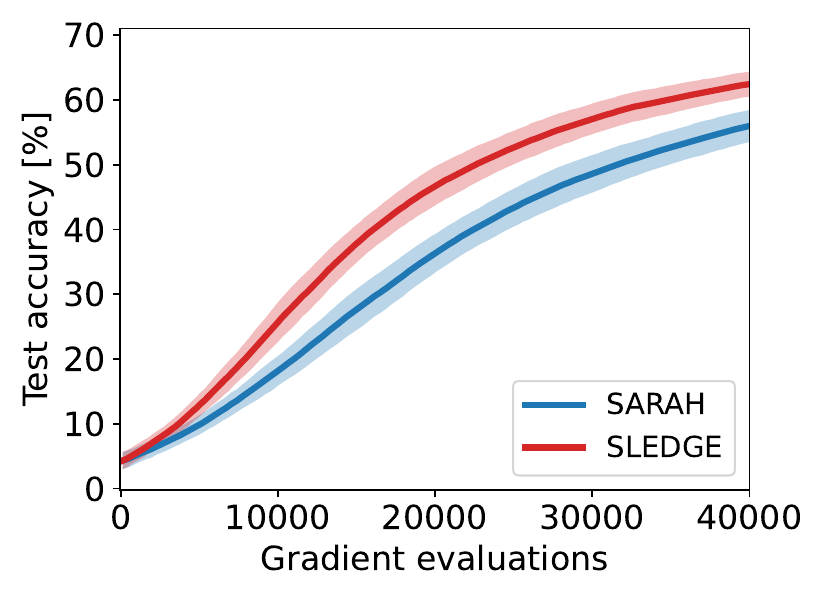}
       \vspace{-2.1mm}\subcaption{Test accuracy ($\eta=0.001$)}\label{fig3:5-4}\end{minipage} 
     \end{tabular}
     \caption{Comparison with SARAH by changing the learning rate}
     \label{fig:AppendixWithSARAH}
  \end{figure}

\Cref{fig:AppendixWithSARAH} shows the result.
We clearly observe that the proposed algorithm SLEDGE slightly faster than SARAH in all range of learning rate $\eta$.
The trajectories of SLEDGE are as stable as SARAH in all settings.
This result shows that we can remove the requirement of periodic full gradient evaluation without hurting the stability during optimization with SLEDGE.

\paragraph{Discrepancy between the gradient estimators and the true gradient}
Here we compare the norm between the gradient estimators and the true gradient because this is the most essential measure that quantify the quality of the gradient estimator.
The setting is completely the same as the previous experiment for \Cref{fig:AppendixWithSARAH}, thus $n=130$.
We compared SLEDGE estimator with SARAH and SAGA \citep{defazio2014saga,reddi2016fast}, taking the minibatch size as $b=12$ and the inner-loop length of SARAH to $m=\lfloor \frac{n}{b}\rfloor=10$.
We set the learning rate to $\eta=0.01$ for all algorithms, since the larger step size tend to increase the discrepancy, meaning that it is not fair to compare algorithms with different step sizes to discuss the discrepancy.
We plotted the mean of the five trials with different random seeds and the sample variance is also shown in the corresponding (lighter) color for each algorithm.

\begin{figure}[htbp]
        \begin{center}
        \includegraphics[width=65mm]{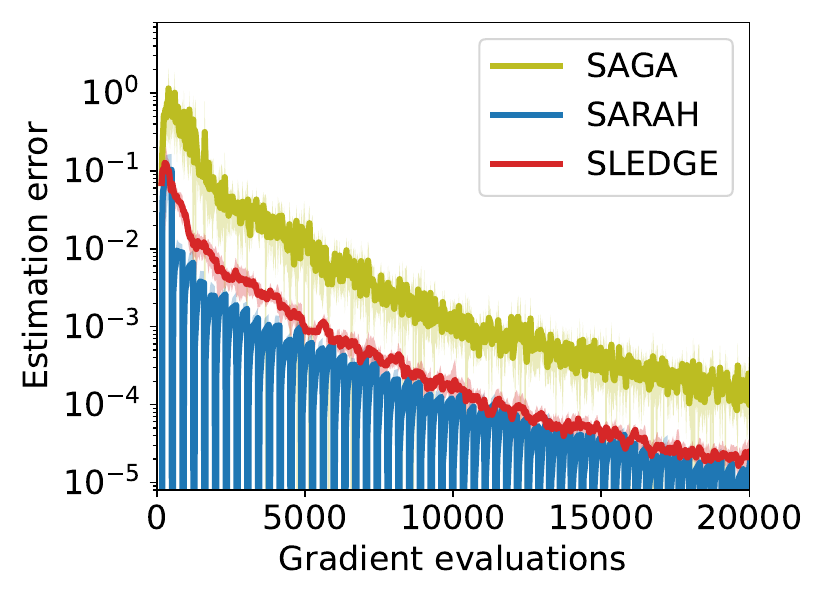}
        \end{center}
        \vspace{-0.55cm}
        \caption{Accuracy of the gradient estimators}
        \label{fig:SLEDGE_discrepancy}
\end{figure}

\Cref{fig:SLEDGE_discrepancy} shows the squared norm $\|v^t-\nabla f(x^t)\|^2$ between the gradient estimator $v^t$ of each algorithm and the true gradient $\nabla f(x^t)$ at each step $t$.
The discrepancy of the SLEDGE estimator is clearly smaller than that of SAGA, and close to that of SARAH.
Note that SARAH estimator is refreshed at every $m=10$ steps.
Remind the discussion in Subsection 3.1.
The SLEDGE estimator is designed to have as small variance as that of SARAH, while removing the need of periodic full gradient computation.
Therefore, this result validates that our strategy actually works well.

\subsection{Additional experiments for FLEDGE}

\paragraph{Escaping saddle points with FLEDGE}
\cref{theorem:D-Second-Main} guarantees second-order optimality of FLEDGE.
To validate this theoretical result, we considered the following experiment.
We first prepared a dataset with the heterogeneity parameter of $q=0.7$ (see \Cref{subsection:Appendix_A2} for details).
Then, we constructed $f_{i,j}$ with the cross-entropy loss and a three-layer fully-connedted neural network, following \citet{murata2022escaping}.
$L_2$-regularizer with a scale of $\lambda = 0.01$ is added to the empirical risk.
We compared FLEDGE with FedAvg \citep{mcmahan2017communication}, SCAFFOLD \citep{karimireddy2020scaffold}, MimeMVR \citep{karimireddy2021breaking}, BVR-L-SGD \citep{murata2021bias}, and BVR-L-PSGD \citep{murata2022escaping}.
Here, we set $P=104$, $p=10$, $b=16$, and $K=10$.
Note that, according to \cref{theorem:D-Second-Main}, setting $p\fallingdotseq \sqrt{P}$ theoretically guarantees that the convergence rate of FLEDGE is not affected by the client sampling and achieves the same number of communication complexity as that of BVR-L-PSGD to find SOSPs.
For FLEDGE and BVR-L-PSGD, we added small noise of $r=0.015$.
We plotted the mean of the five trials with different random seeds.
We omitted the sample variance for clearer presentation.

\begin{figure}[htbp]
        \begin{center}
        \includegraphics[width=150mm]{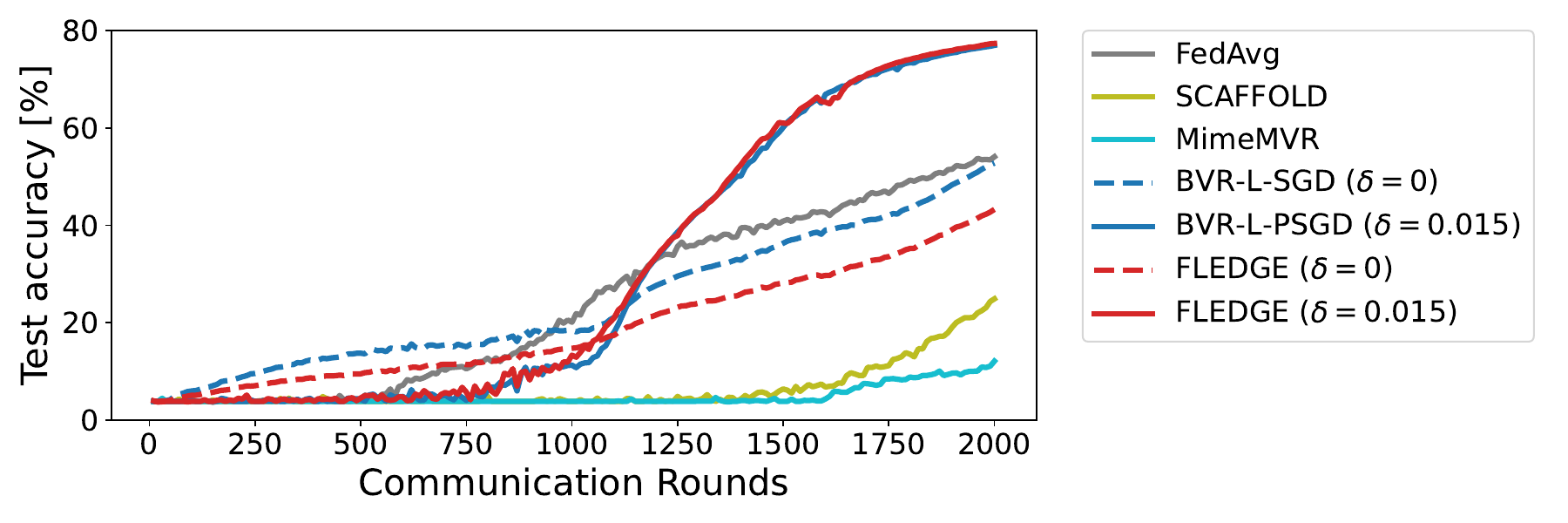}
        \end{center}
        \vspace{-0.55cm}
        \caption{Small perturbation helps faster convergence}
        \label{fig:FLEDGE_escaping_saddle_points}
\end{figure}
The result is shown in \Cref{fig:FLEDGE_escaping_saddle_points}.
We can clearly observe that FLEDGE with small noise and BVR-L-PSGD achieve the highest test accuracy.
Note that BVR-L-PSGD is almost the same as FLEDGE with no client sampling ($p=P$).
Thus, this shows that FLEDGE is not affected by the client sampling with $p\fallingdotseq\sqrt{P}$, which is consistent with the theory.
Ours is as ten times efficient as BVR-L-PSGD, in terms of communication complexity (the number of gradients communicated between the clients).

\begin{figure}[htbp]
        \begin{center}
        \includegraphics[width=85mm]{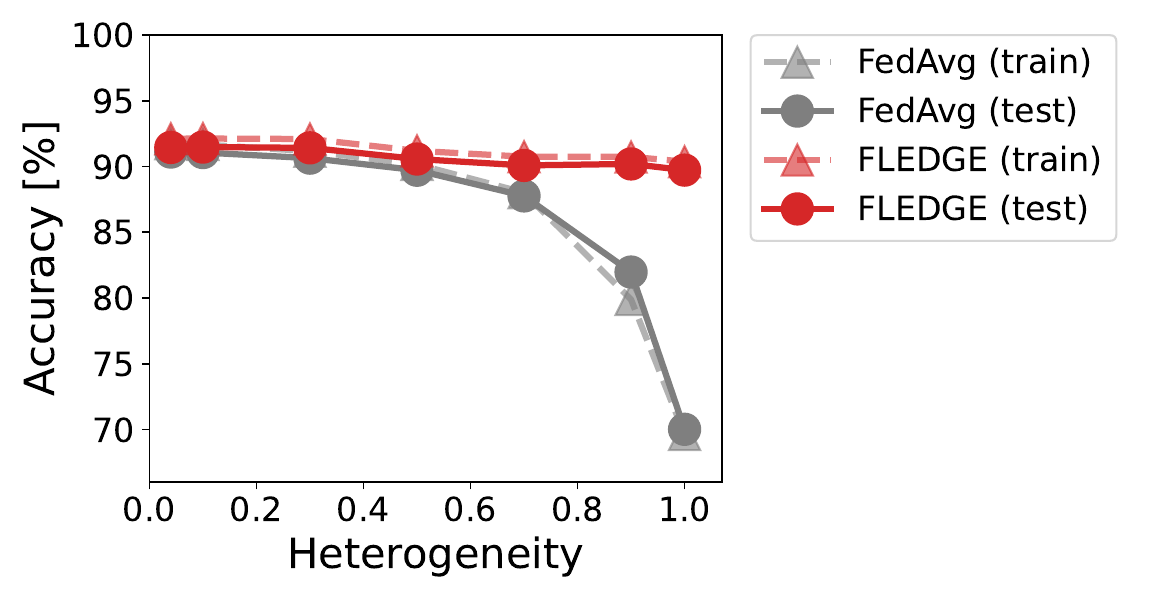}
        \end{center}
        \vspace{-0.55cm}
        \caption{Performance under changing heterogeneity}
        \label{fig:FLEDGE_compare_heterogeneity}
\end{figure}

\paragraph{Performance under changing heterogeneity}
To exhibit how correctly FLEDGE can control the variance between clients, we measured the performance of FLEDGE under changing heterogeneity.
We changed heterogeneity parameter in the range of $q\in \{0.04\text{ (i.i.d.)}, 0.1, 0.3, 0.5, 0.7, 0.9, 1.0\text{ (completely heterogeneous)}\}$, and compared FLEDGE with FedAvg, in terms of both train and test accuracy.
All other settings are the same as that of the experiment for Figure 2.
Note that we chose $p=10\fallingdotseq \sqrt{P}=\sqrt{104}$, where the theory says that the convergence is never affected by sampling of clients.
\Cref{fig:FLEDGE_compare_heterogeneity} shows the average of five trials with different random seeds.

According to \Cref{fig:FLEDGE_compare_heterogeneity}, while FedAvg decreases the train and test accuracy as the heterogeneity increases, the performance of FLEDGE with even $q=1.0$ is only slightly worse than that with $q=0.04$.
The fact that FLEDGE is little affected by the strong heterogeneity in this experiment supports our theoretical guarantee (\cref{theorem:Dist-First-Main,theorem:D-PL}) on the effect from sampling of clients.
That is, setting $p\geq \sqrt{P}$, our algorithm does not affected by sampling of clients and finds $\eps$-first-order stationary points within $\tilde{O}(\frac{1}{K\eps^2}+\frac{\zeta}{\eps^2})$ communication rounds.

\subsection{Computing infrastructures}
\begin{itemize}
    \item OS: Ubuntu 16.04.5
    \item CPU: Intel(R) Xeon(R) CPU E5-2680 v4 \@ 2.40GHz
    \item CPU Memory: 512GB
    \item GPU: Nvidia Tesla V100 (32GB)
    \item Programming language: Python 3.6.13
    \item Deep learning framework: PyTorch 1.7.1
\end{itemize}


\section{Tools}
In this section, we introduce mathematical tools we utilize in the missing proofs.
Before to do so, we weaken \cref{assumption:BoundedGradient} into the following.
\newtheorem*{assumption:BoundedGradient}{\rm\bf Assumption~\ref{assumption:BoundedGradient}'}
\begin{assumption:BoundedGradient}{assumption}[Boundedness of Gradient (formal)]
{\rm(i)}
It holds that $\mathbb{E}_i[\|\nabla f_i(x^0)-\nabla f(x^0)\|^2]\leq \sigma_c^2$, where the expectation $\mathbb{E}_i$ is taken over the choice of $i$.
Moreover, $\|\nabla f_i(x^0)-\nabla f(x^0)\|\leq G_c$ holds for all $i$.
{\rm(ii)} 
For all $i$ and $x$, assume $\mathbb{E}_j[\|\nabla f_{i,j}(x)-\nabla f_{i}(x)\|^2]\leq \sigma^2$, here the expectation $\mathbb{E}_{j}$ is taken about the choice of $j$.
For all $i,j$ and $x$, $\|\nabla f_{i,j}(x)-\nabla f_{i}(x)\|^2\leq G^2$.
\end{assumption:BoundedGradient}
$\sigma_c$ and $\sigma$ in the previous definition are now represented by $G_c$ and $G$.
$\sigma_c$ and $\sigma$ in this definition can be bounded by the previous $\sigma_c$ and $\sigma$.
Thus, we can give tighter evaluation on the dependency on the gradient boundedness with this definition.

\subsection{Concentration Inequalities}
Here, we prepare concentration inequalities for later use.
We first present Bernstein-type bounds.
\begin{proposition}[Matrix Bernstein inequality \cite{tropp2012user}]
Let $X_1,\cdots,X_k$ be a finite sequence of independent random matrices with dimension $d_1\times d_2$.
Assume each random matrix satisfies
\begin{align}
    \mathbb{E}[X_i] = 0 \quad\text{and}\quad \|X_i\|\leq R \quad \text{almost surely.}
\end{align}
Define
\begin{align}
    \sigma^2 = \max\left\{
        \left\|\sum_{i}\mathbb{E}[X_iX_i^\top]\right\|
        , \left\|\sum_{i}\mathbb{E}[X_i^\top X_i]\right\|
    \right\}.
\end{align}
Then, for all $t\geq 0$,
\begin{align}
    \mathbb{P}\left[ \left\|\sum_{i} X_i\right\| \geq t\right]
    \leq (d_1+d_2)\cdot \exp\left(
        \frac{-t^2/2}{\sigma^2 +Rt/3}
    \right).
\end{align}
\end{proposition}
In this paper, we deal only with the vector case. In that case, the inequality is rewritten for bounds with high probability, as follows.
\begin{proposition}[Vector Bernstein inequality]\label{proposition:Bernstein}
Let $x_1,\cdots,x_k$ be a finite sequence of independent, random, $d$-dimensional vectors and $\nu\in (0,1)$. 
Assume that each vector satisfies
\begin{align}
    \|x_i - \mathbb{E}[x_i]\|\leq R \quad \text{almost surely.}
\end{align}
Define
\begin{align}
    \sigma^2 = \sum_{i=1}^k \mathbb{E}[\left\|x_i - \mathbb{E}[x_i]\right\|^2]
\end{align}
Then, with probability at least $1-\nu/{\rm poly}(n,P,T,K)$,
\begin{align}
   \left\|\sum_{i=1}^k (x_i - \mathbb{E}[x_i])\right\|^2\leq C_1^2 \cdot (\sigma^2 + R^2)
\end{align}
where $C_1 = O\left(\log (\nu^{-1}+n+P+d+T+K)\right) = \tilde{O}(1)$.
\end{proposition}
\begin{remark}
Here we do not specify ${\rm poly}(n,P,T,K)$ to apply different polynomials to later.
Whenever we use this inequality with different ${\rm poly}(n,m,T)$, we will reuse $C_1$ for the notational simplicity.
We also use this constant $C_1$ in the following parts to denote constants as large as $O(\log (\nu^{-1}+n+P+d+T+K))$, with a slight abuse of notations.
\end{remark}
Moreover, a similar inequality holds when we consider sampling without replacement.
To the best of our knowledge, Bernstein inequality without replacement for vectors has not been rigorously proven and we attach its complete proof at the end of this subsection.
\begin{proposition}[Vector Bernstein inequality without replacement]\label{proposition:BernsteinNoReplacement}
Let $A=(a_1,a_2,\cdots,a_k)$ be $d$-dimensional fixed vectors, $X=(x_1,\cdots,x_l)$ $(l\leq k)$ be a random sample without replacement from $A$. 
Assume that $\sum_{i=1}^k a_i =0$ and that each vector satisfies
\begin{align}
    \|a_i\|\leq R.
\end{align}
Define
\begin{align}
    \sigma^2 = \frac1k\sum_{i=1}^k \|a_i\|^2.
\end{align}
Then, for each $t\geq 0$ and $l<k$,
\begin{align}
    \mathbb{P}\left[\left\|\sum_{i=1}^l x_i\right\|\geq t\right]
    \leq (d+1)\cdot \exp\left(
        \frac{-t^2}{2l\sigma^2+Rt/3}
    \right)
    .
\end{align}
Moreover, for each $l<k$, with probability at least $1-\nu/{\rm poly}(n,P,T,K)$, 
\begin{align}
   \left\|\sum_{i=1}^l x_i \right\|^2\leq C_1^2 \cdot
   (l\sigma^2+R^2),
\end{align}
where $C_1 = O\left(\log (n+P+d+T+K)\right) = \tilde{O}(1)$.
\end{proposition}

Finally, we need a high-probability version of Azuma-Hoeffding inequality.
\begin{proposition}[Azuma-Hoeffding inequality with high probability \cite{chung2006concentration,tao2015random}]\label{proposition:Azuma}
Let $\{x_i\}$ be a $d$-dimensional vector sequence and martingale with respect to a filtration $\{\mathcal{F}_i\}$.
Assume that each $x_i$ satisfies $\mathbb{E}[x_i|\mathcal{F}_{i-1}]=0$ and
\begin{align}
    \|x_i\|\leq R_i \quad \text{with probability $1-\nu_i$}
\end{align}
for $\nu_i\in(0,1)$ $(i=1,\dots,k)$. 
Then, with probability at least $1 - \nu/{\rm poly}(n,P,T,K) - \sum_{i=1}^k \nu_i$, 
\begin{align}
   \left\|\sum_{i=1}^k x_i\right\|^2\leq C_1^2 \sum_{i=1}^k R_i^2,
\end{align}
where $C_1 = O\left(\log (\nu^{-1}+n+P+d+T+K)\right) = \tilde{O}(1)$.
\end{proposition}

\subsection{Proof of \cref{proposition:BernsteinNoReplacement}}

In order to show \cref{proposition:BernsteinNoReplacement}, we use the Martingale counterpart of Bernstein's Inequality for random matrix. The following is a slightly weaker version of \citet{tropp2011freedman}.
\begin{proposition}[Freedman's inequality for matrix martingales]\label{proposition:MatrixFreedman}
Consider a matrix martingale $\{Y_i\mid\ i=0,1,\cdots \}$ with respect to a filtration $\{\mathcal{F}_i\}$, whose values are matrices with dimension $d_1\times d_2$, and let $\{X_i\mid\ i=1,2,\cdots \}$ be the difference sequence. Assume that each of the difference sequence is uniformly bounded:
\begin{align}
    \|X_i\|^2\leq {R'}^2 \quad \text{almost surely.}
\end{align}
Also, assume that each $i$ satisfies 
\begin{align}
     \max\left\{
        \left\|\mathbb{E}[X_iX_i^\top|\ \mathcal{F}_{i-1}]\right\|
        , \left\|\mathbb{E}[X_i^\top X_i|\ \mathcal{F}_{i-1}]\right\|
    \right\} \leq {\sigma'}^2 \quad \text{almost surely.}
\end{align}
Then, for all $t\geq 0$ and for each $l$,
\begin{align}
    \mathbb{P}\left[ \left\|Y_l\right\| \geq t\right]
    \leq (d_1+d_2) \cdot \exp\left(
        \frac{-t^2/2}{l{\sigma'}^2 +R't/3}
    \right).
\end{align}
\end{proposition}

\begin{proof}[Proof of \cref{proposition:BernsteinNoReplacement}]
First, we consider the case $l\leq \frac{k}{2}$.
Let $y_i = \sum_{j=1}^i x_j$ and consider a filtration $\mathcal{F}_i=\sigma(x_1,\cdots,x_i)$. 
Then, we have
\begin{align}
    \mathbb{E}\left[y_{i+1}\left|\mathcal{F}_{i}\right.\right]
    =y_{i} + \frac{1}{k-i}\left(\sum_{j=1}^n a_j - \sum_{j=1}^i x_j\right)
    =\frac{k-i-1}{k-i}y_{i}
    .
\end{align}
This means that $\left\{\frac{1}{k-i}y_i\right\}_{i=0}^l$ is martingale with respect to $\{\mathcal{F}_i\}$.
We have that this martingale satisfies the assumptions of \cref{proposition:MatrixFreedman} with ${R'}^2 = \frac{R^2}{(k-l)^2}$ and ${\sigma'}^2 =\frac{2\sigma^2}{(k-l)^2}$.
In fact, we have
\begin{align}
    \left\|\frac{1}{k-i-1}y_{i+1} - \mathbb{E}\left[\left.\frac{1}{k-i-1}y_{i+1}\right|\mathcal{F}_{i}\right]\right\|^2
    &= \left\|\frac{1}{k-i-1}x_{i+1} - \mathbb{E}\left[\left.\frac{1}{k-i-1}x_{i+1}\right|\mathcal{F}_{i}\right]\right\|^2
    \\ & \leq \left\|\frac{1}{k-i-1}x_{i+1}\right\|^2
    \leq 
    \frac{R^2}{(k-i-1)^2}
    \leq 
    \frac{R^2}{(k-l)^2},
\end{align}
where the equality follows since $x_1\dots,x_i$ are $\mathcal{F}_i$-measurable, and 
\begin{align}
    \mathbb{E}\left[\left.\left\|\frac{1}{k-i-1}y_{i+1} - \mathbb{E}\left[\left.\frac{1}{k-i-1}y_{i+1}\right|\mathcal{F}_{i}\right]\right\|^2\right|\mathcal{F}_{i}\right]
    & \leq \mathbb{E}\left[\left.\left\|\frac{1}{k-i-1}x_{i+1}\right\|^2\right|\mathcal{F}_{i}\right]
    \\ & = \frac{1}{(k-i-1)^2} \cdot \frac{1}{k-i}\left(\sum_{j=1}^k\|a_j\|^2 - \sum_{j=1}^i\|x_j\|^2\right)
    \\ & \leq \frac{1}{(k-l)^2} \cdot \frac2k \sum_{i=1}^k\|a_i\|^2
    \quad \left(\because k-i \geq \frac{k}{2}\right)
    \\ & =\frac{2\sigma^2}{(k-l)^2}.
\end{align}
Thus, we use \cref{proposition:MatrixFreedman} to obtain
\begin{align}
    \mathbb{P}\left[\left\|y_l\right\|\geq t\right]
    \leq (d+1)\cdot \exp\left(
        \frac{-t^2}{2l\sigma^2+Rt/3}
    \right).
\end{align}
What remains is the case of $l \geq \frac{k}{2}$.
Since $\sum_{i=1}^l x_i = - \sum_{i=l+1}^k x_i$ holds, we can apply the above bound for $ \sum_{i=l+1}^k x_i$.
Thus, we have the first assertion for all $l<k$.
The second assertion follows by setting $t=O\left((l\sigma^2+R)\log (\nu^{-1}+n+m+d+T)\right)$ $=C_1\cdot (l\sigma^2+R)$.

\end{proof}

\subsection{Linear Algebraic Tool}
The following lemma is due to \citet{murata2022escaping}. 
 We provide its proof here. 
\begin{lemma}[\citet{murata2022escaping}]\label{lemma:Murata-MatrixEigenvalue}
Let $A$ be a $d\times d$ symmetric matrix with the smallest and largest eigenvalues $\lambda_{{\rm min}}<0$ and $\lambda_{{\rm max}}<1$, respectively.
Then, for $k=0,1,\cdots$, it holds that
\begin{align}
    \|A(I-A)^k\| \leq -\lambda_{{\rm min}}(1-\lambda_{{\rm min}})^k + \frac{1}{k+1}.
    \end{align}
\end{lemma}
\begin{proof}
Since $A$ is diagonalizable, we write $A=\sum_{i=1}^d \lambda_i e_ie_i^\top$, where $e_1,\dots,e_d$ are normalized eigenvectors and $\lambda_{{\rm min}}=\lambda_1\leq \cdots\leq \lambda_d = \lambda_{{\rm max}}$ are the corresponding eigenvalues.
Then, it holds that 
\begin{align}
    A(I-A)^k = \sum_{i=1}^d \lambda_i (1-\lambda_i)^k e_ie_i^\top
    .
\end{align}
Thus, the remaining is to evaluate $\max_{i}|
\lambda_i (1-\lambda_i)^k|$.
After some algebra, we get
\begin{align}
    0<\lambda (1-\lambda)^k &\leq 
    \begin{cases}
    -\lambda (1-\lambda)^k & \left(\text{if $\lambda\leq 0$}\right)\\
     \frac{1}{k+1} \left(\frac{k}{k+1}\right)^k & \left(\text{if $\lambda> 0$; the equality holds with $\lambda = \frac{1}{1+k}$}\right)
    \end{cases}
    \\ &\leq -\lambda_{{\rm min}}(1-\lambda_{{\rm min}})^k + \frac{1}{k+1},
\end{align}
which concludes the proof.
\end{proof}

\section{Missing Proofs for \AlgMain}
This section provides the missing proofs in \Cref{section:SLEDGE} about the convergence property of \AlgMain.

\subsection{Finding First-order Stationary Points (Proof of \cref{theorem:First-Main})}
In this subsection, we show that {\AlgMain}
finds first-order stationary points with high probability (\cref{theorem:First-Main}). 
For the proof of \cref{theorem:First-Main}, we use the following classical argument (e.g. \citet{ge2019stabilized,li2019ssrgd,li2021page}), which ensures decrease of the function values.

\begin{lemma}[]\label{lemma:DescentLemma}
Let $f$ be an $L$-gradient Lipschitz function and $x^t:=x^{t-1}-\eta v^{t-1}+\xi^{t-1}$ with $\|\xi^{t-1}\|\leq r$.
Then,
\begin{align}
    f(x^t) \leq f(x^{t-1}) + \eta \|\nabla f(x^{t-1}) - v^{t-1}\|^2
    - \frac{\eta}{2}\|\nabla f(x^{t-1})\|^2
    - \left(\frac{1}{2\eta}-\frac{L}{2}\right)\|x^t-x^{t-1}\|^2
    + \frac{r^2}{\eta}
\end{align}
holds.
\end{lemma}
\begin{proof}
Starting from the direct result from $L$-gradient Lipschitzness, we have
\begin{align}
    f(x^t)
    & \leq 
    f(x^{t-1}) + \langle \nabla f(x^{t-1}), x^t-x^{t-1}\rangle + \frac{L}{2}\|x^t-x^{t-1}\|^2
    \\ & =    
    f(x^{t-1}) + \left\langle \nabla f(x^{t-1}) - v^{t-1} + \frac{\xi^{t-1}}{\eta}, x^t-x^{t-1}\right\rangle +
    \left\langle  v^{t-1} - \frac{\xi^{t-1}}{\eta}, x^t-x^{t-1}\right\rangle
    + \frac{L}{2}\|x^t-x^{t-1}\|^2
    \\ &  =      
    f(x^{t-1}) + \left\langle \nabla f(x^{t-1}) - v^{t-1} + \frac{\xi^{t-1}}{\eta}, x^t-x^{t-1}\right\rangle 
    - \left(\frac{1}{\eta}-\frac{L}{2}\right)\|x^t-x^{t-1}\|^2
    \\ &\label{eq:First-FunctionDecrease-1} = 
    f(x^{t-1}) + \frac{\eta}{2} \left\| \nabla f(x^{t-1}) - v^{t-1} + \frac{\xi^{t-1}}{\eta}\right\|^2
    - \frac{\eta}{2}\|\nabla f(x^{t-1})\|^2
    - \left(\frac{1}{2\eta}-\frac{L}{2}\right)\|x^t-x^{t-1}\|^2
    \\ & \label{eq:First-FunctionDecrease-2} \leq
    f(x^{t-1}) + \eta \|\nabla f(x^{t-1}) - v^{t-1}\|^2
    - \frac{\eta}{2}\|\nabla f(x^{t-1})\|^2
    - \left(\frac{1}{2\eta}-\frac{L}{2}\right)\|x^t-x^{t-1}\|^2
    + \eta \left\|\frac{\xi^{t-1}}{\eta}\right\|^2
     \\ & \label{eq:First-FunctionDecrease-3} \leq
    f(x^{t-1}) + \eta \|\nabla f(x^{t-1}) - v^{t-1}\|^2
    - \frac{\eta}{2}\|\nabla f(x^{t-1})\|^2
    - \left(\frac{1}{2\eta}-\frac{L}{2}\right)\|x^t-x^{t-1}\|^2
    + \frac{r^2}{\eta},
\end{align}
where we used $x^t-x^{t-1}=\eta v^{t-1}+\xi^{t-1}$ and $\langle a-b,b\rangle=\frac12 (\|a-b\|^2-\|a\|^2+\|b\|^2)$ for \eqref{eq:First-FunctionDecrease-1}, $\|a+b\|^2\leq 2(\|a\|^2+\|b\|^2)$ for \eqref{eq:First-FunctionDecrease-2}, and $\|\xi^{t-1}\|\leq r$ for \eqref{eq:First-FunctionDecrease-3}.
\end{proof}
Our algorithm uses $v^t = \frac1n\sum_{i=1}^n y^t_i$ as an estimator of $\nabla f(x^t)$. 
To apply \cref{lemma:DescentLemma} for our algorithm, we need to evaluate the term $\|v^t - \nabla f(x^{t})\|^2$, the variance of the gradient estimator.
The next lemma provides its upper bound that holds with high probability.
\setcounter{theorem}{0}
\begin{lemma}
Let $v^t = \frac1n\sum_{i=1}^n y^t_i$ and all the other variables be as stated in Algorithm 1.
Then, by taking $T_1 = \frac{n}{b}C_1$,
\begin{align}
    \|v^t - \nabla f(x^t)\|^2
    \leq
    \begin{cases}
    \displaystyle
        \frac{15C_1^8\zeta^2}{b}\sum_{s =\max\{1,t-T_1+1\}}^{t}\|x^s -x^{s -1}\|^2
        +
        \frac{12C_1^2\mathbbm{1}[t< T_1]}{b}\cdot \left(\sigma_c^2 + \frac{G_c^2}{b}\right)
    &\quad (\text{Option I})\\\displaystyle
    \frac{15C_1^8\zeta^2}{b}\sum_{s =\max\{1,t-T_1+1\}}^{t}\|x^s -x^{s -1}\|^2
&\quad (\text{Option II})
    \end{cases}
\end{align}
holds for all $t=1,\cdots,T$ with probability at least $1-3\nu$.
\end{lemma}
\setcounter{theorem}{16}
We decompose $\|v^t - \nabla f(x^t)\|$ into three parts to each of which one of the following lemmas is applied.
Below, for each $1\leq s\leq t$, we let $\tilde{I}_s^t = [n]\setminus \bigcup_{\tau = s}^t I^t$, which is a set of indexes that are not selected between $s+1$ and $t$.

\begin{lemma}\label{lemma:First-Auxiliary-A}
The following holds uniformly for all $1\leq t \leq T$ with probability at least $1-\nu$:
\begin{align}
    \begin{split}
    \left\|
    \frac{1}{n} \sum_{s=\max\{1,t-T_1+1\}}^t \sum_{i\in \tilde{I}_s^t}(\nabla f_i(x^{s})-\nabla f_i(x^{s-1}) - (\nabla f(x^{s})-\nabla f(x^{s-1})))\right\|^2 
    \\ 
    \leq 
    \frac{C_1^3\zeta^2}{b}\sum_{s=\max\{1,t-T_1+1\}}^t \|x^s-x^{s-1}\|^2
    .
    \end{split}
\end{align}
\end{lemma}
\begin{lemma}\label{lemma:First-Auxiliary-B}
The following inequality holds uniformly for all $1\leq t \leq T$ with probability at least $1-\nu$:
\begin{align}
    \begin{split}
    \left\|\frac{1}{n}\sum_{s=\max\{1,t-T_1+1\}}^t\frac{|\tilde{I}_s^t|}{b}\sum_{i\in I^s} (\nabla f_i(x^{s})-\nabla f_i(x^{s-1})- (\nabla f(x^{s})-\nabla f(x^{s-1})))\right\|^2
    \\ \leq \label{eq:First-Variance-9}
    \frac{4C_1^8\zeta^2}{b}\sum_{s=\max\{1,t-T_1+1\}}^t\|x^s-x^{s-1}\|^2
    .
    \end{split}
\end{align}
\end{lemma}
\begin{lemma}\label{lemma:First-Auxiliary-C}
The following inequality holds with probability at least $1-\nu$:
\begin{align}
    \begin{split}
        \left\|\frac{1}{n}\sum_{i\in \tilde{I}_1^t}(y^0_i-\nabla f_i(x^0))
        \right\|^2
        \leq
        \begin{cases}
            \frac{4C_1^2}{b}\left(\sigma_c^2+\frac{G_c^2}{b}\right) & (\text{Option I, with probability $1-\nu$ uniformly over $1\leq t \leq T$})
        \\
        0& (\text{Option II})
        \end{cases}
            .
    \end{split}
\end{align}
\end{lemma}

\begin{proof}[Proof of \cref{lemma:First-Auxiliary-A}]
First, we have that 
\begin{align}
    ({\rm a})& \coloneqq \left\|
    \frac{1}{n} \sum_{s=\max\{1,t-T_1+1\}}^t \sum_{i\in \tilde{I}_s^t}(\nabla f_i(x^{s})-\nabla f_i(x^{s-1}) - (\nabla f(x^{s})-\nabla f(x^{s-1})))\right\|^2     
    \\ &  \leq 
    \frac{t-\max\{1,t-T_1+1\}+1}{n^2}\sum_{s=\max\{1,t-T_1+1\}}^t
    \left\|
     \sum_{i\in \tilde{I}_s^t}(\nabla f_i(x^{s})-\nabla f_i(x^{s-1}) - (\nabla f(x^{s})-\nabla f(x^{s-1})))\right\|^2 
     \\ & \label{eq:First-Variance-3} \leq 
    \frac{T_1}{n^2}\sum_{s=\max\{1,t-T_1+1\}}^t
    \left\|
     \sum_{i\in \tilde{I}_s^t}(\nabla f_i(x^{s})-\nabla f_i(x^{s-1}) - (\nabla f(x^{s})-\nabla f(x^{s-1})))\right\|^2,
\end{align}
where we use $\|\sum_{i=1}^ma_i\|^2\le m\sum_{i=1}^m\|a_i\|^2$ for the first inequality. 
For each $s$, from \cref{assumption:Heterogeneity}, 
\begin{align}
    \|\nabla f_i(x^{s})-\nabla f_i(x^{s-1}) - (\nabla f(x^{s})-\nabla f(x^{s-1}))\|
    \leq
    \zeta \|x^s-x^{s-1}\|
\end{align}
holds for all $i\in [n]$. 
By vector Bernstein inequality without replacement (\cref{proposition:BernsteinNoReplacement}), for each $t\geq 1$ and $s$ satisfying $\max\{1,t-T_1+1\}\le s\le t$, we have that
\begin{align}
\label{eq:First-Variance-4}
    \left\|
     \sum_{i\in \tilde{I}_s^t}(\nabla f_i(x^{s})-\nabla f_i(x^{s-1}) - (\nabla f(x^{s})-\nabla f(x^{s-1})))\right\|^2
     \leq C_1^2|\tilde{I}_s^t|\zeta^2\|x^s-x^{s-1}\|^2
\end{align}
holds with probability at least $1-\frac{\nu}{T^2}$.
Thus, in \eqref{eq:First-Variance-3}, \eqref{eq:First-Variance-4} holds uniformly for all $t$ and $s$ with probability at least $1-\nu$.
Applying this bound to \eqref{eq:First-Variance-3} yields
\begin{align}
    {\rm (a)} \leq 
    \frac{T_1}{n^2}\sum_{s=\max\{1,t-T_1+1\}}^t
    C_1^2|\tilde{I}_s^t|\zeta^2\|x^s-x^{s-1}\|^2 \leq
    \frac{C_1^3\zeta^2}{b}\sum_{s=\max\{1,t-T_1+1\}}^t
    \|x^s-x^{s-1}\|^2,
\end{align}
where the second inequality follows from $|\tilde{I}_s^t|\leq n$ and $T_1 = \frac{n}{b}C_1$.
\end{proof}

\begin{proof}[Proof of \cref{lemma:First-Auxiliary-B}]
Since $|\tilde{I}_s^t|$ depends not only on $I^s$ but also on $I^s,I^{s+1},\cdots,I^t$, the left-hand side of \eqref{eq:First-Variance-9} is not a sum of martingale variables with respect to the filtration 
$\{\sigma(I^1,\cdots,I^s)\}_{s=1}^t$. Thus, we consider $\mathbb{E}[|\tilde{I}_s^t|]$ instead of $|\tilde{I}_s^t|$ and validate the difference between them later.
We decompose \eqref{eq:First-Variance-9} as
\begin{align}
&
    \left\|\frac{1}{n}\sum_{s=\max\{1,t-T_1+1\}}^t\frac{|\tilde{I}_s^t|}{b}\sum_{i\in I^s} (\nabla f_i(x^{s})-\nabla f_i(x^{s-1})- (\nabla f(x^{s})-\nabla f(x^{s-1})))\right\|^2
    \\ & \leq 
    2\left\|\frac{1}{n}\sum_{s=\max\{1,t-T_1+1\}}^t\frac{\mathbb{E}[|\tilde{I}_s^t|]}{b}\sum_{i\in I^s} (\nabla f_i(x^{s})-\nabla f_i(x^{s-1})- (\nabla f(x^{s})+\nabla f(x^{s-1})))\right\|^2
    \\ & \hspace{5mm} +
    2\left\|\frac{1}{n}\sum_{s=\max\{1,t-T_1+1\}}^t\left(|\tilde{I}_s^t|-\mathbb{E}[|\tilde{I}_s^t|]\right)^2\frac1b\sum_{i\in I^s} (\nabla f_i(x^{s})-\nabla f_i(x^{s-1})- (\nabla f(x^{s})+\nabla f(x^{s-1})))\right\|^2
    \\
    & \leq 
    2\left\|\frac{1}{n}\sum_{s=\max\{1,t-T_1+1\}}^t\frac{\mathbb{E}[|\tilde{I}_s^t|]}{b}\sum_{i\in I^s} (\nabla f_i(x^{s})-\nabla f_i(x^{s-1})- (\nabla f(x^{s})+\nabla f(x^{s-1})))\right\|^2
    \label{eq:First-Variance-10}
    \\ & \hspace{5mm} +
    \frac{2T_1}{n^2}\sum_{s=\max\{1,t-T_1+1\}}^t\left(|\tilde{I}_s^t|-\mathbb{E}[|\tilde{I}_s^t|]\right)^2\left\|\frac1b\sum_{i\in I^s} (\nabla f_i(x^{s})-\nabla f_i(x^{s-1})- (\nabla f(x^{s})+\nabla f(x^{s-1}))\right\|^2
    \label{eq:First-Variance-5}
    .
\end{align}

First, we bound the term \eqref{eq:First-Variance-10}.
We can see that $\frac{\mathbb{E}[|\tilde{I}_s^t|]}{b}\sum_{i\in I^s} (\nabla f_i(x^{s})-\nabla f_i(x^{s-1})- (\nabla f(x^{s})+\nabla f(x^{s-1})))$ is a martingale difference sequence.
Moreover, by the vector Bernstein inequality without replacement (\cref{proposition:BernsteinNoReplacement}) and \cref{assumption:Heterogeneity}, we have
\begin{align}
    \left\|\frac{\mathbb{E}[|\tilde{I}_s^t|]}{b}\sum_{i\in I^s} (\nabla f_i(x^{s})-\nabla f_i(x^{s-1})- (\nabla f(x^{s})-\nabla f(x^{s-1})))\right\|^2 \leq \frac{C_1\mathbb{E}[|\tilde{I}_s^t|]^2\zeta^2}{b}\|x^s-x^{s-1}\|^2
    \leq \frac{C_1n^2\zeta^2}{b}\|x^s-x^{s-1}\|^2
\end{align}
with probability at least $1-\frac{\nu}{5T^2}$ for each $t$ and $s$ in \eqref{eq:First-Variance-10}. This allows us to use the Azuma-Hoeffding inequality with high probability (\cref{proposition:Azuma}).
Consequently, with probability at least $1-\frac{\nu}{5T}-T\cdot \frac{\nu}{5T^2}=1-\frac{2\nu}{5T}$, it holds that 
\begin{align}
    \begin{split}
        \left\|\frac{1}{n}\sum_{s=\max\{1,t-T_1+1\}}^t\frac{\mathbb{E}[|\tilde{I}_s^t|]}{b}\sum_{i\in I^s} (\nabla f_i(x^{s})-\nabla f_i(x^{s-1})- (\nabla f(x^{s})-\nabla f(x^{s-1})))\right\|^2
    \\ \leq 
    \frac{C_1^2\zeta^2}{b}\sum_{s=\max\{1,t-T_1+1\}}^t\|x^s-x^{s-1}\|^2
    \label{eq:First-Variance-11}
    \end{split}
\end{align}
for each $t\in[T]$.
Therefore, \eqref{eq:First-Variance-11} holds for all $t$ with probability $1-\frac{2\nu}{5}$.

As for the term \eqref{eq:First-Variance-5},
by the Bernstein inequality without replacement (\cref{proposition:BernsteinNoReplacement}) and \cref{assumption:Heterogeneity}, we have
\begin{align}
\label{eq:First-Variance-7}
    \left\|\frac{1}{b}\sum_{i\in I_s} (\nabla f_i(x^{s})-\nabla f_i(x^{s-1})- (\nabla f(x^{s})+\nabla f(x^{s-1})))\right\|^2
    \leq \frac{C_1^2\zeta^2}{b}\|x^s-x^{s-1}\|^2,
\end{align}
for all $s$ with probability at least $1-\frac{\nu}{5}$. 
We move to bound the difference $\left(|\tilde{I}_s^t|-\mathbb{E}[|\tilde{I}_s^t|]\right)^2$.
For this purpose, we regard this as a function of (at most) $T_1$ variables 
$I^t,\cdots,I^{s}$ and prepare a ``reverse'' filtration $\tilde{\mathcal{F}}=\{\tilde{\mathcal{F}}^{t}_s\}_{s=t}^{\max\{1,t-T_1+1\}}$ with $\tilde{\mathcal{F}}_s^t = \sigma(I_t,I_{t-1},\cdots,I_s)$.
Then, the sequence $\{|\tilde{I}_s^t|\}_{s=\max\{1,t-T_1+1\}}^{t}$ is a measurable process with respect to $\tilde{\mathcal{F}}$.
We consider the conditional expectation of $|\tilde{I}_s^t|-|\tilde{I}_{s+1}^t|$ with respect to $\tilde{\mathcal{F}}$.
When samples in $\tilde{I}_{s+1}^t$ are not chosen between $t$ to $s+1$, each of them is chosen with probability $\frac{b}{n}$ for the first time at step $s$. Thus, we have
\begin{align}
    \mathbb{E}_s\left[|\tilde{I}_{s+1}^t|-|\tilde{I}_s^t| \mid \tilde{\mathcal{F}}_{s+1}^t\right]
    = \frac{b}{n}|\tilde{I}_{s+1}^t|
    ,
\end{align}
which leads to $\mathbb{E}_s\left[|\tilde{I}_s^t|\left| \tilde{\mathcal{F}}^{t}_{s+1}\right.\right] =  \left(1-\frac{b}{n}\right)|\tilde{I}_{s+1}^t|$.
Hence, the process $\{u_s^t\coloneqq |\tilde{I}_s^t| - \left(1-\frac{b}{n}\right)|\tilde{I}_{s+1}^t|\mid t>s\geq t-T_1+1\}$ is a martingale with respect to $\tilde{\mathcal{F}}$ and satisfies $\mathbb{E}_s\left[u_s^t\left| \tilde{\mathcal{F}}^{t}_{s+1}\right.\right]=0$. 
In addition, let $A=\{\underbrace{1,\cdots,1}_{|\tilde{I}_{s+1}^t|},\underbrace{0,\cdots,0}_{n-|\tilde{I}_{s+1}^t|}\}$ and $\tilde{A}=(\tilde{a}_1,\cdots,\tilde{a}_b)$ be a random sample without replacement from $A$, 
Then, $u_s^t$ conditioned on $\tilde{\mathcal{F}}^{t}_{s+1}$ follows the same distribution as that of $\sum_{l=1}^b\tilde{a}_i-\mathbb{E}\left[\sum_{l=1}^b\tilde{a}_i\right]$.
This means that, using \cref{proposition:BernsteinNoReplacement}, we have
$\|u_s^t\|\leq C_1\sqrt{b}$ with probability at least $1-\frac{1}{5T^2}$.
Finally, we apply \cref{proposition:Azuma} to bound $|\tilde{I}_s^t|=\sum_{\tau=t}^s \left(1-\frac{b}{n}\right)^{(\tau-s)} u_{\tau}^t + n\left(1-\frac{b}{n}\right)^{(t-s+1)}$, which yields that
\begin{align}
    \left||\tilde{I}_s^t|-\mathbb{E}[|\tilde{I}_s^t|]\right|^2 \leq C_1^2 \sum_{\tau=t}^s C_1^2 \left(1-\frac{b}{n}\right)^{(\tau-s)} b \leq C_1^4 b T_1 = C_1^5 n
    \label{eq:First-Variance-8}
\end{align}
with probability at least $1-\frac{\nu}{5T}-T\cdot \frac{\nu}{5T^2}=\frac{2\nu}{5T}$.

Combining \eqref{eq:First-Variance-7} and \eqref{eq:First-Variance-8}, with probability $1-\frac{\nu}{5}$, we get
\begin{align}
        \frac{T_1}{n^2}\sum_{s=\max\{1,t-T_1+1\}}^t\left(|\tilde{I}_s^t|-\mathbb{E}[|\tilde{I}_s^t|]\right)^2\left\|\frac1b\sum_{i\in I^s} (\nabla f_i(x^{s})-\nabla f_i(x^{s-1})- (\nabla f(x^{s})+\nabla f(x^{s-1})))\right\|^2
        \\ \leq  
         \frac{C_1^8\zeta^2}{b^2}\sum_{s=\max\{1,t-T_1+1\}}^t\|x^s-x^{s-1}\|^2,
         \label{eq:First-Variance-12}
\end{align}
by letting $T_1 = \frac{n}{b}C_1$. 
Finally, we get the assertion by combining \eqref{eq:First-Variance-11} and \eqref{eq:First-Variance-12}, and applying $\frac{2C_1^2}{b} + \frac{2C_1^8}{b^2}\leq \frac{4C_1^8}{b}$.

\end{proof}

\begin{proof}[Proof of \cref{lemma:First-Auxiliary-C}]
As for the Option II, the assertion directly follows from the definition $y_i^0 = \nabla f_i(x^0)\ (i=1,\cdots,n)$. 
Henceforth, we prove the bound 
\begin{align}
    \left\|\frac{1}{n}\sum_{i\in \tilde{I}_1^t}(y^0_i-\nabla f_i(x^0))
        \right\|^2
        \leq
        \frac{4C_1^2}{b}\left(\sigma_c^2+\frac{G_c^2}{b}\right)
\end{align}
when we use Option I. 
To this end, we decompose $\left\|\frac{1}{n}\sum_{i\in \tilde{I}_1^t}(y^0_i-\nabla f_i(x_0))\right\|^2$ as
\begin{align}
    \left\|\frac{1}{n}\sum_{i\in \tilde{I}_1^t}(y^0_i-\nabla f_i(x^0))\right\|^2
    &=
    \left\|\frac{1}{n}\left(\frac{|\tilde{I}_1^t|}{b}\sum_{i\in I^1}(\nabla f_i(x^0)-\nabla f(x^0)) + \sum_{i\in \tilde{I}_1^t}(\nabla f(x^0)-\nabla f_i(x^0))\right)\right\|^2
    \\ &\leq \label{eq:First-Variance-13}
    2\left\|\frac{1}{n}\frac{|\tilde{I}_1^t|}{b}\sum_{i\in I^1}(\nabla f_i(x^0)-\nabla f(x^0))\right\|^2
    +
    2\left\|\frac{1}{n}\sum_{i\in \tilde{I}_1^t}(\nabla f_i(x^0)-\nabla f(x^0))\right\|^2,
\end{align}
where we use the inequality $\|a+b\|^2\le 2\|a\|^2+2\|b\|^2$. 
For the first term in \eqref{eq:First-Variance-13},  \cref{proposition:BernsteinNoReplacement} and \cref{assumption:BoundedGradient} imply that 
\begin{align}\label{eq:First-Variance-14}
    \left\|\frac{1}{n}\frac{|\tilde{I}_1^t|}{b}\sum_{i\in I^1}(\nabla f_i(x^0)-\nabla f(x^0))\right\|^2
     \leq  
    \frac{|\tilde{I}_1^t|^2}{n^2b^2}C_1^2b\left(\sigma_c^2 + \frac{G_c^2}{b}\right)
    \leq \frac{C_1^2}{b}\left(\sigma_c^2 + \frac{G_c^2}{b}\right)
    ,
\end{align}
holds with probability at least $1-\frac{\nu}{2}$ for all $t$. 
For the second term in \eqref{eq:First-Variance-13}, we have
\begin{align}\label{eq:First-Variance-15}
    \left\|\frac{1}{n}\sum_{i\in \tilde{I}_1^t}(\nabla f_i(x^0)-\nabla f(x^0))\right\|^2
    \leq \frac{1}{n^2}C_1^2|\tilde{I}_1^t|\left(\sigma_c^2 + \frac{G_c^2}{|\tilde{I}_1^t|}\right)
    \leq \frac{C_1^2}{b}\left(\sigma_c^2 + \frac{G_c^2}{b}\right)
    ,
\end{align}
from \cref{proposition:BernsteinNoReplacement} and \cref{assumption:BoundedGradient}, with probability at least $1-\frac{\nu}{2T}$ for each $t$ and at least $1-\frac{\nu}{2}$ uniformly over all $t$.

Substituting \eqref{eq:First-Variance-13} and \eqref{eq:First-Variance-14} to \eqref{eq:First-Variance-12}, we obtain the desired bound.

\end{proof}

\begin{proof}[Proof of \cref{lemma:First-VarianceBound}]
We first observe that $v^t - \nabla f(x^{t})$ is written as
\begin{align}
\label{eq:First-Variance-1}
    v^t - \nabla f(x^{t}) = \frac{1}{n} \sum_{s=1}^t \left( \frac{|\tilde{I}_s^t|}{b}\sum_{i\in I^s} (\nabla f_i(x^{s})-\nabla f_i(x^{s-1}))
    - \sum_{i\in \tilde{I}_s^t}(\nabla f_i(x^{s})-\nabla f_i(x^{s-1}))
    \right)+\frac{1}{n}\sum_{i\in \tilde{I}_1^t}(y^0_i-\nabla f_i(x^0))
    .
\end{align}
We can ensure that if $t-s$ is sufficiently large, every $f_i$ is sampled at least once between $s+1$ and $t$ with high probability.
Indeed, for each $f_i$ and $\nu>0$, the probability that $f_i$ is not sampled between $s+1$ and $t$ is bounded as
\begin{align}
    \label{eq:First-Variance-2}
    \left(1-\frac{1}{n}\right)^{\sum_{s=\max\{1,t-s+1\}}^t b}
    \leq
    \left(1-\frac{1}{n}\right)^{\sum_{s=\max\{1,t-T_1+1\}}^t b}
    \leq
    \left(1-\frac{1}{n}\right)^{nC_1}
    \leq
    \exp \left(-C_1\right)
    ,
\end{align}
where we use $\sum_{s=\max\{1,t-T_1+1\}}^t b\le T_1b=nC_1$ in the second inequality.
By taking $C_1 = \Omega\left(\log \frac{nT}{\nu}\right)$, the right-hand side of \eqref{eq:First-Variance-2} is bounded by $\frac{\nu}{nT}$.
In other words, $\tilde{I}_{s}^t = \emptyset$ with probability at least $1-\nu$ for every $t$ and $s\leq  t-T_1$. Henceforth, we assume  $\tilde{I}_{s}^t = \emptyset$ and focus on the errors between $\max\{1,t-T_1+1\}\le s\leq t$.

When $\tilde{I}_{s}^t = \emptyset$ holds for $s\leq t-T_1$, the variance term $\|v^t - \nabla f(x^{t})\|^2$ is decomposed as 
\begin{align}
    & \|v^t - \nabla f(x^{t})\|^2
    \\ & =
    \left\|
        \frac{1}{n} \sum_{s=\max\{1,t-T_1+1\}}^t \left( \frac{|\tilde{I}_s^t|}{b}\sum_{i\in I_s} (\nabla f_i(x^{s})-\nabla f_i(x^{s-1}))
    -
    \sum_{i\in \tilde{I}_s^t}(\nabla f_i(x^{s})-\nabla f_i(x^{s-1}))
    \right)+\frac{\mathbbm{1}[t\leq T_1]}{n}\sum_{i\in \tilde{I}_1^t}(y^0_i-\nabla f_i(x^0))
    \right\|^2
    \\ \label{eq:First-Variance-100}\begin{split}
    &\leq
    3\underbrace{
    \left\|
    \frac{1}{n} \sum_{s=\max\{1,t-T_1+1\}}^t \sum_{i\in \tilde{I}_s^t}(\nabla f_i(x^{s})-\nabla f_i(x^{s-1}) - (\nabla f(x^{s})-\nabla f(x^{s-1})))\right\|^2 
    }_{\rm (a)}
    \\ 
    & \hspace{3.5mm}
    + 
    3\underbrace{
    \left\|\frac{1}{n}\sum_{s=\max\{1,t-T_1+1\}}^t\frac{|\tilde{I}_s^t|}{b}\sum_{i\in I^s} (\nabla f_i(x^{s})-\nabla f_i(x^{s-1})- (\nabla f(x^{s})-\nabla f(x^{s-1})))\right\|^2
    }_{\rm (b)}
    +3\underbrace{
    \left\|\frac{\mathbbm{1}[t\leq T_1]}{n}\sum_{i\in \tilde{I}_1^t}(y^0_i-\nabla f_i(x^0))
    \right\|^2
     }_{\rm (c)},
     \end{split}
\end{align}
by the inequality $\|a+b+c\|^2\le 3(\|a\|^2+\|b\|^2+\|c\|^2)$.
Then, we give the bound of \eqref{eq:First-Variance-100} for Option I and Option II, respectively.

\paragraph{Option I} 
According to \cref{lemma:First-Auxiliary-A,lemma:First-Auxiliary-B,lemma:First-Auxiliary-C}, we have
\begin{align}
   & \|v^t - \nabla f(x^{t})\|^2 \\ & \leq
    \underbrace{
    \frac{3C_1^3\zeta^2}{b}\sum_{s=\max\{1,t-T_1+1\}}^t
    \|x^s-x^{s-1}\|^2
    }_{\rm (a)}
    +
    \underbrace{
    \frac{12C_1^8\zeta^2}{b}\sum_{s=\max\{1,t-T_1+1\}}^t
    \|x^s-x^{s-1}\|^2 
    }_{\rm (b)}
    +
     \underbrace{\frac{12C_1^2\left(\sigma^2_c+\frac{G_c^2}{b}\right)\mathbbm{1}[t\leq T_1]}{b}}_{\rm (c)}
    \\ & \leq
    \left(3C_1^3 + 12C_1^8\right)\frac{\zeta^2}{b}\sum_{s=\max\{1,t-T_1+1\}}^t
    \|x^s-x^{s-1}\|^2 
    +\frac{12C_1^2\left(\sigma^2_c+\frac{G_c^2}{b}\right)\mathbbm{1}[t\leq T_1]}{b}
\end{align}
with probability at least $1-3\nu$ uniformly over all $t$.

\paragraph{Option II} 
Almost as well as the previous case, 
we have
\begin{align}
    \|v_t - \nabla f(x^{t})\|^2 & \leq
    \underbrace{
    \frac{3C_1^3\zeta^2}{b}\sum_{s=\max\{1,t-T_1+1\}}^t
    \|x^s-x^{s-1}\|^2
    }_{\rm (a)}
    +
    \underbrace{
    \frac{12C_1^8\zeta^2}{b}\sum_{s=\max\{1,t-T_1+1\}}^t
    \|x^s-x^{s-1}\|^2 
    }_{\rm (b)}
    +
     \underbrace{0}_{\rm (c)}
    \\ & \leq
    \left(3C_1^3 + 12C_1^8\right)\frac{\zeta^2}{b}\sum_{s=\max\{1,t-T_1+1\}}^t
    \|x^s-x^{s-1}\|^2 
\end{align}
with probability at least $1-2\nu$ uniformly over all $t$.

By replacing $C_1$ with $C_1\vee 1$ and applying $3C_1^3 + 12C_1^8\leq 15C_1^8$, we obtain the desired bound for both cases.
\end{proof}

Now, we are ready to prove the first-order convergence of SLEDGE.

\begin{proof}[Proof of \cref{theorem:First-Main}]

Summing up \eqref{eq:First-FunctionDecrease-3} over all $t=1,2,\cdots,T$ and arranging the terms, we get
\begin{align}
\sum_{t=1}^T \|\nabla f(x^{t-1})\|^2 \leq
\frac{2}{\eta}\left[\left(f(x^0)-f(x^{t})\right) -
\sum_{t=1}^T\left(\frac{1}{2\eta}-\frac{L}{2}\right)\|x^t-x^{t-1}\|^2+\eta \sum_{t=1}^T \|\nabla f(x^{t-1}) - v^{t-1}\|^2\right]
+\frac{2Tr^2}{\eta^2}.
\end{align}
Applying \cref{lemma:First-VarianceBound} to this, we obtain that
\begin{align}
& \sum_{t=1}^T \|\nabla f(x^{t-1})\|^2 \\ & \leq
\begin{cases} \displaystyle
\frac{2}{\eta}\left[\left(f(x^0)-f(x^{t})\right) -
\left(\frac{1}{2\eta}-\frac{L}{2}-\frac{15 C_1^9\eta\zeta^2n}{b^2}\right)\sum_{t=1}^T\|x^t-x^{t-1}\|^2
+\frac{12\eta C_1^3 n}{b^2}\left(\sigma_c^2+\frac{G_c^2}{b}\right)\right]
+\frac{2Tr^2}{\eta^2}& (\text{Option I})
\label{eq:First-Variance-15}
\\ \displaystyle
\frac{2}{\eta}\left[\left(f(x^0)-f(x^{t})\right) -
\left(\frac{1}{2\eta}-\frac{L}{2}-\frac{15C_1^9\eta \zeta^2n}{b^2}\right)\sum_{t=1}^T\|x^t-x^{t-1}\|^2
\right]
+\frac{2Tr^2}{\eta^2} & (\text{Option II})
\end{cases}
\end{align}
with probability at least $1-3\nu$.

\paragraph{Option I} 
We take $\eta = \min \left\{\frac{1}{2L},\frac{b}{C_1^4 \zeta \sqrt{60n}} \right\}=\tilde{\Theta}\left(\frac{b}{Lb\lor  \zeta \sqrt{n}}\right)$, which implies $\frac{1}{2\eta}-\frac{L}{2}-\frac{15 C_1^9\eta\zeta^2n}{b^2}\geq 0$.
We have $\frac{2Tr^2}{\eta^2}\leq \frac{T\eps^2}{2}$ by letting $r\leq \frac{\eta\eps}{2}
$.
Also, we have $f(x_0)-f(x^{t})\leq \Delta$ by \cref{assumption:GlobalInfimum}.
Summarizing these, we get \begin{align}
\sum_{t=1}^T \|\nabla f(x^{t-1})\|^2\leq \frac{2\Delta}{\eta}+\frac{12C_1^2n}{b^2}\left(\sigma_c^2+\frac{G_c}{b}\right) + \frac{T\eps^2}{2}.
\end{align}
Thus, by setting $T\geq \left(\frac{2\Delta}{\eta} + \frac{12C_1^3n}{b^2}\left(\sigma_c^2+\frac{G_c^2}{b}\right)\right)\frac{2}{\eps^2}=\tilde{O}\Biggl(\frac{\left(L+ \frac{\zeta\sqrt{n}}{b}\right)\Delta + \frac{n}{b^2}\left(\sigma_c^2+\frac{G_c^2}{b}\right)}{\eps^2}\Biggr)$, we obtain $
\frac{1}{T}\sum_{t=1}^T \|\nabla f(x^{t-1})\|^2\le \epsilon^2
$
with probability at least $1-3\nu$, which implies that there exists some $t$ such that $0\leq t\leq T-1$ and $\|\nabla f(x^{t})\|^2\leq \eps^2$.
From this, the desired conclusion is obtained.

\paragraph{Option II} 
As well as Option I, we take $\eta = \min \left\{\frac{1}{2L},\frac{b}{C_1^4 \zeta \sqrt{60n}} \right\}=\tilde{\Theta}\left(\frac{b}{Lb\lor  \zeta \sqrt{n}}\right)$, which implies $\frac{1}{2\eta}-\frac{L}{2}-\frac{15 C_1^9\eta\zeta^2n}{b^2}\geq 0$.
In addition, by letting $r\leq \frac{\eta\eps}{2}
$, we obtain $\sum_{t=1}^T \|\nabla f(x^{t-1})\|^2\leq \frac{2\Delta}{\eta}+ \frac{T\eps^2}{2}$.
Then, by taking $T\geq \frac{4\Delta}{\eta\eps^2}=\tilde{O}\left(\frac{L\Delta}{\eps^2}\right)$, there exists some $t$ such that $0\leq t\leq T-1$ and $\|\nabla f(x^{t})\|^2\leq \eps^2$, with probability at least $1-3\nu$.
Accordingly, we get the assertion for Option II.

\end{proof}


\subsection{Finding Second-order Stationary Points (Proof of \cref{theorem:Second-Main})}
The goal of this subsection is to show that \AlgMain\ is the single-loop algorithm with theoretical guarantee for finding second-order stationary points.

The argument follows that of \cite{jin2017escape,ge2019stabilized,li2019ssrgd};
Let $x^{\tau_0}$ be a point such as $\lambda_{\rm min}(\nabla f(x^{\tau_0}))\leq - \delta$.
Around that point, we consider two points $x_1$ and $x_2$ such that $\langle x_1,\mathsf{e}\rangle\approx\langle x_2,\mathsf{e}\rangle$, where $\mathsf{e}$ is the eigenvector of $\lambda_{\rm min}(\nabla f(x^{\tau_0}))$. 
Then, two coupled sequences that \AlgMain\ generates from the two initial points ($x_1$ and $x_2$) will be separated exponentially, as long as they are in a small region around the initial points.
This means that if we add some noise to the sequence around a saddle point, then with a certain probability, the algorithm can move away from the saddle point.

We again emphasize that, although this proof outline has been popular, we face the difficulties arising from the single-loop structure of the algorithm. Many existing algorithms compute periodic full gradient and can refresh their gradient estimators around saddle points. In contrast, our single-loop algorithm does not use full gradient, meaning that we have to deal with the error accumulated before that point, and it is not trivial whether such errors can be sufficiently small so that the direction of the negative eigenvalue can be found by the gradient estimator. This is the first difficulty, and we found that taking minibatch size as large as $b\gtrsim \sqrt{n}+\frac{\zeta^2}{\delta^2}$ is sufficient. When $\delta = O(\sqrt{\eps})$, $\frac{\zeta^2}{\delta^2}$ is about $O(\sqrt{n}+\frac{1}{\eps})$, which is usually assumed in existing literature \citep{ge2019stabilized,li2019ssrgd}.
Secondly, our estimator is more correlated due to the $|\tilde{I}_s^t|$ term, thus requiring more delicate analysis than that for SSRGD \citep{li2019ssrgd}, which is based on SARAH \citep{nguyen2017sarah,nguyen2017stochastic}.

We formalize the exponential separation of two sequences in the following lemma.

\begin{lemma}[Small stuck region]\label{lemma:Small-Stuck-Region}
Let $\{x^t\}$ be a sequence generated by \AlgMain\ and suppose that there exists a step $\tau_0$ such that $-\gamma :=\lambda_{\rm min}(\nabla^2 f(x^{\tau_0}))\leq -\delta$ holds.
We denote the smallest eigenvector direction of $\lambda_{\rm min}(\nabla^2 f(x^{\tau_0}))$ by $e$.
Moreover, we define a coupled sequence $\{\tilde{x}^{t}\}$ by running \AlgMain\ with $\tilde{x}^0 = x^0$ and share the same choice of randomness, \textit{i.e.}, minibatches and noises with $\{x^{t}\}$, except for the noise at some step $\tau(>\tau_0)$: $\tilde{\xi}^{\tau}
= \xi^{\tau} - r_e \mathsf{e}$ with $r_e \geq \frac{r\nu }{T\sqrt{d}}$. 
Let $w^t = x^t - \tilde{x}^t$, $v^t = \frac1n\sum_{i=1}^n y_i^t$, $\tilde{v}^t =\frac1n \sum_{i=1}^n\tilde{y}_i^t$, and $g^t = v^t - \nabla f(x^t) - (\tilde{v}^t - \nabla f(\tilde{x}^t))$.
Here $\tilde{y}_i^t$ is the counterpart of $y_i^t$ and corresponds to $\{\tilde{x}^{t}\}$.

Then, there exists constants $\Cb=\tilde{O}(1),{\Cd}=O(1)$, such that if we take $b\geq \sqrt{n}+\frac{\Cb^2\zeta^2}{\delta^2}$, $\eta = \tilde{\Theta}\left(\frac{1}{L}\right)$, and $T_2=\frac{{\Cd}\log \frac{\delta }{ \Cb \rho r_e}}{\eta \gamma}\leq \tilde{O}\left(\frac{L}{\delta}\right)$, with probability at least $1-\frac{\nu}{T}~(\nu \in (0,T))$, it holds that 
\begin{align}
    \max_{\tau_0 \leq t\leq \tau + T_2}\{\|x^t-x^\tau\|,\|\tilde{x}^t-x^\tau\|\} \geq  \frac{\delta}{\Cb\rho}
    .
\end{align}

\end{lemma}
In order to show \cref{lemma:Small-Stuck-Region}, we need the following lemma, which is analogous to \cref{lemma:First-Auxiliary-A,lemma:First-Auxiliary-B,lemma:First-Auxiliary-C}.
\begin{lemma}\label{lemma:Second-Auxiliary-A}
Under the same assumption as that of \cref{lemma:Small-Stuck-Region}, we assume $\max_{\tau_0 \leq t\leq \tau + T_2}\{\|x^t-x^\tau\|,\|\tilde{x}^t-x^\tau\|\} <  \frac{\delta}{\Cb\rho}$. Then, the following holds uniformly for all $t \leq \tau + T_2$ with probability at least $1-\frac{\nu}{T}$:
\begin{align}
    \left\|g^t\right\| 
    \leq \begin{cases}
    0 & (t < \tau )\\
    \displaystyle
    \frac{{\Cc}\zeta r_e}{\sqrt{b}}& (t = \tau)
    \\ 
    \displaystyle
    \frac{{\Cc}\zeta r_e}{\sqrt{b}}
    +
    \frac{{\Cc}\zeta}{\sqrt{b}}\sqrt{\sum_{s=\max\{\tau+1,t-T_1+1\}}^{t}\|w^s - w^{s-1}\|^2}
    +
    \frac{{\Cc}\delta}{\Cb\sqrt{b}}\sqrt{\sum_{s=\max\{\tau,t-T_1+1\}}^{t}\|w^{s}\|^2}& (\text{otherwise}),
    \end{cases}
\end{align}
where $T_1 = \frac{n}{b}C_1$. Here ${\Cc}=\tilde{O}(1)$ is a sufficiently large constant.
\end{lemma}

\begin{proof} 
As for the case $t<\tau$, the assertion directly follows from the definition of $\{\tilde{x}^t\}$.
For the proof of the rest cases, we use notations as follows:
\begin{align}
    H&=\nabla^2f(x^{\tau_0}),\\
    H_i&=\nabla^2 f_i(x^{\tau_0}),\\
    \mathrm{d}H^t&=\int_0^1 (\nabla^2 f(\tilde{x}^t + \theta(x^t-\tilde{x}^t)) - H)\mathrm{d}\theta,\\
    \mathrm{d}H^t_i&=\int_0^1 (\nabla^2 f_i(\tilde{x}^t + \theta(x^t-\tilde{x}^t)) - H_i)\mathrm{d}\theta.
\end{align}
Moreover, to simplify the notation, we denote
\begin{align}
    u_i^s := (\nabla f_i(x^{s})-\nabla f_i(\tilde{x}^{s}))
     - (\nabla f_i(x^{s-1})-\nabla f_i(\tilde{x}^{s-1}))
     - (\nabla f(x^{s})-\nabla f(\tilde{x}^{s}))
     +(\nabla f(x^{s-1})-\nabla f(\tilde{x}^{s-1})).
\end{align}
We have that $\mathbb{E}_i[u_i^s]=0$, where the expectation is taken over the choice of $i$. Furthermore, for $s\geq \tau + 1$, by using the Hessian-heterogenity (\cref{assumption:Heterogeneity}) and the Hessian Lipschitzness (\cref{assumption:HessianLipschitz}), we have that
\begin{align}
    \|u_i^s\| & =
    \|(\nabla f_i(x^{s})-\nabla f_i(\tilde{x}^{s}))
     - (\nabla f_i(x^{s-1})-\nabla f_i(\tilde{x}^{s-1}))
     - (\nabla f(x^{s})-\nabla f(\tilde{x}^{s}))
     +(\nabla f(x^{s-1})-\nabla f(\tilde{x}^{s-1}))\|
    \\ & =
    \left\|\int_0^1 \nabla^2 f_i(\tilde{x}^{s}-\theta (x^{s}-\tilde{x}^{s}))(x^s-\tilde{x}^{s})\mathrm{d}\theta 
    -  \int_0^1 \nabla^2 f_i(\tilde{x}^{s-1}-\theta (x^{s-1}-\tilde{x}^{s-1}))(x^{s-1}-\tilde{x}^{s-1})\mathrm{d}\theta \right.
     \\ & \hspace{5mm} \left.
     -\int_0^1 \nabla^2 f(\tilde{x}^{s}-\theta (x^{s}-\tilde{x}^{s}))(x^s-\tilde{x}^{s}) \mathrm{d}\theta 
    + \int_0^1 \nabla^2 f(\tilde{x}^{s-1}-\theta (x^{s-1}-\tilde{x}^{s-1}))(x^{s-1}-\tilde{x}^{s-1})\mathrm{d}\theta 
    \right\|
    \\ & 
    = \|
    (H_i + \mathrm{d}H_i^s)w^s - (H_i + \mathrm{d}H_i^{s-1})w^{s-1}
    - (H + \mathrm{d}H^s)w^s + (H + \mathrm{d}H^{s-1})w^{s-1}
    \|
    \\ & 
     \leq \|H_i-H\|\|w^s - w^{s-1}\| + (\|\mathrm{d}H_i^s\|+\|\mathrm{d}H^s\|)\|w^s\|
     +(\|\mathrm{d}H_i^{s-1}\|+\|\mathrm{d}H^{s-1}\|)\|w^{s-1}\|
     \\ & \leq
     \zeta \|w^s - w^{s-1}\| + 2\rho \max_{0\leq \theta\leq 1}\{\|\tilde{x}^{s}-\theta (x^{s}-\tilde{x}^{s})-x^\tau\|\}\|w^s\| + 2\rho \max_{0\leq \theta\leq 1}\{\|\tilde{x}^{s-1}-\theta (x^{s-1}-\tilde{x}^{s-1})-x^\tau\|\}\|w^{s-1}\|
     \\ & =
     \zeta \|w^s - w^{s-1}\| + 2\rho \max\{\|x^{s}-x^\tau\|,\|\tilde{x}^{s}-x^\tau\|\}\|w^s\| + 2\rho \max\{\|x^{s-1}-x^\tau\|,\|\tilde{x}^{s-1}-x^\tau\|\}\|w^{s-1}\|
     \\ & <
     \zeta \|w^s - w^{s-1}\| + \frac{2\delta}{\Cb}\|w^s\| + \frac{2\delta}{\Cb}\|w^{s-1}\|
     ,\label{eq::Second-Auxiliary-A-10}
\end{align}
where we use $\max_{\tau_0 \leq t\leq \tau + T_2}\{\|x^t-x^\tau\|,\|\tilde{x}^t-x^\tau\|\} <  \frac{\delta}{\Cb\rho}$ for the last inequality. For $s = \tau $, by \cref{assumption:Heterogeneity}, we have $\|u_i^{\tau}\| = \|(\nabla f_i(x^{\tau})-\nabla f_i(\tilde{x}^{\tau}))
     - (\nabla f(x^{\tau})-\nabla f(\tilde{x}^{\tau}))\|\leq 2\zeta\|x^{\tau}-\tilde{x}^{\tau}\|=2\zeta r_e$.

Recall the discussion in \cref{lemma:First-VarianceBound}, we have 
\begin{align}
    g^{t} &=  v^{t} - \nabla f(x^{t}) - \tilde{v}^{t} + \nabla f(\tilde{x}^{t})
    \\ &= 
    \frac{1}{n} \sum_{s=\max\{\tau,t-T_1+1\}}^t \left( \frac{|\tilde{I}_s^t|}{b}\sum_{i\in I^s} (\nabla f_i(x^{s})-\nabla f_i(x^{s-1}))
    - \sum_{i\in \tilde{I}_s^t}(\nabla f_i(x^{s})-\nabla f_i(x^{s-1}))
    \right)
    \\ &\quad  -
    \frac{1}{n} \sum_{s=\max\{\tau,t-T_1+1\}}^t \left( \frac{|\tilde{I}_s^t|}{b}\sum_{i\in I^s} (\nabla f_i(\tilde{x}^{s})-\nabla f_i(\tilde{x}^{s-1}))
    - \sum_{i\in \tilde{I}_s^t}(\nabla f_i(\tilde{x}^{s})-\nabla f_i(\tilde{x}^{s-1}))
    \right)
    \\ &= 
     \frac{1}{n} \sum_{s=\max\{\tau,t-T_1+1\}}^{t} \left( 
     \frac{|\tilde{I}_s^{t}|}{b}\sum_{i\in I^s} u_i^s - \sum_{i\in \tilde{I}_s^t}u_i^s  \right)
    \\ &= 
    \begin{cases}\displaystyle
     \frac{1}{n} \left( \frac{|\tilde{I}_{\tau}^{\tau}|}{b}\sum_{i\in I^{\tau}} u_i^\tau - \sum_{i\in \tilde{I}_{\tau}^{\tau}} u_i^\tau \right)
      & (t = \tau )\\ \displaystyle
     \frac{1}{n} \left( \frac{|\tilde{I}_{\tau}^{t}|}{b}\sum_{i\in I^{\tau}} u_i^\tau - \sum_{i\in \tilde{I}_{\tau}^{t}} u_i^\tau \right)+
    \frac{1}{n} \sum_{\substack{s=\max\{\tau+1,\\t-T_1+1\}}}^{t} \left( 
     \frac{|\tilde{I}_s^{t}|}{b}\sum_{i\in I^s} u_i^s  \right)- \frac{1}{n} \sum_{\substack{s=\max\{\tau+1,\\t-T_1+1\}}}^{t}  \sum_{i\in \tilde{I}_s^t}u_i^s 
      & (t \geq \tau + 1).
     \end{cases}
     \label{eq::Second-Auxiliary-A-1}
\end{align}

As for the first term in both cases, we have
\begin{align}
    &
    \left\|\frac{1}{n} \left( \frac{|\tilde{I}_{\tau}^{t}|}{b}\sum_{i\in I^{\tau}} u_i^\tau - \sum_{i\in \tilde{I}_{\tau}^{t}} u_i^\tau \right)\right\|
    \leq
    \left\|\frac{1}{n} \frac{|\tilde{I}_{\tau}^{t}|}{b}\sum_{i\in I^{\tau}} u_i^{\tau} \right\| + \left\|\frac{1}{n} \sum_{i\in \tilde{I}_{\tau}^{t}} u_i^{\tau}\right\|
     \leq
     \frac{|\tilde{I}_{\tau}^{t}|}{n}\frac{2C_1 \zeta r_e}{\sqrt{b}}
     \leq \frac{2C_1 \zeta r_e}{\sqrt{b}},
     \label{eq::Second-Auxiliary-A-2}
\end{align}
by using \cref{proposition:BernsteinNoReplacement} and $\|u_i^{\tau}\| \leq 2\zeta r_e$, with probability at least $1-\frac{\nu}{4T}$ for all $t$.

For the second term in the case $t\ge \tau+1$, we follow the same line as the proof of \cref{lemma:First-Auxiliary-B}.
We just replace $\nabla f_i(x^{s})-\nabla f_i(x^{s-1})- (\nabla f(x^{s})-\nabla f(x^{s-1}))$ by $u_i^s$ and use \eqref{eq::Second-Auxiliary-A-10} to obtain that
\begin{align}
\left\|\frac{1}{n} \sum_{s=\max\{\tau+1,t-T_1+1\}}^{t}
     \frac{|\tilde{I}_s^{t}|}{b}\sum_{i\in I^s} u_i^s\right\| &\le     \frac{2C_1^4}{\sqrt{b}}\sqrt{\sum_{s=\max\{1,t-T_1+1\}}^t\left(\zeta \|w^s - w^{s-1}\| + \frac{2\delta}{\Cb}\|w^s\| + \frac{2\delta}{\Cb}\|w^{s-1}\|\right)^2}
    \\ &\le 
    \frac{2C_1^4}{\sqrt{b}}\sqrt{\sum_{s=\max\{1,t-T_1+1\}}^t\left(\zeta \|w^s - w^{s-1}\| + \frac{2\delta}{\Cb}\|w^s\| + \frac{2\delta}{\Cb}\|w^{s-1}\|\right)^2}
.\label{eq:Second-Auxiliary-A-3}
\end{align}
with probability at least $1-\frac{\nu}{4T}$ for all $t$.

Finally, we bound the last term in the case $t\geq \tau + 1$.
By using \cref{proposition:BernsteinNoReplacement}, we obtain
\begin{align}
    \left\|\frac{1}{n} \sum_{s=\max\{\tau+1,t-T_1+1\}}^{t}  \sum_{i\in \tilde{I}_s^t}u_i^s \right\|
   & \leq
    \frac{\sqrt{T_1}}{n} \sqrt{\sum_{s=\max\{\tau+1,t-T_1+1\}}^{t}
    \left\|\sum_{i\in \tilde{I}_s^t}u_i^s \right\|^2}
    \\ & \leq
    \frac{C_1^\frac12}{\sqrt{nb}} \sqrt{\sum_{s=\max\{\tau+1,t-T_1+1\}}^{t}
    C_1^2b\left(\zeta \|w^s - w^{s-1}\| + \frac{2\delta}{\Cb}\|w^s\| + \frac{2\delta}{\Cb}\|w^{s-1}\|\right)^2}
    \label{eq::Second-Auxiliary-A-7} 
    \\ & \leq
    \frac{C_1^\frac32}{\sqrt{b}} \sqrt{\sum_{s=\max\{\tau+1,t-T_1+1\}}^{t}
    \left(\zeta \|w^s - w^{s-1}\| + \frac{2\delta}{\Cb}\|w^s\| + \frac{2\delta}{\Cb}\|w^{s-1}\|\right)^2}
     \label{eq::Second-Auxiliary-A-8}  
\end{align}
with probability at least $1-\frac{\nu}{4T}$ for all $t$.

Combining \eqref{eq::Second-Auxiliary-A-2}, \eqref{eq:Second-Auxiliary-A-3}, and  \eqref{eq::Second-Auxiliary-A-8}, 
we have
\begin{align}
    \|g^{t}\| &\leq \frac{2C_1\zeta r_e}{\sqrt{b}}+\frac{2C_1^4+C_1^\frac32}{\sqrt{b}} \sqrt{\sum_{s=\max\{\tau+1,t-T_1+1\}}^{t}
    \left(\zeta \|w^s - w^{s-1}\| + \frac{2\delta}{\Cb}\|w^s\| + \frac{2\delta}{\Cb}\|w^{s-1}\|\right)^2}
    \\& \leq 
    \frac{{\Cc}\zeta r_e}{\sqrt{b}}
    +
    \frac{{\Cc}\zeta}{\sqrt{b}}\sqrt{\sum_{s=\max\{\tau+1,t-T_1+1\}}^{t}\|w^s - w^{s-1}\|^2}
    +
    \frac{{\Cc}\delta}{\Cb\sqrt{b}}\sqrt{\sum_{s=\max\{\tau,t-T_1+1\}}^{t}\|w^{s}\|^2}
\end{align}
with probability at least $1-\frac{\nu}{T}$ for all $t>\tau$.
Here we take ${\Cc} = \tilde{O}(1)$.
For $t=\tau$, \eqref{eq::Second-Auxiliary-A-2} directly implies the desired bound.
\end{proof}

Now, we move to prove \cref{lemma:Small-Stuck-Region}.
\begin{proof}[Proof of \cref{lemma:Small-Stuck-Region}]
We assume the contrary, i.e.,
$\max_{\tau_0 \leq t\leq \tau + T_2}\{\|x^t-x^\tau\|,\|\tilde{x}^t-x^\tau\|\} <  \frac{\delta}{\Cb\rho}$,
and show the following by induction: for $\tau \leq t\leq \tau + T_2$,
\begin{align}
    &{\rm(a)}\quad \frac12(1+\eta\gamma)^{t-\tau}r_e\leq \|w^t\|\leq 2(1+\eta\gamma)^{t-\tau}r_e\\
    &{\rm(b)}\quad \|w^t - w^{t-1}\|\leq \begin{cases}
    r_e & (\text{for $t=\tau $})\\
    3\eta\gamma (1+\eta\gamma)^{t-\tau}r_e & (\text{for $t\geq \tau + 1$})
    \end{cases}\\
    &{\rm(c)}\quad \|g^t\|\leq 
     \frac{3C_1^\frac12{\Cc}\gamma}{\Cb} (1+\eta\gamma)^{t-\tau}r_e
    .
\end{align}

Then, (a) yields contradiction by taking $t-\tau = T_2 = O\left(\frac{\log\frac{\delta}{\Cb\rho r_e}}{\eta\delta}\right)$ since it holds that
\begin{align}
    \max_{\tau_0 \leq t\leq \tau + T_2}\{\|x^t-x^\tau\|,\|\tilde{x}^t-x^\tau\|\}\ge \frac{1}{2}\|x^t-\tilde{x}^t\| = \frac{1}{2}\|w^t\|\ge \frac{\delta}{\Cb\rho}.
\end{align}
It is easy to check (a) and (b) for $t=\tau$.
As for (c), 
by taking $b \geq \frac{\zeta^2}{\delta^2}$, $\|g^t\|\leq {\Cc} \delta r_e \leq {\Cc} \gamma (1+\eta\gamma)^{t-\tau}r_e$ holds with probability at least $1-\frac{\nu}{T}$ by \cref{lemma:Second-Auxiliary-A}. 

Now, we derive that (a), (b), and (c) are true for $t+1$ if they are true for $t=\tau,\tau + 1, \cdots,t$.
For $t\geq \tau + 1$, we can decompose $w^t$ as 
\begin{align}
    w^t &= w^{t-1} - \eta\left(v^{t-1} - \tilde{v}^{t-1} \right) 
    \\ & =
    w^{t-1} - \eta\left(\nabla f(x^{t-1})-\nabla f(\tilde{x}^{t-1})+v^{t-1} - \nabla f(x^{t-1}) - \tilde{v}^{t-1} + \nabla f(\tilde{x}^{t-1})\right) 
     \\ & =   
    w^{t-1} - \eta\left(\int_0^1 \nabla^2 f(\tilde{x}^{t-1} + \theta(x^{t-1}-\tilde{x}^{t-1}))(x^{t-1} - \tilde{x}^{t-1})\mathrm{d}\theta +v^{t-1} - \nabla f(x^{t-1}) - \tilde{v}^{t-1} + \nabla f(\tilde{x}^{t-1})\right)
    \\ & = 
    w^{t-1} - \eta\left((\mathrm{d}H^{t-1}+H) w^{t-1}+v^{t-1} - \nabla f(x^{t-1}) - \tilde{v}^{t-1} + \nabla f(\tilde{x}^{t-1})\right)   
    \\ & = 
    (I-\eta H)w^{t-1} - \eta (\mathrm{d}H^{t-1}w^{t-1} + g^{t-1})
    \\ & = 
    (I-\eta H)^{t-\tau}w^{\tau}
    -\eta \sum_{s=\tau}^{t-1}(I-\eta H)^{t-1-s}(\mathrm{d}H^{s}w^{s} + g^{s})
    \\ & =
    (1+\eta \gamma)^{t-\tau}r_e\mathsf{e}
    -\eta \sum_{s=\tau}^{t-1}(I-\eta H)^{t-1-s}(\mathrm{d}H^{s}w^{s} + g^{s}),
    \label{eq:Second-SmallStuckRegion-1}
\end{align}
where we use the same notation as the proof of \cref{lemma:Second-Auxiliary-A}. 
According to this decomposition, we verify (a), (b), and (c).

\paragraph{Verifying (a)}
The first term of \eqref{eq:Second-SmallStuckRegion-1} satisfies
\begin{align}
    \|(1+\eta \gamma)^{t+1-\tau}r_e\mathsf{e}\|
    =(1+\eta \gamma)^{t+1-\tau}r_e
    .
\end{align}
Thus, it suffices to bound the norm of $\eta \sum_{s=\tau}^{t-1}(I-\eta H)^{t-1-s}(\mathrm{d}H_{s}w_{s} + y_{s})$ by $\frac12 (1+\eta \gamma)^{t-\tau}r_e$.
We have
\begin{align}
    \left\|\eta \sum_{s=\tau}^{t}(I-\eta H)^{t-s}\mathrm{d}H_{s}w_{s} \right\|
    & \leq
    \eta \sum_{s=\tau}^{t}\|I-\eta H\|^{t-s} \left\|\mathrm{d}H_{s}\right\|\left\|w_{s}\right\|
    \\ & \leq \label{eq:Second-SmallStuckRegion-2} 
    \eta (1+\eta \gamma)^{t-\tau }r_e \sum_{s=\tau}^{t}\left\|\mathrm{d}H_{s}\right\|
    \\ & \leq \label{eq:Second-SmallStuckRegion-3}
    \eta (1+\eta \gamma)^{t-\tau}r_e T_2\frac{\delta}{\Cb}
    \\ & \leq \label{eq:Second-SmallStuckRegion-4} 
     \frac{\delta\eta T_2}{\Cb} (1+\eta \gamma)^{t-\tau}r_e
    \\ & \leq \label{eq:Second-SmallStuckRegion-5} 
     \frac14 (1+\eta \gamma)^{t-\tau}r_e    
     .
\end{align}
For \eqref{eq:Second-SmallStuckRegion-2}, we used the facts that the maximum eigenvalue of $\eta H$ is at most $\eta L\leq 1$ when $\eta \leq \frac1L$ and that the minimum eigenvalue is $-\eta \gamma$, which imply $\|I-\eta H\|\le 1+\eta\gamma$. 
\eqref{eq:Second-SmallStuckRegion-3} follows from the assumptions on $\|w_s\|$.
For \eqref{eq:Second-SmallStuckRegion-4}, we use $t\leq \tau + T_2$ and 
\begin{align}
    \|\mathrm{d}H^{s}\| &= \left\|\int_0^1 (\nabla^2 f(\tilde{x}^s + \theta(x^s-\tilde{x}^s)) - H)\mathrm{d}\theta\right\|
    \\    &\leq \max_{0\leq \theta\leq 1}\rho \|\tilde{x}^s + \theta(x^s-\tilde{x}^s)-x^{\tau_0}\|
    \\
    &= \max_{0\leq \theta\leq 1}\rho \max\{\|x^s-x^{\tau_0}\|,\|\tilde{x}^s-x^{\tau_0}\|\}
    < \rho \frac{\delta}{\Cb\rho}=\frac{\delta}{\Cb},
\end{align}
where the first inequality follows from the hessian Lipschitzness (\cref{assumption:HessianLipschitz}). 
The final inequality \eqref{eq:Second-SmallStuckRegion-5} holds when we take $\Cb$ as $\Cb\leq 4 {\Cd}\log \frac{\delta }{ \Cb \rho r_e}$ (this is satisfied by taking $\Cb=\tilde{O}({\Cd})=\tilde{O}(1)$). 

In addition, we have
\begin{align}
      \left\|\eta \sum_{s=\tau}^{t}(I-\eta H)^{t-s}g^s \right\|
    & \leq 
    \eta \sum_{s=\tau}^{t}\|I-\eta H\|^{t-s}
    \left\|g^{s}\right\|
    \\ & \leq \label{eq:Second-SmallStuckRegion-6}
    \eta \sum_{s=\tau}^{t-1}(1+\eta \gamma)^{t-s}\frac{3{\Cc}\gamma}{\Cb} (1+\eta\gamma)^{s-\tau}r_e
    \\ & = 
    \eta T_2\frac{3{\Cc}\gamma}{\Cb} (1+\eta\gamma)^{t-\tau}r_e
    \\ & \leq \label{eq:Second-SmallStuckRegion-7}
    \frac{3{\Cc}{\Cd}\log \frac{\delta}{\Cb\rho r_e}}{\Cb}(1+\eta \gamma)^{t-\tau}\left({\Cc} \delta + \gamma\right) r_e
    \\ & \leq \label{eq:Second-SmallStuckRegion-8}    
    \frac14 (1+\eta \gamma)^{t-\tau }r_e.
\end{align}
Note that \eqref{eq:Second-SmallStuckRegion-6} can be checked by the same argument as \eqref{eq:Second-SmallStuckRegion-2} and the inductive hypothesis.
\eqref{eq:Second-SmallStuckRegion-8} holds when we take $\Cb$ sufficiently large such that $\frac{3{\Cc}{\Cd}\log \frac{\delta}{\Cb\rho r_e}}{\Cb} \leq \frac14$ holds. 

Combining \eqref{eq:Second-SmallStuckRegion-5} and \eqref{eq:Second-SmallStuckRegion-8}, we can bound the second term of \eqref{eq:Second-SmallStuckRegion-1} as desired, which concludes (a) holds for $t\ge \tau+1$. 

\paragraph{Verifying (b)} 
For $t\geq \tau + 1$, we have
\begin{align}
& w_{t+1} - w_t \\ 
&= 
    (1+\eta \gamma)^{t-\tau+1}r_e\mathsf{e}-\eta \sum_{s=\tau}^{t}(I-\eta H)^{t-s}(\mathrm{d}H^{s}w^{s} + g^{s})
    -
    \left((1+\eta \gamma)^{t-\tau}r_e\mathsf{e}-\eta \sum_{s=\tau}^{t-1}(I-\eta H)^{t-1-s}(\mathrm{d}H_{s}w^{s} + g^{s})\right)
\\ &=   \eta\gamma (1+\eta \gamma)^{t-\tau} r_e \mathsf{e}
    - \eta \sum_{s=\tau}^{t-1}\eta H(I-\eta H)^{t-1-s}(\mathrm{d}H^{s}w^{s} + g^{s})
     -\eta (\mathrm{d}H^{t}w^{t} + g^{t})
    .
\end{align}
As for the first term, we can bound its norm as
\begin{align}
    \|\eta\gamma (1+\eta \gamma)^{t-\tau} r_e \mathsf{e}\|\leq \eta \gamma (1+\eta \gamma)^{t-\tau}r_e.
\end{align}
The norm of the second term can be bounded by using (a) and (b) for $\tau+1,\cdots, t-1$ and \cref{lemma:Murata-MatrixEigenvalue} as follows:
\begin{align}
    &\left\|\eta \sum_{s=\tau}^{t-1}\eta H(I-\eta H)^{t-1-s}(\mathrm{d}H_{s}w_{s} + y_{s})\right\|
    \\ & \leq 
    \sum_{s=\tau}^{t-1}\eta \left\|\eta H(I-\eta H)^{t-1-s}\right\|(\|\mathrm{d}H_{s}\|\|w_{s}\| + \|y_{s}\|)
    \\ & \leq 
    \sum_{s=\tau}^{t-1}\eta \left\|\eta H(I-\eta H)^{t-1-s}\right\|\left(\frac{\delta}{\Cb}(1+\eta\gamma)^{s-\tau}r_e + \frac{3C_1^\frac12{\Cc}\gamma}{\Cb} (1+\eta\gamma)^{s-\tau}r_e\right)
    \\ & \leq 
    \sum_{s=\tau}^{t-1}\eta \left\|\eta H(I-\eta H)^{t-1-s}\right\| \left(\frac{\delta}{\Cb}+\frac{3C_1^\frac12{\Cc}\gamma}{\Cb}\right)(1+\eta\gamma)^{s-\tau}r_e
    \\ & \leq 
    \sum_{s=\tau}^{t-1}\eta \left(\eta \gamma (1+\eta \gamma)^{t-1-s} + \frac{1}{t-s}\right)
        \left(\frac{\delta}{\Cb}+\frac{3C_1^\frac12{\Cc}\gamma}{\Cb}\right)(1+\eta\gamma)^{s-\tau}r_e
    \\ & \leq 
    \eta\left(\eta \gamma T_2 + \log T_2 \right)\left(\frac{\delta}{\Cb}+ \frac{3C_1^\frac12{\Cc}\gamma}{\Cb}\right) (1+\eta \gamma)^{t-\tau}r_e.
\end{align}
Since $T_2 = \tilde{O}\left(\frac{1}{\eta\delta}\right)$ and $\gamma \geq \delta$, setting $\Cb=\tilde{O}(1)$ and $\eta = \tilde{O}\left(\frac1L\right)$ with sufficiently large hidden constants yields $\left(\eta \gamma T_2 + \log T_2 \right)\left(\frac{\delta}{\Cb}+\frac{3C_1^\frac12{\Cc}\gamma}{\Cb} \right)\leq \gamma$. Thus, the second term is bounded by $\eta\gamma (1+\eta \gamma)^{t-\tau}r_e$.

Finally, we consider the third term.
We have $\|\mathrm{d}H^{t}w^{t}\|\leq \frac{\delta}{\Cb}r_e (1+\eta \gamma)^{t-\tau}r_e$ and $\|g^t\|\leq \frac{3C_1^\frac12{\Cc}\gamma}{\Cb}(1+\eta \gamma)^{t-\tau}r_e$ by the inductive hypothesis.
Thus, taking $\Cb$ sufficiently large, the third term is bounded by $\eta\gamma (1+\eta \gamma)^{t-\tau}r_e$.

Combining these bounds, we get (b) for $t+1$.

\paragraph{Verifying (c)}
By using \cref{lemma:Second-Auxiliary-A} and the inductive hypothesis, we have
\begin{align}
    \|g^{t+1}\| 
    & \leq \frac{{\Cc}\zeta r_e}{\sqrt{b}}
    + \frac{{\Cc}\zeta}{\sqrt{b}}\sqrt{\sum_{s=\max\{\tau,t-T_1+1\}}^{t}\|w^s - w^{s-1}\|^2}
    +
    \frac{{\Cc}\delta}{\Cb\sqrt{b}}\sqrt{\sum_{s=\max\{\tau,t-T_1+1\}}^{t}\|w^{s}\|^2}
    \\ & \leq
    \frac{{\Cc} \zeta}{\sqrt{b}}r_e + \frac{3C_1{\Cc} \zeta \sqrt{n}\eta \gamma}{b} (1+\eta\gamma)^{t-\tau}r_e
    + \frac{C_1^\frac12{\Cc} \sqrt{n}\delta}{\Cb b} (1+\eta\gamma)^{t-\tau}r_e
    \\ & \leq
    \left(\frac{{\Cc} \zeta }{\sqrt{b}}+3C_1{\Cc} \zeta \eta \gamma+\frac{C_1^\frac12{\Cc}\delta}{\Cb}\right) (1+\eta\gamma)^{t-\tau}r_e
\end{align}
with probability at least $1-\frac{\nu}{T}$ for all $t$.
Taking $b\geq \frac{\Cb^2\zeta^2}{\delta^2}$, $\eta = \tilde{\Theta}\left(\frac{1}{L}\right)$, and $\Cb = O(C_1{\Cc}) = \tilde{O}(1)$ gives
$\frac{{\Cc} \zeta}{\sqrt{b}}+C_1{\Cc} \zeta \eta \gamma+\frac{C_1^\frac12{\Cc}\delta}{\Cb}\leq \frac{3C_1^\frac12{\Cc}}{\Cb} \gamma$.
Thus, we obtain that (c) holds for $t+1$.

Thus, we complete the induction step, and hence, the assertion follows.
\end{proof}

From \cref{lemma:Small-Stuck-Region}, we can ensure that \AlgMain\ escapes saddle points with high probability.
\begin{lemma}\label{lemma:Escape-Small-Stuck-Region}
Let $\{x^t\}$ be a sequence generated by \AlgMain\ and $\tau_0(\geq 0)$ be a step where $-\gamma :=\lambda_{\rm min}(\nabla^2 f(x^{\tau_0}))\leq -\delta$ holds.
We denote the eigenvector with the eigenvalue $\lambda_{\rm min}(\nabla^2 f(x^{\tau_0}))$ by $\mathsf{e}$.
We take $b\geq \sqrt{n}+\frac{\zeta^2}{\delta^2}$, $\eta = \tilde{\Theta}\left(\frac{1}{L}\right)$, and $T_2=\frac{{\Cd}\log \frac{\delta }{ \Cb \rho r_e}}{\eta \gamma}\lesssim \tilde{O}\left(\frac{L}{\delta}\right)$ with a constant ${\Cd}=O(1)$.
Then, for arbitrary $\tau > \tau_0$, it holds that
\begin{align}
    \mathbb{P}\left[\max_{\tau_0\leq t\leq \tau + T_2}\|x^t-x^{\tau_0}\|\geq \frac{\delta}{\Cb\rho}\mid I^0,\cdots,I^\tau,\xi^1,\cdots,\xi^\tau\right]
    \geq 1-\frac{2\nu}{T},
\end{align}
\end{lemma}
\begin{proof}

Let $A$ be a subset of $B(0,r)$ such that each $a\in A$ satisfies
\begin{align}
    \mathbb{P}\left[\max_{\tau_0\leq t\leq \tau + T_2}\|x^t-x^{\tau_0}\| > \frac{\delta}{\Cb\rho}\mid I^0,\cdots,I^\tau,\xi^1,\cdots,\xi^\tau,\xi^{\tau+1}=a\right]\leq 1-\frac{\nu}{T}
    .
\end{align}
Then, no two elements, $\xi_{\tau+1}$ and $\tilde{\xi}_{\tau+1}$ such that $\xi_{\tau+1}-\tilde{\xi}_{\tau+1}=r_e\mathsf{e}$ with $r_e\geq \frac{r\nu}{T\sqrt{d}}$, can be elements of $A$ at the same time since by \cref{lemma:Small-Stuck-Region}, it holds that
\begin{align}
    \max_{\tau_0\leq t\leq \tau + T_2}\{\|x^t-x^{\tau_0}\|,\|\tilde{x}^t-x^{\tau_0}\|\} \geq \frac{\delta}{\Cb\rho}
\end{align}
with probability at least $1-\frac{\nu}{T}$. 
Let $V_d(r)$ be the volume of Euclidean ball with radius $r$ in $\mathbb{R}^d$. 
Then, we have
\begin{align}
    \frac{{\rm Vol}(A)}{V_d(r)}\leq \frac{r_eV_{d-1}(r)}{V_d(r)}
    =\frac{r_e \Gamma(\frac{d}{2}+1)}{\sqrt{\pi}r\Gamma(\frac{d}{2} + \frac12)}
    \leq \frac{r_e}{\pi r}\left(\frac{d}{2}+1\right)^\frac12
    \leq
    \frac{r_e\sqrt{d}}{r}
    \leq \frac{\nu}{T}.
\end{align}
This means that $A$ occupies at least $1-\frac{\nu}{T}$ of the volumes of $B(0,r)$.
From this fact and the definition of $A$, we have
\begin{align}
\mathbb{P}\left[\max_{\tau_0\leq t\leq \tau + T_2}\|x^t-x^{\tau_0}\|\geq \frac{\delta}{\Cb\rho}\mid I^0,\cdots,I^\tau,\xi^1,\cdots,\xi^\tau\right]
    \geq 1-\frac{\nu}{T}-\frac{\nu}{T}=1-\frac{2\nu}{T},
\end{align}
which gives the conclusion.
\end{proof}

Then, we move to the proof of the main theorem of this subsection, which guarantees that the algorithm finds $(\eps,\delta)$-second-order stationary point with high probability.

\begin{proof}[Proof of \cref{theorem:Second-Main}]
Since $T_2=\frac{{\Cd}\log \frac{\delta }{ \Cb \rho r_e}}{\eta \gamma}$ depends on $x^{\tau_0}$ (since $\gamma$ depends on $\nabla^2f(x^{\tau_0})$), we take $T_2=\frac{{\Cd}\log \frac{\delta }{ \Cb \rho r_e}}{\eta \delta}$ instead from now.
Note that this replacement does not affect whether \cref{lemma:Escape-Small-Stuck-Region} holds.

We devide $\{t=0,1,\cdots,T-1\}$ into $\lceil\frac{T}{2T_2}\rceil$ \textit{phases}: $P^\tau = \{2\tau T_2\leq t < 2(\tau+1)T_2\}\ \left(\tau=0,\cdots,\lceil\frac{T}{2T_2}\rceil-1\right)$.
For each phase, we define $a^\tau$ as a random variable defined by
\begin{align}
    a^\tau = 
    \begin{cases}
    1 & \left(\text{if $\sum_{t\in P^\tau}\mathbbm{1}[\|\nabla f(x^t)\|>\eps]> T_2$}\right),\\
    2 & \left(\text{if there exists $t$ such that $2\tau T_2 \leq t< (2\tau+1)T_2$, $\|\nabla f(x^t)\|\leq \eps$ and $\lambda_{\rm min}(\nabla^2 f(x^t) ) \leq -\delta$}\right),\\
    3 & \left(\text{if there exists $t$ such that $2\tau T_2 \leq t< (2\tau+1)T_2$, $\|\nabla f(x^t)\|\leq \eps$ and $\lambda_{\rm min}(\nabla^2 f(x^t) ) > -\delta$}\right).
    \end{cases}
\end{align}
Note that $\mathbb{P}[a^\tau \in\{ 1, 2, 3\}]=1$ for each $\tau$.
This is because if there does not exist $t$ between $2\tau T_2 \leq t< (2\tau+1)T_2$ such that $\|\nabla f(x^t)\|\leq \eps$ (i.e., neither $a^\tau=2$ nor $3$), then we have $\sum_{t\in P^\tau}\mathbbm{1}[\|\nabla f(x^t)\|>\eps]\geq \sum_{t=2\tau T_2}^{(2\tau+1)T_2-1}\mathbbm{1}[\|\nabla f(x^t)\|>\eps]=T_2$, meaning $a^\tau = 1$.
We denote $N_1 = \sum_{\tau=0}^{\frac{T}{2T_2}-1}\mathbbm{1}[a^\tau = 1]$, $N_2 = \sum_{\tau=0}^{\frac{T}{2T_2}-1}\mathbbm{1}[a^\tau = 2]$, and $N_3 = \sum_{\tau=0}^{\frac{T}{2T_2}-1}\mathbbm{1}[a^\tau = 3]$.

According to \cref{lemma:Escape-Small-Stuck-Region}, with probability $1-2\nu$ over all $\tau$, it holds that if $a^\tau = 2$ then that phase successes escaping saddle points; i.e., there exists $2\tau T_2\leq t<(2\tau+1)T_2$ such that
\begin{align}
    \max_{t\leq s\leq t + T_2}\|x^s-x^t\|> \frac{\delta}{\Cb\rho}
    \label{eq:Second-SmallStuckRegion-31}
\end{align}
holds.
\eqref{eq:Second-SmallStuckRegion-31} further leads to
\begin{align}
    T_2\sum_{t=2\tau T_2}^{2(\tau+1) T_2-1}\|x^{t+1}-x^t\|^2
    > \left(\frac{\delta}{\Cb\rho}\right)^2
    \left(\iff
    \sum_{t=2\tau T_2}^{2(\tau+1) T_2-1}\|x^{t+1}-x^t\|^2
    > \frac{\delta^2}{T_2\Cb^2\rho^2}\right). 
    \label{eq:Second-SmallStuckRegion-32}
\end{align}

On the other hand, in \cref{theorem:First-Main}, we derived that 
\begin{align}
& \sum_{t=1}^T \|\nabla f(x^{t-1})\|^2 \\ & \leq
\begin{cases} \displaystyle
\frac{2}{\eta}\left[\left(f(x^0)-f(x^{t})\right) -
\left(\frac{1}{2\eta}-\frac{L}{2}-\frac{15 C_1^9\eta\zeta^2n}{b^2}\right)\sum_{t=1}^T\|x^t-x^{t-1}\|^2
+\frac{\eta C_1^3T_1}{b}\left(\sigma_c^2+\frac{G_c^2}{b}\right)\right]
+\frac{2Tr^2}{\eta^2}& (\text{Option I})
\\ \displaystyle
\frac{2}{\eta}\left[\left(f(x^0)-f(x^{t})\right) -
\left(\frac{1}{2\eta}-\frac{L}{2}-\frac{15 C_1^9\eta\zeta^2n}{b^2}\right)\sum_{t=1}^T\|x^t-x^{t-1}\|^2
\right]
+\frac{2Tr^2}{\eta^2} & (\text{Option II})
\end{cases}
\end{align}
with probability $1-3\nu$.
By taking $\eta=\tilde{\Theta}\left(\frac1L\right)$, applying $b\geq \sqrt{n}$ and $f(x^0)-f(x^t)\leq \Delta$, and rearranging terms, we obtain
\begin{align}
& \sum_{t=1}^T \|\nabla f(x^{t-1})\|^2 
+\frac{1}{2\eta^2}\sum_{t=1}^T\|x^t-x^{t-1}\|^2
 \leq
\begin{cases} \displaystyle
\frac{2}{\eta}\left[\Delta 
+\frac{12\eta C_1^3 n}{b^2}\left(\sigma_c^2+\frac{G_c^2}{b}\right)\right]
+\frac{2Tr^2}{\eta^2}& (\text{Option I}),
\\ \displaystyle
\frac{2\Delta}{\eta}
+\frac{2Tr^2}{\eta^2} & (\text{Option II}).
\end{cases}
\end{align}
From the definition of $a^\tau = 1$ and \eqref{eq:Second-SmallStuckRegion-32}, that the left-hand side is bounded as
\begin{align}
    \sum_{t=1}^T \|\nabla f(x^{t-1})\|^2 
+\frac{1}{2\eta^2}\sum_{t=1}^T\|x^t-x^{t-1}\|^2
& \geq 
N_1T_2\eps^2
+
\frac{\delta^2N_2}{2\eta^2T_2\Cb^2\rho^2}.
\end{align}
Thus, it holds that 
\begin{align}
\label{eq:Second-SmallStuckRegion-33}
    \max\left\{N_1T_2\eps^2
,
N_2T_2\cdot\frac{\delta^2}{2\eta^2T_2^2\Cb^2\rho^2}\right\}
\le
\begin{cases} \displaystyle
\frac{2}{\eta}\left[\Delta 
+\frac{12\eta C_1^3 n}{b^2}\left(\sigma_c^2+\frac{G_c^2}{b}\right)\right]
+\frac{2Tr^2}{\eta^2}& (\text{Option I}),
\\ \displaystyle
\frac{2\Delta}{\eta}
+\frac{2Tr^2}{\eta^2} & (\text{Option II}).
\end{cases}
\end{align}

By the parameter settings, we have $\frac{2\eta^2T_2^2\Cb^2\rho^2}{\delta^2} = \tilde{O}\left(\frac{\rho^2}{\delta^4}\right)$.
From this, $(N_1 + N_2)T_2 \leq \tilde{O}\left(\frac{1}{\eps^2}+\frac{\rho^2}{\delta^4}\right)\times (\text{the right-hand side of \eqref{eq:Second-SmallStuckRegion-33}})$.
Taking $T\geq 2(N_1+N_2+1)T_2$, there exists $\tau$ such that $a^\tau =3$, which concludes the proof.
\end{proof}
\begin{remark}\label{remark:ReduceDeltaDependency}
Although our main interest in this paper is to develop a simple algorithm with convergence to second-order stationary points, it can be easily shown that adaptive selection of minibatch size can reduce the gradient complexity.
In \cref{lemma:Small-Stuck-Region}, if we carefully check the proof, we can see that the condition $b\gtrsim \sqrt{n}+\frac{\zeta^2}{\delta^2}$ is needed only for the step $\tau$. On the other hand, for all $\tau_0\leq t\leq \tau+T_2$ except for $t=\tau$, $b=\sqrt{n}$ is sufficient.


If we take $b = \sqrt{n}+\frac{\zeta^2}{\delta^2}$ only at $t=(2\tau + 1)T_2~\left(\tau=0,\cdots,\frac{T}{2T_2}-1\right)$ and $b =\sqrt{n}$ at the other steps, the above argument still holds with a slight modification. 
Then, 
the gradient complexity is reduced to 
\begin{align}
    \tilde{O}\left(\left(L\Delta + \sigma_c^2+\frac{G_c^2}{b}\right)\left(\frac{\sqrt{n}}{\eps^2}+\frac{\rho^2\sqrt{n}}{\delta^4}+\frac{\zeta^2}{L\eps^2\delta}+\frac{\zeta^2\rho^2}{L\delta^5}\right)\right)
    \quad (\text{\rm Option I}),
    \\ 
    \tilde{O}\left(n+L\Delta\left(\frac{\sqrt{n}}{\eps^2}+\frac{\rho^2\sqrt{n}}{\delta^4}+\frac{\zeta^2}{L\eps^2\delta}+\frac{\zeta^2\rho^2}{L\delta^5}\right)\right)
    \quad (\text{\rm Option II})  .
\end{align}
In the classical setting $\delta = O(\sqrt{\rho\eps})$, this bound is no worse than 
SPIDER-$\text{SFO}^+$(+Neon2) \citep{fang2018spider,allen2018neon2}, no matter what $n$ and $\delta$ are.

Finally, we note that if $\delta$ is too small,  $\frac{L^2}{\delta^2}$ can be as large as $n$.
In such case, it is more efficient to replace sampling such number of samples is replaced by full gradient computation. 
Then, the complexity gets
\begin{align}
    \tilde{O}\left(\left(L\Delta + \sigma_c^2+\frac{G}{b}\right)\left(\frac{\sqrt{n}}{\eps^2}+\frac{\rho^2\sqrt{n}}{\delta^4}+\frac{n\delta}{L\eps^2}+\frac{n\rho^2}{L\delta^3}\right)\right)
    \quad (\text{\rm Option I}),
    \\ 
    \tilde{O}\left(n+L\Delta\left(\frac{\sqrt{n}}{\eps^2}+\frac{\rho^2\sqrt{n}}{\delta^4}+\frac{n\delta}{L\eps^2}+\frac{n\rho^2}{L\delta^3}\right)\right)
    \quad (\text{\rm Option II})  .
\end{align}
When $\delta = O(\sqrt{\rho\eps})$, this bound is no worse than NestedSVRG+Neon2 \citep{zhou2020stochastic,allen2018neon2}.
However, it is unusual to assume $\frac{L^2}{\delta^2}=n$ in the first place.
In fact, carefully looking the proof of SSRGD \citep{li2019ssrgd}, we find that they implicitly limits their analysis to the case of $\frac{L^2}{\delta^2}\lesssim n$.
\end{remark}

\subsection{Convergence under PL condition (proof of \cref{theorem:PL-Main})}
In this subsection, we provide the proof of the convergence under \cref{assumption:PL}, i.e., PL condition holds for the objective function.
\begin{proof}
According to the descent lemma (\cref{lemma:DescentLemma}) and PL condition (\cref{assumption:PL}), we have that 
\begin{align}
    f(x^t)
    & \leq
    f(x^{t-1}) + \eta \|\nabla f(x^{t-1}) - v^{t-1}\|^2
    - \frac{\eta}{2}\|\nabla f(x^{t-1})\|^2
    - \left(\frac{1}{2\eta}-\frac{L}{2}\right)\|x^t-x^{t-1}\|^2
    + \frac{r^2}{\eta}
    \\ & \leq 
    f(x^{t-1}) + \eta \|\nabla f(x^{t-1}) - v^{t-1}\|^2
    - \eta\mu(f(x^{t-1})-f(x^*))
    - \left(\frac{1}{2\eta}-\frac{L}{2}\right)\|x^t-x^{t-1}\|^2
    + \frac{r^2}{\eta} .   
\end{align}
Rearranging the terms yields
\begin{align}
    f(x^t)-f^*
    \leq (1-\eta\mu)(f(x^{t-1})-f^*) + \eta \|\nabla f(x^{t-1}) - v^{t-1}\|^2    - \left(\frac{1}{2\eta}-\frac{L}{2}\right)\|x^t-x^{t-1}\|^2
    + \frac{r^2}{\eta}
    .
\end{align}

By applying \cref{lemma:First-VarianceBound} to this, we obtain that with probability at least $1-3\nu$,
\begin{align}
    f(x^t)-f^*
    &\leq  (1-\eta\mu)(f(x^{t-1})-f^*)     - \left(\frac{1}{2\eta}-\frac{L}{2}\right)\|x^t-x^{t-1}\|^2
    + \frac{r^2}{\eta}
    \\ & \hspace{5mm}+         \begin{cases}
    \displaystyle
        \frac{15C_1^8\eta\zeta^2}{b}\sum_{s =\max\{1,t-T_1+1\}}^{t}\|x^s -x^{s -1}\|^2
        +
        \frac{12C_1^2\eta\mathbbm{1}[t< T_1]}{b}\cdot \left(\sigma_c^2 + \frac{G_c^2}{b}\right)
    &\quad (\text{Option I})\\\displaystyle
    \frac{15C_1^8\eta\zeta^2}{b}\sum_{s =\max\{1,t-T_1+1\}}^{t}\|x^s -x^{s -1}\|^2
&\quad (\text{Option II})
    \end{cases}
\end{align}
holds for all $t=1,\cdots,T$. 
Multiplying both sides by $(1-\eta\mu)^{T-t}$ and summing up over all $t=1,2,\cdots,T$ and arranging the terms, we get
\begin{align}
    &f(x^T)-f^*\\ &\leq
    (1-\eta \mu)^T(f(x^0)-f^*)+\sum_{t=1}^T(1-\eta \mu)^{T-t}\frac{r^2}{\eta}
        -\sum_{t=1}^T(1-\eta \mu)^{T-t}\left(\frac{1}{2\eta}-\frac{L}{2}-\frac{15C_1^9\eta\zeta^2n(1-\eta \mu)^{-T_1}}{b^2}\right)\|x^t-x^{t-1}\|^2
    \\ & 
    \hspace{5mm}+\begin{cases}\displaystyle
    (1-\eta \mu)^{T-T_1}\frac{12C_1^3\eta n}{b^2}\left(\sigma_c^2 \frac{G_c^2}{b}\right)
    &(\text{Option I})
    \\ \displaystyle
    0
    &(\text{Option II}).
    \end{cases}
\end{align}
Note that $T_1 = \frac{n}{b}C_1$. According to this, we take $\eta$ as
\begin{align}
    \eta = \Theta\left(\frac{1}{L}\land \frac{b}{C_1^{4.5}\zeta\sqrt{n}}\land\frac{b}{\mu C_1n}\right)
\end{align}
so that 
$\frac{1}{2\eta}-\frac{L}{2}-\frac{15C_1^9\eta\zeta^2n(1-\eta \mu)^{-T_1}}{b^2}\geq 0$ holds.
Then, we have that
\begin{align}
    f(x^t)-f^* \leq
    (1-\eta \mu)^T(f(x^0)-f^*)+\sum_{t=1}^T(1-\eta \mu)^{T-t}\frac{r^2}{\eta}
    +\begin{cases}\displaystyle
    (1-\eta \mu)^{T-T_1}\frac{12C_1^3\eta n}{b^2}\left(\sigma_c^2 + \frac{G_c^2}{b}\right)
    &\quad (\text{Option I})
    \\ \displaystyle
    0
    &\quad (\text{Option II})
    \end{cases}
\end{align}
The first term $(1-\eta \mu)^T(f(x^0)-f^*)$ is smaller than $\frac{\eps}{3}$ if we take $T = O \left(\frac{1}{\eta\mu}\log \frac{\Delta}{\eps}\right)$.
The second term is bounded by $\frac{r^2}{\eta^2\mu}$, which is smaller than $\frac{\eps}{3}$ if we take $r\leq \eta\sqrt{\frac{\eps\mu}{3}}$.
The third term for Option I, $(1-\eta \mu)^{T-T_1}\frac{12\eta C_1^3n}{b^2}\left(\sigma_c^2 + \frac{G}{b}\right)$, is also bounded by $\frac{\eps}{3}$, if we take 
$T = T_1+ O \Bigl(\frac{1}{\eta\mu}\log \frac{\frac{C_1^3 \eta n}{b^2} \left(\sigma_c^2+\frac{G_c^2}{b}\right)}{\eps}\Bigr)=O\left(\frac{n}{b}C_1+ \frac{C_1}{\eta\mu}\log \frac{\left(\sigma_c+G_c\right)}{\eps}\right)$. 

Thus, for Option I, taking 
 \begin{align}T=O^*\left(\frac{n}{b}C_1 + C_1\left(\frac{L}{\mu}\lor \frac{C_1^{4.5}\zeta \sqrt{n}}{\mu b}\lor \frac{C_1n}{b}\right)\log \frac{\Delta+\sigma_c+G_c}{\eps}\right),
\end{align} yields the desired bound with probability at least $1-3\nu$.

And for Option II, taking
\begin{align}T=O\left( \left(\frac{L}{\mu}\lor \frac{C_1^{4.5}\zeta \sqrt{n}}{\mu b}\lor \frac{C_1n}{b}\right)\log \frac{\Delta}{\eps}\right)
\end{align}
yields the desired bound.

Note that $T$ depends on $\eps^{-1}$ only logarithmically, which means that $C_1$ depends on $\eps^{-1}$ in only $\log\log$ order and $C_1 = O^*(\log(n+\mu^{-1}+\nu^{-1}))$, where $O^*$ suppresses $\log\log$ factors.
\end{proof}

\section{Missing Statements and Proofs for \AlgDist}\label{section:Appendix:FLEDGE} 
This Section provides the missing information of \AlgDist\ that we abbreviate in \cref{section:FLEDGE} and gives the proofs of the theorems in \cref{section:FLEDGE} about the convergence property of \AlgDist. 
First, we provide the full version of FLEDGE, including Option I.

\begin{algorithm}[ht]
    \caption{\AlgDist$(x^0, \eta, p, b, T, K ,r)$ (formal)}
    \begin{algorithmic}[1]
        \State{{\bf Option I:}}
        \State{\hspace{5mm}Randomly select $p$ agents $I^{0}$}
        \State{\hspace{5mm}{\bf for }$i\in I^0$ in parallel {\bf do}}
        \State{\hspace{10mm}Randomly select $bK$ samples $J^0_{i}$}
        \State{\hspace{10mm}$y^0_i\leftarrow \frac{1}{bK}\sum_{j\in J^0_{i}}\nabla f_{i,j}(x^0)$}
        \State{\hspace{5mm}Communicate $\{y^{0}_i\}_{i\in I^{0}}$ between $I^{0}$}
        \State{\hspace{5mm}$y^0_i \leftarrow \frac{1}{p}\sum_{i\in I^0}y_i^0\ (i=1,\cdots,n)$ \footnotesize // we do not need to explicitly communicate this between all the clients} 
        \State{{\bf Option II:}}
        \State{\hspace{5mm}{\bf for }$i\in I^0 = I$ in parallel {\bf do}}
        \State{\hspace{10mm}Randomly select $bK$ samples $J_i^0$}
        \State{\hspace{10mm}$y^0_i\leftarrow \frac{1}{bK}\sum_{j\in J_i^0} \nabla f_{i,j}(x^0)$}
            \For{$t=1$ to $T$}
            \State{Randomly sample one agent $i_t$}
            \State{Communicate $\{\frac1P\sum_{i=1}^Py_i^{t-1}\}$, 
            and $x^{t-1}$ between $I^{t-1}\cup \{i^t\}$ and the server}
            \State{$x^{t,0}\leftarrow x^{t-1},\ z^{t,0}\leftarrow 0$}
            \For{$k=1$ to $K$}
            \State{$\displaystyle x^{t,k} \leftarrow x^{t,k-1} - \left(\frac{1}{P} \sum_{i=1}^P y^{t-1}_i + z^{t,k-1}\right)+\xi^{t,k}\quad (\xi^{t,k}\sim B(0,r))$}
            \State{randomly select $b$ samples $J^{t,k}_{i_t}$}
            \State{$z^{t,k} \leftarrow z^{t,k-1} + \frac{1}{b} \sum_{j\in J^{t,k}_{i_t}} (\nabla f_{i_t,j}(x^{t,k}) - \nabla f_{i_t,j} (x^{t,k-1}))$}
            \EndFor
            \State{$x^{t}\leftarrow x^{t,K}$}
            \State{Randomly select $p$ agents $I^{t}$}
            \State{Communicate $x^t$ between $I^{t}\cup\{i_t\}$}
            \For{$i\in I^t$ in parallel}
            \State{Randomly select $b$ samples $J^{t}_i$}
            \State{$y^t_{i} \leftarrow \frac{1}{bK}\sum_{j\in J^{t}_i}\nabla f_{i,j}(x^t)$}
            \State{$\Delta y^t_{i} \leftarrow \frac{1}{bK}\sum_{j\in J^{t}_{i}}(\nabla f_{i,j}(x^t)-f_{i,j}(x^{t-1}))$}
            \EndFor
            \State{Communicate $\{\Delta h_i^t\}_{i\in I^t}$ between $I^t$ and the server}
            \State{$y^t_{i} \leftarrow y^{t-1}_{i} + \frac{1}{p}\sum_{i\in I^t}\Delta y^t_i\ (\text{for }i\notin I^t)$ \footnotesize // Practically, we update only $\frac1P\sum_{i=1}^Py_i^t$ in the server in $O(p)$ time.}
            \EndFor
    \end{algorithmic}
    \label{alg:dist-Appendix}
\end{algorithm}

\subsection{Finding First-order Stationary Points (Proof of \cref{theorem:Dist-First-Main})}
In this subsection, we show that \AlgNumDist finds first-order stationary points with high probability. 
First, we describe the formal statement of \cref{theorem:Dist-First-Main}.

\newtheorem*{theorem:FLEDFEfirst}{\rm\bf Theorem~\ref{theorem:Dist-First-Main}'}
\begin{theorem:FLEDFEfirst}
\label{theorem:Dist-First-Main-Appendix}
Let $r\leq \frac{\eta\eps}{2\sqrt{2}}$ and 
$PKb \geq \tilde{\Omega}\left(\frac{\sigma^2}{\eps^2}+\frac{G}{\eps}\right)$. 
Under \cref{assumption:GradLipschitz,assumption:GlobalInfimum,assumption:BoundedGradient,assumption:Heterogeneity}, if we choose
\begin{align}
    \eta =\tilde{\Theta}\left(\frac{1}{L}\land \frac{p\sqrt{b}}{\zeta\sqrt{PK}}\land \frac{1}{\zeta K}\land \frac{\sqrt{b}}{L\sqrt{K}}\right),
\end{align}
\AlgNumDist with Option I finds an $\eps$-first-order stationary point for problem \eqref{eq:Intro-Federated} by using

\begin{align}
   \tilde{O}\left(
    \left(L\lor\frac{\zeta\sqrt{P}K}{p}\lor \frac{\zeta\sqrt{PK}}{p\sqrt{b}}\lor \zeta K\lor \frac{L\sqrt{K}}{\sqrt{b}}\right)\frac{\Delta pb}{\eps^2}\land
    \left(\frac{\sigma^2P}{p^2b}\lor \frac{PG^2}{p^3Kb^2}\lor\frac{PK\sigma^2_c}{p^2} \lor \frac{PKG^2_c}{p^3}\right)\frac{pb}{\eps^2}
    \right)
   \text{ stochastic gradients and }\\
   \tilde{O}\left(
    \left(\frac{L}{K}\lor\frac{\zeta\sqrt{P}}{p}\lor \frac{\zeta\sqrt{P}}{p\sqrt{Kb}}\lor \zeta \lor \frac{L}{\sqrt{Kb}}\right)\frac{\Delta}{\eps^2}\land
    \left(\frac{\sigma^2P}{p^2Kb}\lor \frac{PG^2}{p^3K^2b^2}\lor\frac{P\sigma^2_c}{p^2} \lor \frac{PG^2_c}{p^3}\right)\frac{1}{\eps^2}
    \right)
   \text{ communication rounds}
\end{align}
with probability at least $1-8\nu$.

Moreover, under the same assumptions, 
\AlgNumDist with Option II finds an $\eps$-first-order stationary point for problem \eqref{eq:Intro-Federated} by using
\begin{align}
   \tilde{O}\left(PKb + 
    \left(L\lor\frac{\zeta\sqrt{P}K}{p}\lor \frac{\zeta\sqrt{PK}}{p\sqrt{b}}\lor \zeta K\lor \frac{L\sqrt{K}}{\sqrt{b}}\right)\frac{\Delta pb}{\eps^2}
    \right)
   \text{ stochastic gradients and }\\
   \tilde{O}\left(
    1+\left(\frac{L}{K}\lor\frac{\zeta\sqrt{P}}{p}\lor \frac{\zeta\sqrt{P}}{p\sqrt{Kb}}\lor \zeta \lor \frac{L}{\sqrt{Kb}}\right)\frac{\Delta}{\eps^2}
    \right)
   \text{ communication rounds}
\end{align}
with probability at least $1-8\nu$.
\end{theorem:FLEDFEfirst}
Let $v^{t,k-1}=\frac1P\sum_{i=1}^P y_i^{t-1}+z^{t,k-1}$ and $K(t)$ be the last inner loop step in the $t$-th outer loop as stated in the algorithm.
The descent lemma (\cref{lemma:DescentLemma}) also works here: as was discussed for Algorithm 1, for each $t$ and $k$ $(1\leq t\leq T, 1\leq k\leq K(t))$, it holds that 
\begin{align}
\label{eq:Dist-First-FunctionDecrease-1}
    f(x^{t,k})\leq
    f(x^{t,k-1}) + \eta \|\nabla f(x^{t,k-1}) - v^{t,k-1}\|^2
    - \frac{\eta}{2}\|\nabla f(x^{t,k-1})\|^2
    - \left(\frac{1}{2\eta}-\frac{L}{2}\right)\|x^{t,k}-x^{t,k-1}\|^2
    + \frac{r^2}{\eta}.
\end{align}
Our strategy is to bound the variance term $\|v^{t,k} - \nabla f(x^{t,k})\|^2$ with high probability, as summarized in the following lemma.
\begin{lemma}\label{lemma:Dist-First-VarianceBound}
Let $v^{t,k}=\frac1P\sum_{i=1}^P y_i^{t-1}+z^{t,k}$ and all the other variables be as stated in \cref{alg:dist-Appendix}.
Then, with taking $T_3 = \frac{P}{p}C_1$, we have
\begin{align}
    &\left\|v^{t,k} - \nabla f(x^{t,k}) \right\|^2
    \\ & \leq 
    \left(\frac{120C_1^8\zeta^2K}{p}+\frac{32C_1^{10}\zeta^2}{pb}\right) \sum_{s=\max\{1,t-T_3\}}^{t-1}\sum_{l=1}^K\|x^{s,l}-x^{s,l-1}\|^2
    +
    \left(4\zeta^2K+\frac{4C_1^2L^2}{b}\right)\sum_{l=1}^k\|x^{t,l}-x^{t,l-1}\|^2
    \\ & \quad +
    \frac{8C_1^2}{PKb}\left(\sigma^2 + \frac{G^2}{PKb}\right)+\begin{cases}
    \displaystyle
         96C_1^2\mathbbm{1}[t\leq T_3]\left(\frac{\sigma^2}{pKb} + \frac{G^2}{p^2K^2b^2}+\frac{\sigma^2_c}{p} + \frac{G^2_c}{p^2}\right) &\quad (\text{\rm Option I})
    \\\displaystyle
    0&\quad (\text{\rm Option II})
    \end{cases}
\end{align}
for all $t,k$ $(1\leq t\leq T, 0\leq k\leq K-1)$, with probability at least $1-8\nu$.
\end{lemma}
For the proof of \cref{lemma:Dist-First-VarianceBound}, we utilize the four following auxiliary lemmas. 
Below, we define $\tilde{y}^0_i$ by
\begin{align}
    \tilde{y}^0_i \coloneqq 
    \begin{cases} 
    \frac{1}{p}\sum_{i\in I^0}\nabla f_i(x^0) & \quad (\text{Option I}),
    \\ \nabla f_i(x^0) & \quad (\text{Option II}).
    \end{cases}
\end{align}
As well as the previous section, we define $\tilde{I}_s^t \coloneqq [n]\setminus \bigcup_{\tau = s}^t I^t$ for $1\leq s\leq t$. 
In addition, for each $s, t\ (s\leq t)$ and $i\in [P]$, we let $T_4(t,i)$ as $T_4(t,i):=\max\{s\mid s=0 \text{ or } 1\leq s\leq t \text{ with } s\in I^s\}$, i.e., the last step when $y_i^s$ is updated before $t$. 
We remark that the setting $T_3 = \frac{P}{p}C_1$ gives $\tilde{I}^t_s=\emptyset$ with probability at least $1-\nu$ for all $t$ and $s\leq t-T_3$.
\begin{lemma}\label{lemma:Dist-First-Auxiliary-D}
With probability at least $1-\nu$, the following holds for all $t=1,\cdots,T$:
\begin{align}
    \left\|\sum_{s=\max\{1,t-T_3+1\}}^t \frac{|\tilde{I}^t_s|}{PpKb}\sum_{i\in I^s} \sum_{j\in  
    J^s_i}(\nabla f_{i,j}(x^s)-\nabla f_{i,j}(x^{s-1})-(\nabla f_i(x^{s})-\nabla f_i(x^{s-1})))\right\|^2
    \\ \leq \frac{4C_1^{10}\zeta^2}{pKb} \sum_{s=\max\{1,t-T_3+1\}}^t\|x^s-x^{s-1}\|^2.
\end{align}
\end{lemma}

\begin{proof}
First, we decompose the left hand side as 
\begin{align}
    &\left\|\sum_{s=\max\{1,t-T_3+1\}}^t \frac{|\tilde{I}^t_s|}{PpKb}\sum_{i\in I^s} \sum_{j\in 
    J^s_i}(\nabla f_{i,j}(x^s)-\nabla f_{i,j}(x^{s-1})-(\nabla f_i(x^{s})-\nabla f_i(x^{s-1})))
    \right\|^2
    \\ & 
    \leq 2\left\|\sum_{s=\max\{1,t-T_3+1\}}^t \frac{\mathbb{E}[|\tilde{I}^t_s|]}{PpKb}\sum_{i\in I^s} \sum_{j\in  
    J^s_i}(\nabla f_{i,j}(x^s)-\nabla f_{i,j}(x^{s-1})-(\nabla f_i(x^{s})-\nabla f_i(x^{s-1})))\right\|^2   
    \\ & \hspace{5mm} +
    2\left\|\sum_{s=\max\{1,t-T_3+1\}}^t \frac{|\tilde{I}^t_s|-\mathbb{E}[|\tilde{I}^t_s|]}{PpKb}\sum_{i\in I^s} \sum_{j\in  
    J^s_i}(\nabla f_{i,j}(x^s)-\nabla f_{i,j}(x^{s-1})-(\nabla f_i(x^{s})-\nabla f_i(x^{s-1})))\right\|^2 
    \\ & 
    \leq 2\left\|\sum_{s=\max\{1,t-T_3+1\}}^t \frac{\mathbb{E}[|\tilde{I}^t_s|]}{PpKb}\sum_{i\in I^s} \sum_{j\in  
    J^s_i}(\nabla f_{i,j}(x^s)-\nabla f_{i,j}(x^{s-1})-(\nabla f_i(x^{s})-\nabla f_i(x^{s-1})))\right\|^2   
    \\ & \hspace{5mm} +
    2\sum_{s=\max\{1,t-T_3+1\}}^t \frac{T_3\left(|\tilde{I}^t_s|-\mathbb{E}[|\tilde{I}^t_s|]\right)^2}{P^2p^2K^2b^2}\left\|\sum_{i\in I^s} \sum_{j\in  
    J^s_i}(\nabla f_{i,j}(x^s)-\nabla f_{i,j}(x^{s-1})-(\nabla f_i(x^{s})-\nabla f_i(x^{s-1})))\right\|^2 
    \\  \label{eq:Dist-First-Auxiliary-D-4}
    \begin{split}& 
    = 2\left\|\sum_{s=\max\{1,t-T_3+1\}}^t \frac{\mathbb{E}[|\tilde{I}^t_s|]}{PpKb}\sum_{i\in I^s} \sum_{j\in  
    J^s_i}(\nabla f_{i,j}(x^s)-\nabla f_{i,j}(x^{s-1})-(\nabla f_i(x^{s})-\nabla f_i(x^{s-1})))\right\|^2   
    \\ & \hspace{5mm} +
    2\sum_{s=\max\{1,t-T_3+1\}}^t \frac{C_1\left(|\tilde{I}^t_s|-\mathbb{E}[|\tilde{I}^t_s|]\right)^2}{Pp^3K^2b^2}\left\|\sum_{i\in I^s} \sum_{j\in  
    J^s_i}(\nabla f_{i,j}(x^s)-\nabla f_{i,j}(x^{s-1})-(\nabla f_i(x^{s})-\nabla f_i(x^{s-1})))\right\|^2.
    \end{split}
\end{align}

To bound the first term, by applying \cref{proposition:Bernstein} to the choice of $I^s$ and $J_i^s$, we have
\begin{align}
    \left\|\sum_{i\in I^s}\sum_{j\in  
    J^s_i}(\nabla f_{i,j}(x^s)-\nabla f_{i,j}(x^{s-1})-(\nabla f_i(x^{s})-\nabla f_i(x^{s-1})))\right\|^2\leq C_1^4\zeta^2pKb\|x^s-x^{s-1}\|^2.
    \label{eq:Dist-First-Auxiliary-D-1}
\end{align}
with probability at least $1-\frac{\nu}{4T^2}$.
Then, we use \cref{proposition:Azuma} to obtain
\begin{align}
    \left\|\sum_{s=\max\{1,t-T_3+1\}}^t \frac{\mathbb{E}[|\tilde{I}^t_s|]}{PpKb}\sum_{i\in I^s}\sum_{j\in  
    J^s_i}(\nabla f_{i,j}(x^s)-\nabla f_{i,j}(x^{s-1})-(\nabla f_i(x^{s})-\nabla f_i(x^{s-1})))\right\|^2
    &\leq \frac{C_1^6\zeta^2}{pKb}\|x^s-x^{s-1}\|^2
    \\&\leq 
    \frac{C_1^6\zeta^2}{pKb}\|x^s-x^{s-1}\|^2
    \label{eq:Dist-First-Auxiliary-D-2}
\end{align}
with probability at least $1-\frac{\nu}{4T}-T \cdot\frac{\nu}{4T^2}=1-\frac{\nu}{2T}$.


For the second term, following the same argument in \cref{lemma:First-Auxiliary-B}, we can show that $|\tilde{I}_s^t-\mathbb{E}[\tilde{I}_s^t]|^2\leq C_1^5 P$ with probability at least $1-\frac{\nu}{4T^2}$ for each $s,t$.
Combining this with \eqref{eq:Dist-First-Auxiliary-D-1}, we have
\begin{align}
    \sum_{s=\max\{1,t-T_3+1\}}^t \frac{||\tilde{I}^t_s|-\mathbb{E}[|\tilde{I}^t_s|]|^2C_1}{Pp^3K^2b^2}\left\|\sum_{i\in I^s} \sum_{j\in  
    J^s_i}(\nabla f_{i,j}(x^s)-\nabla f_{i,j}(x^{s-1})-(\nabla f_i(x^{s})-\nabla f_i(x^{s-1})))\right\|^2 
    \\ \leq
    \sum_{s=\max\{1,t-T_3+1\}}^t \frac{C_1^{10}\zeta^2P}{p^2Kb}\|x^s-x^{s-1}\|^2
    \label{eq:Dist-First-Auxiliary-D-3}
\end{align}
with probability at least $1-T\cdot \frac{\nu}{4T^2}-T\cdot \frac{\nu}{4T^2}=1-\frac{\nu}{2T}$.

Finally, substituting \eqref{eq:Dist-First-Auxiliary-D-2} and \eqref{eq:Dist-First-Auxiliary-D-3} for \eqref{eq:Dist-First-Auxiliary-D-4}, we obtain the assertion.
\end{proof}

\begin{lemma}\label{lemma:Dist-First-Auxiliary-E}
With probability at least $1-\nu$, the following holds for all $t=1,\cdots,T$:
\begin{align}
\left\|\frac{1}{PbK} \sum_{i=1}^P \mathbbm{1}[T_4(t,i)\geq t-T_3]\sum_{j\in J_i^{T_4(t,i)}}(\nabla f_{i,j}(x^{T_4(t,i)}) - \nabla f_{i}(x^{T_4(t,i)}))\right\|^2
\leq \frac{C_1^2}{PKb}\left(\sigma^2+\frac{G^2}{PKb}\right).
\end{align}
\end{lemma}
\begin{proof}
We condition the events on $\{T_4(i,s)\}$ and apply the Bernstein's inequality to obtain the desired bound. 
\end{proof}

\begin{lemma}\label{lemma:Dist-First-Auxiliary-C}
With probability at least $1-\nu$, the following holds for all $t=1,\cdots,T$:
\begin{align}
        \left\|\frac{\mathbbm{1}[t\leq T_3]}{P}\sum_{i\in \tilde{I}_1^t}(y^0_i-\tilde{y}^0_i)\right\|^2\leq 
        \begin{cases}\displaystyle
        \frac{C_1^2\mathbbm{1}[t\leq T_3]}{pKb}\left(\sigma^2 + \frac{G^2}{pKb}\right)
        & (\text{\rm Option I})
        \\ \displaystyle
        \frac{C_1^2}{PKb}\left(\sigma^2 + \frac{G^2}{PKb}\right)
        & (\text{\rm Option II})
        \end{cases}
\end{align}
\end{lemma}

\begin{proof} 

Recall the definition of $\tilde{y}_i^0$: 
\paragraph{Option I}
By conditioning $I^0$, \cref{proposition:Bernstein} yields that
\begin{align}
    \left\|\frac{\mathbbm{1}[t\leq T_3]}{P}\sum_{i\in \tilde{I}_1^t}(y^0_i-\tilde{y}^0_i)\right\|^2
    =
    \left\|\frac{|\tilde{I}_1^t|\mathbbm{1}[t\leq T_3]}{P}
    \cdot \frac{1}{pKb}\sum_{i\in I^0}\sum_{j\in J_i^0}(\nabla f_{i,j}(x^0)-\nabla f_{i}(x^0))
    \right\|^2
    \leq \frac{C_1^2\mathbbm{1}[t\leq T_3]}{pKb}\left(\sigma^2 + \frac{G^2}{pKb}\right),
\end{align}
with probability at least $1-\frac{\nu}{T}$ for each $t$.

\paragraph{Option II}
In this case, \cref{proposition:Bernstein} directly yields that
\begin{align}
    \left\|\frac{\mathbbm{1}[t\leq T_3]}{P}\sum_{i\in \tilde{I}_1^t}(y^0_i-\tilde{y}^0_i)\right\|^2
    =
    \left\|\frac{|\tilde{I}_1^t|\mathbbm{1}[t\leq T_3]}{P}
    \cdot \frac{1}{PKb}\sum_{i\in I}\sum_{j\in J_i^0}(\nabla f_{i,j}(x^0)-\nabla f_{i}(x^0))
    \right\|^2
    \leq \frac{C_1^2}{PKb}\left(\sigma^2 + \frac{G^2}{PKb}\right),
\end{align}
with probability at least $1-\frac{\nu}{T}$ for each $t$.

\end{proof}

\begin{lemma}\label{lemma:Dist-First-Auxiliary-F}
With probability at least $1-\nu$, the following holds for all $t=1,\cdots,T-1$ and $k=1,\cdots,K-1$:
\begin{align}
    \|z^{t,k} - (\nabla f(x^{t,k})-\nabla f(x^{t,0}))\|^2
    \leq \left(2\zeta^2K+\frac{2C_1^2L^2}{b}\right)\sum_{l=1}^k\|x^{t,l}-x^{t,l-1}\|^2
\end{align}
\end{lemma}

\begin{proof}
We decompose the $z^{t,k} - (\nabla f(x^{t,k})-\nabla f(x^{t,0}))$ as
\begin{align}
    &\|z^{t,k} - (\nabla f(x^{t,k})-\nabla f(x^{t,0}))\|^2
    \\ & \leq
    2\|z^{t,k}-(\nabla f_{i_t}(x^{t,k})-\nabla f_{i_t}(x^{t,0}))\|^2
    +
    2\|\nabla f_{i_t}(x^{t,k})-\nabla f_{i_t}(x^{t,0})-(\nabla f(x^{t,k})-\nabla f(x^{t,0}))\|^2
    \\ & \leq
    2\left\| \sum_{l=1}^k\frac{1}{b} \sum_{j\in J^{t,l}_{i_t}} (\nabla f_{i_t,j}(x^{t,l}) - \nabla f_{i_t,j} (x^{t,l-1}))\right\|^2
    + 2\zeta^2\|x^{t,k}-x^{t,0}\|^2
    \\ & \leq
    2\left\| \sum_{l=1}^k\frac{1}{b} \sum_{j\in J^{t,l}_{i_t}} (\nabla f_{i_t,j}(x^{t,l}) - \nabla f_{i_t,j} (x^{t,l-1}))\right\|^2
    + 2\zeta^2K\sum_{l=1}^k\|x^{t,l}-x^{t,l-1}\|^2,
    \label{eq:Dist-First-Auxiliary-F-1}
\end{align}
where we use \cref{assumption:Heterogeneity} for the second inequality.

We apply \cref{proposition:Bernstein} to $\frac{1}{b} \sum_{j\in J^{t,l}_{i_t}} (\nabla f_{i_t,j}(x^{t,l}) - \nabla f_{i_t,j} (x^{t,l-1}))$ and obtain that
\begin{align}
     \left\|\frac{1}{b} \sum_{j\in J^{t,l}_{i_t}} (\nabla f_{i_t,j}(x^{t,l}) - \nabla f_{i_t,j} (x^{t,l-1}))\right\|^2
     \leq \frac{C_1^2L^2}{b}\|x^{t,l}-x^{t,l-1}\|^2
\end{align}
with probability at least $1-\frac{1}{2TK^2}$ for each $(t,l)$.
Using \cref{proposition:Azuma}, with probability $1-K\cdot\frac{1}{2TK^2}-\frac{1}{2TK}=1-\frac{1}{TK}$, we have
\begin{align}
    \left\| \sum_{l=1}^k\frac{1}{b} \sum_{j\in J^{t,l}_{i_t}} (\nabla f_{i_t,j}(x^{t,l}) - \nabla f_{i_t,j} (x^{t,l-1}))\right\|^2
    \leq \frac{C_1^4L^2}{b}\sum_{l=1}^k\|x^{t,l}-x^{t,l-1}\|^2
    \label{eq:Dist-First-Auxiliary-F-2}
\end{align}
for each $(t,k)$.

By substituting \eqref{eq:Dist-First-Auxiliary-F-2} to \eqref{eq:Dist-First-Auxiliary-F-1}, we obtain the desired bound.
\end{proof}

\begin{proof}[Proof of \cref{lemma:Dist-First-VarianceBound}]
First, we observe that
\begin{align}
    \label{eq:DistAuxiliary100}
    \begin{split}
    \left\|\frac1P \sum_{i=1}^P y^{t-1}_i + z^{t,k} - \nabla f(x^{t,k})\right\|^2
 &= \left\|\frac1P \sum_{i=1}^P y^{t-1}_i - \nabla f(x^{t-1}) + z^{t,k} - (\nabla f(x^{t,k})-\nabla f(x^{t,0}))\right\|^2\\
 &\le 2\left\|\frac1P \sum_{i=1}^P y^{t-1}_i - \nabla f(x^{t-1})\right\|^2+2\left\|\nabla f(x^{t-1}) + z^{t,k} - (\nabla f(x^{t,k})-\nabla f(x^{t,0}))\right\|^2
    \end{split}
\end{align}
 
We first bound $\left\| \frac1P \sum_{i=1}^P y^{t-1}_i - \nabla f(x^{t-1})\right\|^2$. 

Similarly to \cref{lemma:First-VarianceBound}, with probability at least $1-\nu$, $\tilde{I}^t_s=\emptyset$ holds for all $s\leq t-T_3$ and we can expand $\frac1P \sum_{i=1}^P y^{t}_i - \nabla f(x^{t})$ as
\begin{align}
\label{eq:Dist-First-Variance-1}
    & \frac1P \sum_{i=1}^P y^{t}_i - \nabla f(x^{t}) 
    \\ & = \underbrace{
    \frac{1}{P} \sum_{s=\max\{1,t-T_3+1\}}^t \left( \frac{|\tilde{I}_s^t|}{p}\sum_{i\in I^s} (\nabla f_i(x^{s})-\nabla f_i(x^{s-1}))
    - \sum_{i\in \tilde{I}_s^t}(\nabla f_i(x^{s})-\nabla f_i(x^{s-1}))
    \right)+\frac{\mathbbm{1}[t\leq T_3]}{P}\sum_{i\in \tilde{I}_1^t}(\tilde{y}^0_i-\nabla f_i(x^0))    }_{\rm (a)}
    \\ & \hspace{5mm}
    + \underbrace{\frac{1}{P} \sum_{s=\max\{1,t-T_3+1\}}^t \frac{|\tilde{I}^t_s|}{pKb}\sum_{i\in I^s} \sum_{j\in  
    J^s_i}(\nabla f_{i,j}(x^s)-\nabla f_{i,j}(x^{s-1})-(\nabla f_i(x^{s})-\nabla f_i(x^{s-1})))
    }_{\rm (b)}\\ & \hspace{5mm}
    + \underbrace{\frac{1}{PKb} \sum_{i=1}^P \mathbbm{1}[T_4(t,i)\geq t-T_3]\sum_{j\in J_i^{T_4(t,i)}}(\nabla f_{i,j}(x^{T_4(t,i)}) - \nabla f_{i}(x^{T_4(t,i)}))}_{\rm (c)}
    +
    \underbrace{
    \frac{\mathbbm{1}[t\leq T_3]}{P}\sum_{i\in \tilde{I}_1^t}(y^0_i-\tilde{y}^0_i)
    }_{\rm (d)}
\end{align}
for all $t$.

The norm of the part (a) can be bounded by using \cref{lemma:First-VarianceBound}, just replacing $n$ by $P$, i.e., 
\begin{align}
    \|{\rm (a)}\|^2 \leq \frac{15C_1^8\zeta^2}{p}\sum_{s=\max\{1,t-T_3+1\}}\|x^s-x^{s-1}\|^2 + 
    \frac{12C_1^2\mathbbm{1}[t\leq T_3]}{p}\left(\sigma_c^2 + \frac{G_c^2}{p}\right) 
\end{align}
for Option I and
\begin{align}
    \|{\rm (a)}\|^2 \leq \frac{15C_1^8\zeta^2}{p}\sum_{s=\max\{1,t-T_3+1\}}\|x^s-x^{s-1}\|^2 
\end{align}
for Option II, with probability at least $1-3\nu$ for all $t$.
For the bound of (b), (c) and (d), we apply \cref{lemma:Dist-First-Auxiliary-D}, \cref{lemma:Dist-First-Auxiliary-E}, and \cref{lemma:Dist-First-Auxiliary-C}, respectively.

Then, by summarizing all these and using $\|x^s-x^{s-1}\|^2\leq K\sum_{l=1}^K \|x^{s,l}-x^{s,l-1}\|^2$, we get
\begin{align}
    \left\|\frac1P \sum_{i=1}^P y^{t}_i - \nabla f(x^{t}) \right\|^2 
     & \le 4\|{\rm (a)}\|^2+4\|{\rm (b)}\|^2+4\|{\rm (c)}\|^2+4\|{\rm (d)}\|^2
    \\ & \leq 
    \left(\frac{60C_1^8\zeta^2}{p}+\frac{16C_1^{10}\zeta^2}{pKb}\right) \sum_{s=\max\{1,t-T_3+1\}}^t\|x^s-x^{s-1}\|^2
    +
    \frac{4C_1^2}{PKb}\left(\sigma^2 + \frac{G^2}{PKb}\right)
    \\ &\qquad\qquad +
    48C_1^2\mathbbm{1}[t\leq T_3]\left(\frac{\sigma^2}{pKb} + \frac{G^2}{p^2K^2b^2}+\frac{\sigma^2_c}{p} + \frac{G^2_c}{p^2}\right)
    \\ & \leq 
    \left(\frac{60C_1^8\zeta^2K}{p}+\frac{16C_1^{10}\zeta^2}{pb}\right) \sum_{s=\max\{1,t-T_3+1\}}^t\sum_{l=1}^K \|x^{s,l}-x^{s,l-1}\|^2
    +
    \frac{4C_1^2}{PKb}\left(\sigma^2 + \frac{G^2}{PKb}\right)
    \\ &\qquad\qquad +
    48C_1^2\mathbbm{1}[t\leq T_3]\left(\frac{\sigma^2}{pKb} + \frac{G^2}{p^2K^2b^2}+\frac{\sigma^2_c}{p} + \frac{G^2_c}{p^2}\right)
\end{align}
for Option I and
\begin{align}
    \left\|\frac1P \sum_{i=1}^P y^{t}_i - \nabla f(x^{t}) \right\|^2
    &\leq 
    \left(\frac{60C_1^8\zeta^2}{p}+\frac{16C_1^{10}\zeta^2}{pKb}\right) \sum_{s=\max\{1,t-T_3+1\}}^t\|x^s-x^{s-1}\|^2
    +
    \frac{4C_1^2}{PKb}\left(\sigma^2 + \frac{G^2}{PKb}\right)\\
    &    \leq 
    \left(\frac{60C_1^8\zeta^2K}{p}+\frac{16C_1^{10}\zeta^2}{pb}\right) \sum_{s=\max\{1,t-T_3+1\}}^t\sum_{l=1}^K \|x^{s,l}-x^{s,l-1}\|^2
    +
    \frac{4C_1^2}{PKb}\left(\sigma^2 + \frac{G^2}{PKb}\right)
\end{align}
for Option II, with probability $1-7\nu$ for all $t$.

Also, we have $\|z^{t,k}-(\nabla f(x^{t,k}-\nabla f(x^{t,0}))\|^2\leq  \left(2\zeta^2K+\frac{2C_1^2L^2}{b}\right)\sum_{l=1}^k\|x^{t,l}-x^{t,l-1}\|^2$ by \cref{lemma:Dist-First-Auxiliary-F}, with probability $1-\nu$ over all $t,k$.

By substituting these bound to \eqref{eq:DistAuxiliary100}, we obtain the desired bound.
\end{proof}

Now, we are ready to prove the first-order convergence of FLEDGE.

\begin{proof}[Proof of \cref{theorem:Dist-First-Main}']
Summing up \eqref{eq:Dist-First-FunctionDecrease-1} over all $t$ and $k$ and rearranging the terms, we get
\begin{align}
& \sum_{t=1}^T\sum_{k=1}^{K} \|\nabla f(x^{t,k-1})\|^2 \\ & \leq
\frac{2}{\eta}\left[\left(f(x^0)-f(x^{T})\right) -
\sum_{t=1}^T\sum_{k=1}^{K}\left(\frac{1}{2\eta}-\frac{L}{2}\right)\|x^{t,k}-x^{t,k-1}\|^2+\eta \sum_{t=1}^T\sum_{k=1}^{K} \|\nabla f(x^{t,k-1}) - v_{t,k-1}\|^2
+\frac{2TKr^2}{\eta^2}\right].
\end{align}

Applying \cref{lemma:Dist-First-VarianceBound} to this, we have that
\begin{align}
& \sum_{t=1}^T\sum_{k=1}^{K} \|\nabla f(x^{t,k-1})\|^2 \\ & \leq
\frac{2}{\eta}\left(f(x^0)-f(x^{T})\right) -
\frac{2}{\eta}\sum_{t=1}^T\sum_{k=1}^{K}\left(\frac{1}{2\eta}-\frac{L}{2}-\eta \left(\frac{120C_1^9\zeta^2PK^2}{p^2}+\frac{128C_1^{11}\zeta^2PK}{p^2b}+4\zeta^2K^2+\frac{4C_1^2L^2K}{b}\right)\right)\|x^{t,k}-x^{t,k-1}\|^2
    \\ & \hspace{10mm}+\begin{cases}
    \displaystyle
    \frac{16C_1^2T}{Pb}\left(\sigma^2 + \frac{G^2}{PKb}\right)+\frac{2TKr^2}{\eta^2}+
         192C_1^3\left(\frac{\sigma^2P}{p^2b} + \frac{PG^2}{p^3Kb^2}+\frac{PK\sigma^2_c}{p^2} + \frac{PKG^2_c}{p^3}\right)&\quad (\text{\rm Option I})
    \\\displaystyle
     \frac{16C_1^2T}{Pb}\left(\sigma^2 + \frac{G^2}{PKb}\right)+\frac{2r^2}{\eta^2}&\quad (\text{\rm Option II})
    \end{cases}
\end{align}
with probability at least $1-8\nu$.

\paragraph{Option I}
We set $\eta$ as
\begin{align}
    \eta = \min\left\{\frac{1}{2L},\left(\frac{480C_1^9\zeta^2P}{p^2}+\frac{604C_1^{11}\zeta^2PK}{p^2b}+16\zeta^2K^2+\frac{16C_1^2L^2K}{b}\right)^{-\frac12} \right\}=\tilde{\Theta}\left(\frac{1}{L}\land\frac{p}{\zeta \sqrt{P}K}\land \frac{p\sqrt{b}}{\zeta\sqrt{PK}}\land \frac{1}{\zeta K}\land \frac{\sqrt{b}}{L\sqrt{K}}\right),
\end{align}
so that 
$\frac{1}{2\eta}-\frac{L}{2}-\eta \left(\frac{152C_1^{11}\zeta^2PK}{p^2b}+4\zeta^2K^2+\frac{4C_1^2L^2K}{b}\right)\geq 0$ holds.
By taking $r\leq \frac{\eta\eps}{2\sqrt{2}}$ and 
$PKb \geq \frac{128C_1^2\sigma^2}{\eps^2}+\frac{8\sqrt{2}C_1G}{\eps}$
 , we obtain $\frac{2TKr^2}{\eta^2}\leq \frac{TK\eps^2}{4}$ and 
$ \frac{16C_1^2T}{Pb}\left(\sigma^2 + \frac{G^2}{PKb}\right)\leq \frac{TK\eps^2}{4}$, respectively.
Moreover, we apply $f(x^0)-f(x^t)\leq \Delta$.
Summarizing these, we get 
\begin{align}
    \sum_{t=1}^T\sum_{k=1}^{K} \|\nabla f(x^{t,k-1})\|^2 
\leq \frac{2\Delta}{\eta}+
\frac{TK\eps^2}{2}+
         192C_1^3\left(\frac{\sigma^2P}{p^2b} + \frac{PG^2}{p^3Kb^2}+\frac{PK\sigma^2_c}{p^2} + \frac{PKG^2_c}{p^3}\right)
.
\end{align}
Hence, taking
\begin{align}
    TK &\geq \frac{4\Delta}{\eta\eps^2}+\frac{384C_1^3}{\eps^2}\left(\frac{\sigma^2P}{p^2b} + \frac{PG^2}{p^3Kb^2}+\frac{PK\sigma^2_c}{p^2} + \frac{PKG^2_c}{p^3}\right)
    \\ &=\tilde{O}\left(
    \left(L\lor
    \frac{\zeta\sqrt{P}K}{p} \lor
    \frac{\zeta\sqrt{PK}}{p\sqrt{b}}\lor \zeta K\lor \frac{L\sqrt{K}}{\sqrt{b}}\right)\frac{\Delta}{\eps^2}\land
    \left(\frac{\sigma^2P}{p^2b}\lor \frac{PG^2}{p^3Kb^2}\lor\frac{PK\sigma^2_c}{p^2} \lor \frac{PKG^2_c}{p^3}\right)\frac{1}{\eps^2}
    \right)
\end{align}
results in 
\begin{align}
    \frac{1}{TK}\sum_{t=1}^T\sum_{k=1}^{K} \|\nabla f(x^{t,k-1})\|^2\le\epsilon^2,
\end{align}
which implies that {\AlgDist} can find $\eps$-first order stationary points with probability at least $1-8\nu$.
Thus, the gradient complexity and the communication complexity are bounded as stated.

\paragraph{Option II}
We set $\eta$ as the same as that for Option I, so that $\frac{1}{2\eta}-\frac{L}{2}-\eta \left(\frac{152C_1^{11}\zeta^2PK}{p^2b}+4\zeta^2K^2+\frac{4C_1^2L^2K}{b}\right)\geq 0$ holds as well.
We take $r\leq \frac{\eta\eps}{2\sqrt{2}}$ and 
$PKb \geq \frac{128C_1^2\sigma^2}{\eps^2}+\frac{8\sqrt{2}C_1G}{\eps}$.
Then, we get 
\begin{align}
    \sum_{t=1}^T\sum_{k=1}^{K} \|\nabla f(x^{t,k-1})\|^2 
\leq \frac{2\Delta}{\eta}+
\frac{TK\eps^2}{2}
.
\end{align}
Therefore, by the similar argument to Option I, taking $TK \geq \frac{2\Delta}{\eta\eps^2}=\tilde{O}\left(
    \left(L\lor
    \frac{\zeta\sqrt{P}K}{p}\lor \frac{\zeta\sqrt{PK}}{p\sqrt{b}}\lor \zeta K\lor \frac{L\sqrt{K}}{\sqrt{b}}\right)\frac{\Delta}{\eps^2}
    \right)$ ensures that {\AlgDist} finds $\epsilon$-first order stationary points with probability at least $1-8\nu$.
\end{proof}

\subsection{Finding Second-order Stationary Points (Proof of \cref{theorem:D-Second-Main})}

Here, we show that \AlgDist\ can efficiently find second-order stationary points.
With a slight abuse of notations, we sometimes identify $(t,k)$ with $(t',k')$ when $tK+k = t'K+k'$ holds.
Moreover, we say $(t_1,k_1)>(t_2,k_2)$ when $t_1K+k_1>t_2K+k_2$.

First, we state the formal version of \cref{theorem:D-Second-Main} as follows:
\newtheorem*{theorem:FLEDFEsecond}{\rm\bf Theorem~\ref{theorem:D-Second-Main}'}
\begin{theorem:FLEDFEsecond}\label{theorem:D-Second-Main-Appendix}
We assume  \cref{assumption:GradLipschitz,assumption:GlobalInfimum,assumption:BoundedGradient,assumption:Heterogeneity,assumption:HessianLipschitz}, and $\delta <\frac{1}{\zeta}$.
Let $p\geq \sqrt{P}+\frac{\Ce^2\zeta^2}{\delta^2}+\frac{\Ce^2L^2}{Kb\delta^2}$ with $\Ce=\tilde{O}(1)$, $b\geq K$, $K=\tilde{O}\left(\frac{L}{\zeta}\right)$, $\eta = \tilde{\Theta}\left(\frac{1}{L}\right)$, $r=\tilde{O}\left(\frac{\eps}{L}\right)$, $PKb\geq O\left(\frac{\sigma^2}{\eps^2}+\frac{G}{\eps}\right)$ and $\nu\in (0,1)$.
Then,  \AlgDist\ with Option I finds $(\eps,\delta)$-second-order stationary points using
\begin{align}
    \tilde{O}\left(\left(L\Delta +
         \left(\frac{\sigma^2}{pb} + \frac{G^2}{pKb^2}+K\sigma^2_c+ \frac{KG^2_c}{p}\right)\right)\left(\frac{1}{K\eps^2}+\frac{\rho^2}{K\delta^4}\right)pKb\right)
    \quad \text{stochastic gradients and}\\ \quad
    \tilde{O}\left(\left(L\Delta +
         \left(\frac{\sigma^2}{pb} + \frac{G^2}{pKb^2}+K\sigma^2_c+ \frac{KG^2_c}{p}\right)\right)\left(\frac{1}{K\eps^2}+\frac{\rho^2}{K\delta^4}\right)\right)
    \quad \text{communication rounds},
\end{align}
with probability at least $1-12\nu$.

Moreover, \AlgDist\ with Option II finds $(\eps,\delta)$-second-order stationary points using
\begin{align}
   \tilde{O}\left(PKb+L\Delta \left(\frac{1}{\eps^2}+\frac{\rho^2}{\delta^4}\right)pKb\right)
    \quad \text{stochastic gradients and}\\ \quad
   \tilde{O}\left(L\Delta \left(\frac{1}{\eps^2}+\frac{\rho^2}{\delta^4}\right)\right)
    \quad \text{communication rounds},
\end{align}
with probability at least $1-12\nu$.
\end{theorem:FLEDFEsecond}

Similarly to the previous section, the key argument is the exponential separation of two coupled trajectories with different initial values. 
\begin{lemma}[Small Stuck Region]\label{lemma:D-Small-Stuck-Region}
Assume $\delta <\frac{1}{\zeta}$. Let $\{x^{t,k}\}$ be a sequence generated by \AlgDist\ and $(\tau_0, \kappa_0)\ (0\leq \kappa_0<K)$ be a step where $-\gamma :=\lambda_{\rm min}(\nabla^2 f(x^{\tau_0,\kappa_0}))\leq -\delta$ holds.
We denote the eigenvector with the eigenvalue $\lambda_{\rm min}(\nabla^2 f(x^{\tau_0,\kappa_0}))$ by $\mathsf{e}$.
Moreover, let $\{\tilde{x}^{t,k}\}$ by a coupled sequence that is generated by \AlgDist\ with $\tilde{x}^{0} = x^0$ and shares the same choice of randomness with $\{x_t\}$ i.e., client samplings, minibatches and noises, except for the noise at a step $(\tau_0,K)> (\tau_0,\kappa_0)$: $\tilde{\xi}^{\tau_0,K}
= \xi^{\tau_0,K} - r_e \mathsf{e}$ with $r_e \geq \frac{r\nu }{TK\sqrt{d}}$.
Let $w^{t,k} = x^{t,k} - \tilde{x}^{t,k}$, $w^{t} = x^{t} - \tilde{x}^{t}$, $v^{t,k} = \frac{1}{P}\sum_{i=1}^P y_i^{t-1}+z^{t,k}$, $\tilde{v}^{t,k} = \frac{1}{P}\sum_{i=1}^P\tilde{y}_i^{t-1}+z^{t,k}$, $g^t = \frac{1}{P}\sum_{i=1}^P y_i^{t}-\nabla f(x^t)- \left(\frac{1}{P}\sum_{i=1}^P \tilde{y}_i^{t}-\nabla f(\tilde{x}^t)\right)$, and $h^{t,k} = (z^{t,k} - (\nabla f(x^{t,k})-\nabla f(x^{t,0})))-(\tilde{z}^{t,k} - (\nabla f(\tilde{x}^{t,k})-\nabla f(\tilde{x}^{t,0})))$.
Using these notations, $v^{t,k} - \nabla f(x^{t,k}) - (\tilde{v}^{t,k} - \nabla f(\tilde{x}^{t,k}))=g^{t-1}+h^{t,k}$ holds.

Then, there exists a sufficiently large constants $\Ce=\tilde{O}(1)$ and $\Cf=O(1)$ with which the following holds:
If we take $p\geq \sqrt{P}+\frac{\Ce^2\zeta^2}{\delta^2}+\frac{\Ce^2L^2}{Kb\delta^2}, b\geq K, K = O(\frac{L}{\zeta})$, $\eta = \tilde{\Theta}\left(\frac{1}{L}\right)$, and $T_5=\frac{\Cf\log \frac{\delta }{ \Ce \rho r_e}}{\eta \gamma}\lesssim \tilde{O}\left(\frac{L}{\delta }\right)$, with probability $1-\frac{3\nu}{TK}~(\nu \in (0,1))$, we have
\begin{align}
    \max_{(\tau_0,\kappa_0)\leq  (t,k)< (\tau_0 + 1,T_5)}\{\|x^{\tau,k}-x^{\tau_0,\kappa_0}\|,\|\tilde{x}^{\tau,\kappa}-x^{\tau_0,\kappa_0}\|\} \geq  \frac{\delta}{\Ce\rho}
    .
\end{align}

\end{lemma}
In order to show \cref{lemma:D-Small-Stuck-Region}, we prepare the two following lemmas, which bound the difference between gradient estimation errors of the two sequence.
\begin{lemma}\label{lemma:D-Second-Auxiliary-A}
Under the same assumption as that of \cref{lemma:D-Small-Stuck-Region}, we assume $    \max_{(\tau_0,\kappa_0)\leq  (t,k)< (\tau_0 + 1,T_5)}\{\|x^{\tau,k}-x^{\tau_0,\kappa_0}\|,\|\tilde{x}^{\tau,\kappa}-x^{\tau_0,\kappa_0}\|\} < \frac{\delta}{\Ce\rho}$. Then, the following holds uniformly for all $(\tau_0,\kappa_0)\leq  (t,k)\leq (\tau_0 + 1,T_5)$ with probability at least $1-\frac{\nu}{TK}$:
\begin{align}
    \left\|g^t\right\| 
    \leq \begin{cases}
    0 & (t< \tau_0),\\
    \displaystyle
    \left(\frac{\zeta}{\sqrt{p}}+\frac{L}{\sqrt{pKb}}\right)\Cg r_e& (t=\tau_0),
    \\ 
    \displaystyle
    \left(\frac{\zeta}{\sqrt{p}}+\frac{L}{\sqrt{pKb}}\right)\Cg r_e
    +
    \left(\frac{\zeta \sqrt{K}}{\sqrt{p}}+\frac{L}{\sqrt{pb}}\right)\Cg\sqrt{\sum_{s=\max\{\tau_0+1,t-T_3+1\}}^{t}\sum_{k=1}^K\|w^{s,k} - w^{s,k-1}\|^2}
    \\
    \displaystyle
    \qquad\qquad + \frac{\Cg\delta}{\Ce\sqrt{p}}\sqrt{\sum_{s=\max\{\tau_0,t-T_3+1\}}^{t}\|w^{s}\|^2}
    & (t\geq \tau_0+1),
    \end{cases}
\end{align}
where $T_3 = \frac{P}{p}C_1$, and $\Cg=\tilde{O}(1)$ is a sufficiently large constant.
\end{lemma}

\begin{lemma}\label{lemma:D-Second-Auxiliary-B}
Under the same assumption as that of \cref{lemma:D-Small-Stuck-Region}, the following holds uniformly for all $t \geq \tau_0+1$ and $k\geq 0$ with probability at least $1-\frac{2\nu}{TK}$:
\begin{align}
    \left\|h^{t,k}\right\|
    \leq \zeta\sum_{l=1}^k\|w^{t,l}-w^{t,l-1}\|+\frac{2\delta}{\Ce}\|w^{t,k}\|+\frac{2\delta}{\Ce}\|w^{t,0}\|+\frac{C_1^2}{\sqrt{b}}\sqrt{
    \sum_{l=1}^k\left(L\|w^{t,l}-w^{t,l-1}\|+\frac{2\delta}{\Ce}\|w^{t,l}\|+\frac{2\delta}{\Ce}\|w^{t,l-1}\|\right)^2
    }.
\end{align}
For $t< \tau_0+1$, we have $\left\|h^{t,k}\right\|=0$.
\end{lemma}



\begin{proof}[Proof of \cref{lemma:D-Second-Auxiliary-A}]
As for the case $t<\tau_0$, the assertion directly follows from the definition of $\{\tilde{x}^{t,k}\}$.
For the proof of the rest cases, we use notations as follows:
\begin{align}
    H&=\nabla^2f(x^{\tau_0,\kappa_0}),\\
    H_i&=\nabla^2 f_i(x^{\tau_0,\kappa_0})\\
    H_{i,j}&=\nabla^2 f_{i,j}(x^{\tau_0,\kappa_0}),\\
    \mathrm{d}H^{t,k}&=\int_0^1 (\nabla^2 f(\tilde{x}^{t,k} + \theta(x^{t,k}-\tilde{x}^{t,k})) - H)\mathrm{d}\theta,\\
    \mathrm{d}H^{t,k}_i&=\int_0^1 (\nabla^2 f_i(\tilde{x}^{t,k}+ \theta(x^{t,k}-\tilde{x}^{t,k})) - H_i)\mathrm{d}\theta,\\
    \mathrm{d}H^{t,k}_{i,j}&=\int_0^1 (\nabla^2 f_{i,j}(\tilde{x}^{t,k} + \theta(x^{t,k}-\tilde{x}^{t,k})) - H_{i,j})\mathrm{d}\theta.
\end{align}
Moreover,  we denote 
\begin{align}
    u_i^s \coloneqq (\nabla f_i(x^{s})-\nabla f_i(\tilde{x}^{s}))
     - (\nabla f_i(x^{s-1})-\nabla f_i(\tilde{x}^{s-1}))
     - (\nabla f(x^{s})-\nabla f(\tilde{x}^{s}))
     +(\nabla f(x^{s-1})-\nabla f(\tilde{x}^{s-1}))    
\end{align}
and 
\begin{align}
    u_{i,j}^s := (\nabla f_{i,j}(x^{s})-\nabla f_{i,j}(\tilde{x}^{s}))
     - (\nabla f_{i,j}(x^{s-1})-\nabla f_{i,j}(\tilde{x}^{s-1}))
     - (\nabla f_i(x^{s})-\nabla f_i(\tilde{x}^{s}))
     +(\nabla f_i(x^{s-1})-\nabla f_i(\tilde{x}^{s-1})).
\end{align}
Note that $\mathbb{E}_i[u_i^s]=0$ (expectation with respect to the choice of $i$) and $\mathbb{E}_j[u_{i,j}^s]=0$ (expectation with respect to the choice of $j$) hold.
Using \cref{assumption:GradLipschitz,assumption:HessianLipschitz,assumption:Heterogeneity} and $\max_{(\tau_0,\kappa_0)\leq  (t,k)< (\tau_0 + 1,T_5)}\{\|x^{\tau,k}-x^{\tau_0,\kappa_0}\|,\|\tilde{x}^{\tau,\kappa}-x^{\tau_0,\kappa_0}\|\} < \frac{\delta}{\Ce\rho}$, we can derive that
\begin{align}
    \|u_i^s\| \leq
     \zeta \|w^s - w^{s-1}\| + \frac{2\delta}{\Ce}\|w^s\| + \frac{2\delta}{\Cb}\|w^{s-1}\|
     \quad\text{ and }\quad
    \|u_{i,j}^s\| \leq
     L \|w^s - w^{s-1}\| + \frac{2\delta}{\Ce}\|w^s\| + \frac{2\delta}{\Cb}\|w^{s-1}\|   
     \label{eq:D-Second-Auxiliary-A-1}
\end{align}
for $s\geq \tau_0+1, $ by similar argument to the proof of \cref{lemma:Second-Auxiliary-A}.
For $t = \tau_0 $, we have $\|u_i^{\tau_0}\| = \|(\nabla f_i(x^{\tau_0})-\nabla f_i(\tilde{x}^{\tau_0}))
     - (\nabla f(x^{\tau_0})-\nabla f(\tilde{x}^{\tau_0}))\|\leq \zeta\|x^{\tau_0}-\tilde{x}^{\tau_0}\|=\zeta r_e$ and $\|u_{i,j}^{\tau_0}\| = \|(\nabla f_{i,j}(x^{\tau_0})-\nabla f_{i,j}(\tilde{x}^{\tau_0}))
     - (\nabla f(x^{\tau_0})-\nabla f(\tilde{x}^{\tau_0}))\|\leq L\|x^{\tau_0}-\tilde{x}^{\tau_0}\|=L r_e$.

As we did in \cref{lemma:Second-Auxiliary-A}, for $t\geq \tau_0+1$, we have 
\begin{align}
    g^{t}&=
     \underbrace{\frac{1}{P} \left( \frac{|\tilde{I}_{\tau_0}^{t}|}{p}\sum_{i\in I^{\tau_0}} u_i^{\tau_0} - \sum_{i\in \tilde{I}_{\tau_0}^{t}} u_i^{\tau_0} \right)}_{(a)}
     +
     \underbrace{\frac{1}{PKb} \left( \frac{|\tilde{I}_{\tau_0}^{\tau_0}|}{p}\sum_{i\in I^{\tau_0}}\sum_{j\in J^{\tau_0}_i} u_{i,j}^{\tau_0} - \sum_{i\in \tilde{I}_{\tau_0}^{\tau_0}}\sum_{j\in J^{\tau_0}_i} u_{i,j}^{\tau_0} \right)}_{(b)}
     \\
   &\hspace{5mm}+
    \underbrace{\frac{1}{P} \sum_{s=\max\{\tau_0+1,t-T_1+1\}}^{t} \left( 
     \frac{|\tilde{I}_s^{t}|}{p}\sum_{i\in I^s} u_i^s  \right)- \frac{1}{P} \sum_{s=\max\{\tau_0+1,t-T_1+1\}}^{t}  \sum_{i\in \tilde{I}_s^t}u_i^s }_{(c)}
      \\
     &\hspace{5mm}+\underbrace{\frac{1}{PKb} \sum_{s=\max\{\tau_0+1,t-T_1+1\}}^{t} \left( 
     \frac{|\tilde{I}_s^{t}|}{p}\sum_{i\in I^s} \sum_{j\in J^s_i}u_{i,j}^s  \right)- \frac{1}{PKb} \sum_{s=\max\{\tau_0+1,t-T_1+1\}}^{t}  \sum_{i\in \tilde{I}_s^t}\sum_{j\in J^s_i}u_{i,j}^s }_{(d)}
     \label{eq:Second-Auxiliary-A-2}
\end{align}
with probability $1-\frac{\nu}{8TK}$ for all $t$. For $t=\tau_0$, $g^{\tau_0} = {\rm (a)}+{\rm (b)}$ holds.

Recall the argument in \cref{lemma:Second-Auxiliary-A}. We have that
\begin{align}
    \|{\rm (a)}\|\leq \frac{2C_1\zeta r_e}{\sqrt{p}} \quad \text{and} \quad
    \|{\rm (c)}\|\leq 
    \frac{2C_1^4 + C_1^\frac32}{\sqrt{p}}\sqrt{
    \sum_{s=\max\{\tau_0+1,t-T_3+1\}}\left(\zeta\|w^s-w^{s-1}\|+\frac{2\delta}{\Ce}\|w^s\|+\frac{2\delta}{\Ce}\|w^{s-1}\|\right)^2
    }
\end{align}
hold with probability at least $1-\frac{1}{8TK}$ for all $t$.

Moreover, observe that (b) and (d) are obtained just by replacing $u^t_i$ in (a) and (c) by $\frac{1}{Kb}\sum_{j\in J^t_i}u_{i,j}^s$.
Note that $\frac{1}{Kb}\sum_{j\in J^{\tau_0}_i}u_{i,j}^{\tau_0}$ is mean-zero and its norm is bounded by 
    $ \frac{C_1Lr_e}{\sqrt{Kb}}$ for $s={\tau_0}$, with probability $1-\frac{1}{8T^2K}$.
Thus, \cref{proposition:Azuma} yields that
\begin{align}
    &
    \left\|{\rm (b)}\right\|
    \leq
    \left\|\frac{1}{P} \frac{|\tilde{I}_{\tau_0}^{t}|}{p}\sum_{i\in I^{\tau_0}}\sum_{j\in J^{\tau_0}_i}u_{i,j}^{\tau_0} \right\| + \left\|\frac{1}{P} \sum_{i\in \tilde{I}_{\tau_0}^{t}} \sum_{j\in J^{\tau_0}_i}u_{i,j}^{\tau_0} \right\|
     \leq
     \frac{|\tilde{I}_{\tau_0}^{t}|}{P}\frac{C_1^2 L r_e}{\sqrt{pKb}}+\frac{\sqrt{|\tilde{I}_{\tau_0}^{t}|}C_1^2 L r_e}{P\sqrt{Kb}}
     \leq \frac{2C_1^2 L r_e}{\sqrt{pKb}},
     \label{eq:D-Second-Auxiliary-A-3}
\end{align}
with probability $1-\frac{1}{8TK}-T\cdot\frac{1}{8T^2K}=1-\frac{1}{4TK}$, where we use $|\tilde{I}_{\tau_0}^{t}|\le P$ and $p\le P$ for the last inequality.
    
For the first term of (d), we first observe that $\frac{1}{Kb}\sum_{i\in I^s}\sum_{j\in J^t_i}u_{i,j}^s$ is mean-zero and its norm is bounded by $
    \frac{C_1\sqrt{p}}{\sqrt{Kb}} L \|w^s - w^{s-1}\| + \frac{2C_1\delta\sqrt{p}}{\Ce\sqrt{Kb}}\|w^s\| + \frac{2C_1\delta\sqrt{p}}{\Ce\sqrt{Kb}}\|w^{s-1}\|  $ (for $s\geq \tau_0 + 1$) with probability at least$1-\frac{1}{8T^2K}$.
    Then, we apply the same argument as \cref{lemma:First-Auxiliary-B}. 
    This yields
\begin{align}
 \left\|\frac{1}{PKb} \sum_{s=\max\{\tau_0+1,t-T_1+1\}}^{t}
     \frac{|\tilde{I}_s^{t}|}{p}\sum_{i\in I^s} \sum_{j\in J^s_i}u_{i,j}^s\right\|
    \leq
    \frac{2C_1^4}{\sqrt{pKb}}\sqrt{\sum_{s=\max\{\tau_0+1,t-T_1+1\}}^t\left(L \|w^s - w^{s-1}\| + \frac{2\delta}{\Ce}\|w^s\| + \frac{2\delta}{\Ce}\|w^{s-1}\|\right)^2}
.\label{eq:D-Second-Auxiliary-A-4}
\end{align}
with probability $1-T\cdot\frac{\nu}{8T^2K}-\frac{\nu}{8TK}=1-\frac{\nu}{4TK}$ for all $t$.
As for the second term of (d), by applying \cref{proposition:BernsteinNoReplacement}, we get
\begin{align}
    \left\|\frac{1}{PKb} \sum_{s=\max\{\tau_0+1,t-T_1+1\}}^{t}  \sum_{i\in \tilde{I}_s^t} \sum_{j\in J^s_i}u_{i,j}^s\right\|
   & \leq
    \frac{\sqrt{T_3}}{P} \sqrt{\sum_{s=\max\{\tau_0+1,t-T_1+1\}}^{t}
    \left\|\frac{1}{Kb}\sum_{i\in \tilde{I}_s^t} \sum_{j\in J^s_i}u_{i,j}^s\right\|^2}
    \\ & \leq
    \frac{C_1^\frac12}{\sqrt{Pp}} \sqrt{\sum_{s=\max\{\tau_0+1,t-T_1+1\}}^{t}
    \frac{C_1^2p}{Kb}\left(L \|w^s - w^{s-1}\| + \frac{2\delta}{\Ce}\|w^s\| + \frac{2\delta}{\Ce}\|w^{s-1}\|\right)^2}
    \\ & \leq
    \frac{C_1^\frac32}{\sqrt{pKb}} \sqrt{\sum_{s=\max\{\tau_0+1,t-T_1+1\}}^{t}
    \left(L \|w^s - w^{s-1}\| + \frac{2\delta}{\Ce}\|w^s\| + \frac{2\delta}{\Ce}\|w^{s-1}\|\right)^2}
     \label{eq:D-Second-Auxiliary-A-8}  
\end{align}
with probability $1-T\cdot\frac{\nu}{8T^2K}-\frac{\nu}{8TK}=1-\frac{\nu}{4TK}$ for all $t$.

By combining all these, 
we have
\begin{align}
    \|g^{t}\| 
    &
    \leq 
    \left(\frac{\zeta}{\sqrt{p}}+\frac{L}{\sqrt{pKb}}\right)\Cg\zeta r_e
    +
    \left(\frac{\zeta}{\sqrt{p}}+\frac{L}{\sqrt{pKb}}\right)\Cg\sqrt{\sum_{s=\max\{\tau_0+1,t-T_3+1\}}^{t}\|w^s - w^{s-1}\|^2}
    \\ & \hspace{5mm}+
    \frac{\Cg\delta}{\Ce}\sqrt{\sum_{s=\max\{\tau_0,t-T_3+1\}}^{t}\|w^{s}\|^2}
    \\ & \leq
    \left(\frac{\zeta}{\sqrt{p}}+\frac{L}{\sqrt{pKb}}\right)\Cg\zeta r_e
    +
    \left(\frac{\zeta \sqrt{K}}{\sqrt{p}}+\frac{L}{\sqrt{pb}}\right)\Cg\sqrt{\sum_{s=\max\{\tau_0+1,t-T_3+1\}}^{t}\sum_{k=1}^K\|w^{s,k} - w^{s,k-1}\|^2}
    \\ & \hspace{5mm}+
    \frac{\Cg\delta}{\Ce\sqrt{p}}\sqrt{\sum_{s=\max\{\tau_0,t-T_3+1\}}^{t}\|w^{s}\|^2}
\end{align}
with probability at least $1-\frac{\nu}{TK}$ for all $t\geq\tau_0+1$.
Here we take $\Cg = \tilde{O}(1)$, which is independent of $\Ce$. 
Thus, we get the assertion for $t\ge \tau_0+1$.
For $t=\tau_0$, the bounds on (a) and (b) imply the desired bound.
\end{proof}

\begin{proof}[Proof of \cref{lemma:D-Second-Auxiliary-B}]
Let 
\begin{align}
u_i^{t,l} \coloneqq (\nabla f_i(x^{t,l})-\nabla f_i(\tilde{x}^{t,l}))
     - (\nabla f_i(x^{t,0})-\nabla f_i(\tilde{x}^{t,0}))
     - (\nabla f(x^{t,l})-\nabla f(\tilde{x}^{t,l}))
     +(\nabla f(x^{t,0})-\nabla f(\tilde{x}^{t,0}))
\end{align}
     and 
\begin{align}
    u_{i,j}^{t,l} \coloneqq &(\nabla f_{i,j}(x^{t,l})-\nabla f_{i,j}(\tilde{x}^{t,l}))
     - (\nabla f_{i,j}(x^{t,l-1})-\nabla f_{i,j}(\tilde{x}^{t,l-1}))\\
     &- (\nabla f_i(x^{t,l})-\nabla f_i(\tilde{x}^{t,l}))
     +(\nabla f_u(x^{t,l-1})-\nabla f_i(\tilde{x}^{t,l-1}))
\end{align}
By their definitions, $h^{t,k} =u_{i_t}^{t,l} + \frac{1}{b}\sum_{l=1}^k\sum_{j\in J_{i_t}}u_{i_t,j}^{t,l}$ holds.
We can bound the norm of them as 
\begin{align}
    \|u_i^{t,k}\|\leq \zeta \|w^{t,k}-w^{t,0}\|+\frac{2\delta}{\Ce}\|w^{t,k}\|+\frac{2\delta}{\Ce}\|w^{t,0}\|
 \leq
    \zeta\sum_{l=1}^k\|w^{t,l}-w^{t,l-1}\|+\frac{2\delta}{\Ce}\|w^{t,k}\|+\frac{2\delta}{\Ce}\|w^{t,0}\|
    \label{eq:D-Second-Auxiliary-B-1}
\end{align}
and 
\begin{align}
   \|u_{i_t}^{t,l}\|\leq L \|w^{t,l}-w^{t,l}\|+\frac{2\delta}{\Ce}\|w^{t,l}\|+\frac{2\delta}{\Ce}\|w^{t,l-1}\| .
\end{align}

Thus, applying \cref{proposition:Bernstein} and \cref{proposition:Azuma} to $\frac{1}{b}\sum_{l=1}^k\sum_{j\in J_{i_t}}u_{i_t,j}^{t,l}$, we get
\begin{align}
    \left\|\frac{1}{b}\sum_{l=1}^k\sum_{j\in J_{i_t}}u_{i_t,j}^{t,l}\right\|
    \leq \frac{C_1^2}{\sqrt{b}}\sqrt{
    \sum_{l=1}^k\left(L\|w^{t,l}-w^{t,l-1}\|+\frac{2\delta}{\Ce}\|w^{t,l}\|+\frac{2\delta}{\Ce}\|w^{t,l-1}\|\right)^2
    }
    \label{eq:D-Second-Auxiliary-B-2}
\end{align}
with probability at least $1-\frac{1}{TK}$ for all $t$ and $K$.

Substituting \eqref{eq:D-Second-Auxiliary-B-1} and \eqref{eq:D-Second-Auxiliary-B-2} to $h^{t,k} =u_{i_t}^{t,l} + \frac{1}{b}\sum_{l=1}^k\sum_{j\in J_{i_t}}u_{i_t,j}^{t,l}$, we get the desired bound.
\end{proof}

Now, we are ready to prove \cref{lemma:D-Small-Stuck-Region}.
\begin{proof}[Proof of \cref{lemma:D-Small-Stuck-Region}]
We assume the contrary and show the following by induction, for $(\tau_0+1,0) \leq (t,k)\leq (\tau_0 +1,T_5)$:
\begin{align}
    &{\rm(a)}\quad \frac12(1+\eta\gamma)^{(t-\tau_0-1)K + k}r_e\leq \|w^{t,k}\|\leq 2(1+\eta\gamma)^{(t-\tau_0-1)K + k}r_e\\
    &{\rm(b)}\quad \|w^{t,k} - w^{t,k-1}\|\leq 
    \begin{cases}
    r_e  &(\text{for $(t,k)= (\tau_0 + 1,0)$})\\
    3\eta\gamma (1+\eta\gamma)^{(t-\tau_0-1)K + k}r_e & (\text{for $(t,k)> (\tau_0 + 1,0)$})
    \end{cases}
   \\
    &{\rm(c)}\quad \|g^{t-1}+h^{t,k}\|\leq 
    \frac{2\Ch\gamma }{\Ce} (1+\eta\gamma)^{(t-\tau_0-1)K + k}r_e
    .
\end{align}
Here $\Ch=\tilde{O}(1)$ is a sufficiently large constant independent of $\Ce$.
Then, (a) yields contradiction by taking $(t,k)-(\tau_0+1,0) = T_5 = O\left(1+\frac{\log\frac{\delta}{\Cb\rho r_e}}{\eta\delta K}\right)$ to break the assumption.

It is easy to check (a) and (b) for and $t=\tau_0 +1$ and $k=0$.
As for (c), checking the initial condition at $(t,k)=(\tau_0 +1,0)$ requires assumption on the size of $p$.
According to \cref{lemma:D-Second-Auxiliary-A}, taking $p \geq \frac{\zeta^2}{\delta^2}+\frac{L^2}{\delta^2Kb}$, $\|g^{\tau_0}\|\leq 2\Cg \delta r_e \leq 2\Cg \gamma r_e$ holds. 

Now, we derive that (a), (b) and (c) are true for $(t,k+1)$, assuming that they are true for all $(\tau_0+1,0),\cdots,(t,k)$.
To this end, we consider the decomposition of $w^{t,k}$ as follows:
\begin{align}
    w^{t,k+1} &= w^{t,k} - \eta\left(v^{t,k} - \tilde{v}^{t,k} \right) 
    \\ & =
    (1+\eta \gamma)^{(t-\tau_0-1)K+k+1}r_e\mathsf{e}
    -\eta \sum_{(s,l)=(\tau_0+1,0)}^{(t,k)}(I-\eta H)^{(t-s)K+(k-l)}(\mathrm{d}H^{s,l}w^{s,l} + g^{s-1}+ h^{s,l}),
    \label{eq:D-Second-SmallStuckRegion-1}
\end{align}
for $(t,k+1)\geq (\tau_0 + 1,1)$.

\paragraph{Verifying (a)}
The first term $(1+\eta \gamma)^{(t-\tau_0-1)K+k+1}r_e\mathsf{e}$ of \eqref{eq:D-Second-SmallStuckRegion-1} satisfies
\begin{align}
    \|(1+\eta \gamma)^{(t-\tau_0-1)K+k+1}r_e\mathsf{e}\|
    =(1+\eta \gamma)^{(t-\tau_0-1)K+k+1}r_e
    .
\end{align}
Then, focus on bounding $\eta \sum_{(s,l)=(\tau_0+1,0)}^{t,k}(I-\eta H)^{(t-s)K+(k-l)}(\mathrm{d}H^{s,l}w^{s,l} + g^{s-1}+ h^{s,l})$ by $\frac12 (1+\eta \gamma)^{(t-\tau_0-1)K+k+1}r_e$.
We have
\begin{align}
    \left\|\eta \sum_{(s,l)=(\tau_0+1,0)}^{(t,k)}(I-\eta H)^{(t-s)K+(k-l)}\mathrm{d}H^{s,l}w^{s,l} \right\|
    & \leq
    \eta \sum_{(s,l)=(\tau_0+1,0)}^{(t,k)}\|I-\eta H\|^{(t-s)K+(k-l)} \left\|\mathrm{d}H^{s,l}\right\|\left\|w^{s,l}\right\|
    \\ & \leq \label{eq:D-Second-SmallStuckRegion-2} 
    2\eta (1+\eta \gamma)^{(t-s)K+(k-l)+(s-\tau_0-1)K+l}r_e \sum_{(s,l)=(\tau_0+1,0)}^{t,k}\left\|\mathrm{d}H^{s,l}\right\|
    \\ & \leq \label{eq:D-Second-SmallStuckRegion-3}
    2\eta (1+\eta \gamma)^{(t-\tau_0-1)K+k}r_e T_5K\frac{\delta}{\Ce}
    \\ & \leq \label{eq:D-Second-SmallStuckRegion-4} 
     \frac{2\eta\delta T_5}{\Ce} (1+\eta \gamma)^{(t-\tau_0-1)K+k}r_e
    \\ & \leq \label{eq:D-Second-SmallStuckRegion-5} 
     \frac14 (1+\eta \gamma)^{(t-\tau_0-1)K+k}r_e    
     .
\end{align}
The last inequality follows from the definition of $T_5=\frac{\Cf\log \frac{\delta }{ \Ce \rho r_e}}{\eta \gamma }$ and sufficiently large $\Ce$.

In addition, we have
\begin{align}
    \left\|\eta \sum_{(s,l)=(\tau_0+1,0)}^{(t,k)}(I-\eta H)^{(t-s)K+(k-l)}(g^{s-1}+h^{s,l}) \right\|
    & \leq
    \eta \sum_{(s,l)=(\tau_0+1,0)}^{(t,k)}\|I-\eta H\|^{(t-s)K+(k-l)} \left\|g^{s-1}+h^{s,l}\right\|
    \\ & \leq \label{eq:D-Second-SmallStuckRegion-6} 
    \eta \sum_{(s,l)=(\tau_0+1,0)}^{(t,k)}(1+\eta\gamma)^{(t-s)K+(k-l)}
    \frac{2\Ch\gamma}{\Ce}(1+\eta\gamma)^{(s-\tau_0-1)K+l}
    \\ & \leq \label{eq:D-Second-SmallStuckRegion-7}
    \frac{2\eta\gamma T_5}{\Ce}(1+\eta \gamma)^{(t-\tau_0-1)K+k}
    \\ & \leq \label{eq:D-Second-SmallStuckRegion-8} 
     \frac14 (1+\eta \gamma)^{(t-\tau_0-1)K+k}r_e    
     .
\end{align}
For the final inequality, we again use $T_5=\frac{\Cf\log \frac{\delta }{ \Ce \rho r_e}}{\eta \gamma}$ with sufficiently large $\Ce$.

Combining \eqref{eq:D-Second-SmallStuckRegion-5} and \eqref{eq:D-Second-SmallStuckRegion-8}, we get (a) for $(t,k+1)$ as desired. 
\paragraph{Verifying (b)}
For $(t,k)\geq (\tau_0 + 1,0)$, we have
\begin{align}
&w^{t,k+1} - w^{t,k} \\ 
& =
    (1+\eta \gamma)^{(t-\tau_0-1)K+k+1}r_e\mathsf{e}
    -\eta \sum_{(s,l)=(\tau_0+1,0)}^{(t,k)}(I-\eta H)^{(t-s)K+(k-l)}(\mathrm{d}H^{s,l}w^{s,l} + g^{s-1}+ h^{s,l})
   \\ &\hspace{5mm} -
    (1+\eta \gamma)^{(t-\tau_0-1)K+k}r_e\mathsf{e}
    -\eta \sum_{(s,l)=(\tau_0+1,0)}^{(t,k-1)}(I-\eta H)^{(t-s)K+(k-l)}(\mathrm{d}H^{s,l}w^{s,l} + g^{s-1}+ h^{s,l})  
\\ &=   \eta\gamma (1+\eta \gamma)^{(t-\tau_0-1)K+k} r_e \mathsf{e}
   \\ & \hspace{5mm} - \eta \sum_{(s,l)=(\tau_0+1,0)}^{(t,k-1)}\eta H(I-\eta H)^{(t-s)K+(k-l)}(\mathrm{d}H^{s,l}w^{s,l} + g^{s-1}+ h^{s,l})  
     -\eta (\mathrm{d}H_{t}w_{t} + g^{t-1}+h^{t,k})
    .
\end{align}
As for the first term, we can bound it as
\begin{align}
    \|\eta\gamma (1+\eta \gamma)^{(t-\tau_0-1)K+k} r_e \mathsf{e}\|\leq \eta \gamma (1+\eta \gamma)^{(t-\tau_0-1)K+k}r_e.
\end{align}
Evaluating the second term requires (a) and (b) for $(\tau_0+1,0),\cdots,(t,k-1)$ and \cref{lemma:Murata-MatrixEigenvalue}:
\begin{align}
    &\left\|\eta \sum_{(s,l)=(\tau_0+1,0)}^{(t,k-1)}\eta H(I-\eta H)^{(t-s)K+(k-l)}(\mathrm{d}H^{s,l}w^{s,l} + g^{s-1}+ h^{s,l})  \right\|
    \\ & \leq 
   \sum_{(s,l)=(\tau_0+1,0)}^{(t,k-1)}\eta \left\|\eta H(I-\eta H)^{(t-s)K+(k-l)}\right\|\left(\|\mathrm{d}H^{s,l}\|\|w^{s,l}\| + \|g^{s-1}+ h^{s,l}\|\right)
    \\ & \leq 
   \sum_{(s,l)=(\tau_0+1,0)}^{(t,k-1)}\eta \left\|\eta H(I-\eta H)^{(t-s)K+(k-l)}\right\|\left(\frac{\delta}{\Ce}(1+\eta\gamma)^{(s-\tau_0-1)K+l}r_e + \frac{2\Ch\gamma}{\Ce} (1+\eta\gamma)^{(s-\tau_0-1)K+l}r_e\right)
    \\ & \leq 
    \sum_{(s,l)=(\tau_0+1,0)}^{(t,k-1)}\eta \left(\eta \gamma (1+\eta \gamma)^{(t-s)K+(k-l)} + \frac{1}{(t-s)K+(k-l)}\right)
        \left(\frac{\delta}{\Ce}+\frac{2\Ch\gamma}{\Ce}\right) (1+\eta\gamma)^{(s-\tau_0-1)K+l}r_e
    \\ & \leq 
    \eta\left(\eta \gamma T_5 + \log T_5 \right)\left(\frac{\delta}{\Ce}+\frac{2\Ch\gamma}{\Ce}\right) (1+\eta\gamma)^{(t-\tau_0-1)K+k}r_e.
\end{align}
Since $T_5 = \tilde{O}\left(\frac{1}{\eta\delta }\right)$ and $\gamma \geq \delta$, setting $\Ce=\tilde{O}(1)$ with sufficiently large $\Ce$ yields $\left(\eta \gamma T_5 + \log T_5 \right)\left(\frac{\delta}{\Ce}+\frac{2\Ch\gamma}{\Ce}\right)\leq \gamma$. Thus, the second term is bounded by $\eta\gamma(1+\eta\gamma)^{(t-\tau_0-1)K+k}r_e$.

Finally, we consider the third term.
We have $\|\mathrm{d}H^{t,k}w^{t,k}\|\leq \frac{\delta}{\Ce}r_e (1+\eta \gamma)^{(t-\tau_0-1)K+k}r_e$ and $\|g^{t-1}+h^{t,k}\|\leq \frac{2\Ch\gamma}{\Ce}(1+\eta \gamma)^{(t-\tau_0-1)K+k}r_e$.
Thus, by taking $\Ce$ sufficiently large, the third term is bounded by $\eta\gamma (1+\eta \gamma)^{(t-\tau_0-1)K+k}r_e$.

By combining these bounds, we get (b) for $(t,k+1)$.

\paragraph{Verifying (c)}
Using \cref{lemma:D-Second-Auxiliary-A} and assumptions, we have
\begin{align}
       & \|g^{t+1}\| 
      \\ & \leq \left(\frac{\zeta}{\sqrt{p}}+\frac{L}{\sqrt{pKb}}\right)\Cg r_e
    +
    \left(\frac{\zeta \sqrt{K}}{\sqrt{p}}+\frac{L}{\sqrt{pb}}\right)\Cg\sqrt{\sum_{s=\max\{\tau_0+1,t-T_3+1\}}^{t}\sum_{k=1}^K\|w^{s,k} - w^{s,k-1}\|^2}\\
    &\hspace{30mm}+
    \frac{\Cg\delta}{\Ce\sqrt{p}}\sqrt{\sum_{s=\max\{\tau_0,t-T_3+1\}}^{t}\|w^{s}\|^2}
    \\ & \leq
    \left[
    \frac{\zeta \Cg}{\sqrt{p}} + \frac{L\Cg}{\sqrt{pKb}}
    +\left(\frac{\Cg\zeta KT_3^\frac12}{\sqrt{p}}+\frac{\Cg LK^\frac12T_3^\frac12}{\sqrt{pb}}\right)3\eta\gamma(1+\eta\gamma)^{(t-\tau_0-1)K+K}+\frac{2\Cg T_3^\frac12\delta}{\Ce\sqrt{pK}}(1+\eta\gamma)^{(t-\tau_0-1)K+K}
    \right]r_e
    \\ & =
    \left[
    \frac{\zeta \Cg}{\sqrt{p}} + \frac{L\Cg}{\sqrt{pKb}}
    +\left(\frac{C_1^\frac12\Cg\zeta P^\frac12K}{p}+\frac{C_1^\frac12\Cg L\sqrt{PK}}{p\sqrt{b}}\right)3\eta\gamma(1+\eta\gamma)^{(t-\tau_0-1)K+K}+\frac{2C_1^\frac12\Cg\sqrt{P}\delta}{\Ce p}(1+\eta\gamma)^{(t-\tau_0-1)K+K}
    \right]r_e
\end{align}
with probability at least $1-\frac{\nu}{TK}$ for all $t$.
Taking $p\geq \sqrt{P}+\frac{\Ce^2\zeta^2}{\delta^2}+\frac{\Ce^2L^2}{\delta^2 Kb}$, $\eta=\Theta(\frac1L)$, $b \geq K, K = O\left(\frac{L}{\zeta}\right)$,  
and $\|g^{t+1}\|\leq \frac{\Ch\gamma}{\Ce}(1+\eta\gamma)^{(t-\tau_0)K}$ with sufficiently large constant $\Ch$, that only depends on $C_1$, $\Cg$, and sufficiently small $\eta = \tilde{\Theta}(\frac1L)$.


Moreover, \cref{lemma:D-Second-Auxiliary-B} states that, for $k<K$,
\begin{align}
    &\left\|h^{t,k+1}\right\|
    \\ &\leq \zeta\sum_{l=1}^{k+1}\|w^{t,l}-w^{t,l-1}\|+\frac{2\delta}{\Ce}\|w^{t,k+1}\|+\frac{2\delta}{\Ce}\|w^{t,0}\|+\frac{C_1^2}{\sqrt{b}}\sqrt{
    \sum_{l=1}^{k+1}\left(L\|w^{t,l}-w^{t,l-1}\|+\frac{2\delta}{\Ce}\|w^{l,k}\|+\frac{2\delta}{\Ce}\|w^{t,l-1}\|\right)^2
    }
\end{align}
holds with probability at least $1-\frac{\nu}{TK}$.
If (a) and (b) hold for all $(s,l)\leq (t,k+1)$, then we have
\begin{align}
     \left\|h^{t,k+1}\right\| & \leq
    3 \zeta K \eta \gamma (1+\eta\gamma)^{(t-\tau_0-1)K+k+1}
     +
     \frac{8\delta}{\Ce} (1+\eta\gamma)^{(t-\tau_0-1)K+k+1}
     \\ & \hspace{5mm}+
     \frac{3C_1^2\sqrt{K}}{\sqrt{b}}L\eta\gamma  (1+\eta\gamma)^{(t-\tau_0-1)K+k+1}
     +
     \frac{8C_1^2\sqrt{K}\delta}{\sqrt{b}} (1+\eta\gamma)^{(t-\tau_0-1)K+k+1}
     .
\end{align}
Taking $b\geq K$ and $K = O\left(\frac{L}{\zeta}\right)$, with sufficiently large $\Ch$ and sufficiently small $\eta$, we have $\|h^{t,k+1}\|\leq \frac{\Ch\gamma}{\Ce}(1+\eta\gamma)^{(t-\tau_0-1)K+k+1}$.

Thus, we obtain that (c) holds for $(t,k+1)$.

Therefore, we have completed the induction step and have $\frac12(1+\eta\gamma)^{(t-\tau_0-1)K+k}r_e\leq \|w^t\|$ for all $(\tau_0+1,0) \leq (t,k)< (\tau_0+1,T_5)$ with $T_5=\frac{\Cf\log \frac{\delta }{ \Ce \rho r_e}}{\eta \gamma}$.
Taking $\Cf$ sufficiently large, we have $\frac12(1+\eta\gamma)^{(\tau_0+1-\tau_0-1)K+T_5}r_e\geq \frac{\delta}{\Ce\rho}$.
This yields contradiction against the assumption and the desired assertion follows.

\end{proof}

From \cref{lemma:D-Small-Stuck-Region}, we can show that \AlgDist\ escapes saddle points with high probability.
We have the following lemma, and the proof is essentially the same as that of \cref{lemma:Escape-Small-Stuck-Region}.
\begin{lemma}\label{lemma:D-Escape-Small-Stuck-Region}
Let $\{x^{t,k}\}$ be a sequence generated by \AlgDist\ and $(\tau_0,\kappa_0) \ (0\leq \kappa_0<K)$ be a step where $-\gamma :=\lambda_{\rm min}(\nabla^2 f(x^{\tau_0,\kappa_0}))\leq -\delta$ holds.
We take $p\geq \sqrt{P}+\frac{\Ce^2\zeta^2}{\delta^2}+\frac{\Ce^2L^2}{Kb\delta^2}, b\geq \sqrt{K}$ and, $\eta = \tilde{\Theta}\left(\frac{1}{L}\right)$, and $T_5=\frac{\Cf\log \frac{\delta }{ \Ce \rho r_e}}{\eta \gamma}\lesssim \tilde{O}\left(\frac{L}{\delta}\right)$, with sufficiently large $\Ce,\Cf=\tilde{O}(1)$.
Then, 
\begin{align}
    \mathbb{P}\left[\max_{(\tau_0,\kappa_0)\leq (t,k)< (\tau_0 +1 ,T_5)}\|x^{t,k}-x^{\tau_0,\kappa_0+1}\|\geq \frac{\delta}{\Ce\rho}\mid I^{0},\cdots,I^\tau,i_{0},\cdots,i_{\tau_0},\xi^{0,0},\cdots,\xi^{\tau_0,\kappa_0}\right]
    \geq 1-\frac{4\nu}{TK}.
\end{align}
\end{lemma}

Finally, we show the main theorem of this subsection, which guarantees that the algorithm finds $(\eps,\delta)$-second-order stationary point with high probability.

\begin{proof}[Proof of \cref{theorem:Second-Main}]
Since $T_5=\frac{\Cf\log \frac{\delta }{\Ce \rho r_e}}{\eta \gamma }$ depends on $x^{\tau_0}$, we take $T_5=\frac{\Cf\log \frac{\delta }{ \Ce \rho r_e}}{\eta \delta }$ from now instead.
This change does not affect whether \cref{lemma:D-Escape-Small-Stuck-Region} holds.
Also, we let $T_6 = \lceil 1 + \frac{T_5}{K}\rceil$.

We divide $\{t=0,1,\cdots,T-1\}$ into the following $\lfloor\frac{T}{2T_6} \rfloor$ \textit{phases}: $P^\tau = \{2\tau T_6\leq t < 2(\tau+1)T_6\}\ \left(\tau=0,\cdots,\lfloor\frac{T}{2T_6} \rfloor-1\right)$.
For each phase, we define $a^\tau$ as a random variable taking values
\begin{align}
    a^\tau = 
    \begin{cases}
    1 & \left(\text{if $\sum_{t\in P^\tau}\sum_{k=0}^K\mathbbm{1}[\|\nabla f(x^{t,k})\|>\eps]> KT_6$}\right)\\
    2 & \left(\text{if there exists $t$ such that $(2\tau T_6,0) \leq (t,k)< ((2\tau+1) T_6,0)$, $\|\nabla f(x^{t,k})\|\leq \eps$ and $\lambda_{\rm min}(\nabla^2 f(x^{t,k}) ) \leq -\delta$}\right)\\
    3 & \left(\text{if there exists $t$ such that $(2\tau T_6,0) \leq (t,k)< ((2\tau+1) T_6,0)$, $\|\nabla f(x^{t,k})\|\leq \eps$ and $\lambda_{\rm min}(\nabla^2 f(x^{t,k}) ) >-\delta$}\right).
    \end{cases}
\end{align}
Note that $\mathbb{P}[a^\tau = 1, 2, 3]=1$ for each $\tau$.
This is because if there does not exist $t$ between $(2\tau T_6,0) \leq (t,k)< ((2\tau+1)T_6,0)$ such that $\|\nabla f(x^{t,k})\|\leq \eps$ (i.e., neither $a^\tau=2$ nor $3$), then we have $\sum_{t\in P^\tau}\sum_{k=0}^K\mathbbm{1}[\|\nabla f(x^{t,k})\|>\eps]\geq \sum_{t=2\tau T_2}^{(2\tau+1)T_6-1}\sum_{k=0}^K\mathbbm{1}[\|\nabla f(x^{t,k})\|>\eps]=T_6K$, meaning $a^\tau = 1$.
We denote $N_1 = \sum_{\tau=0}^{\lfloor\frac{T}{2T_6}\rfloor}\mathbbm{1}[a^\tau = 1]$, $N_2 = \sum_{\tau=0}^{\lfloor\frac{T}{2T_6}\rfloor}\mathbbm{1}[a^\tau = 2]$, and $N_3 = \sum_{\tau=0}^{\lfloor\frac{T}{2T_6}\rfloor}\mathbbm{1}[a^\tau = 3]$.

According to \cref{lemma:D-Escape-Small-Stuck-Region}, with probability $1-4\nu$ over all $\tau$, it holds that if $a^\tau = 2$ then that phase successes escaping saddle points; i.e., there exists $(2\tau T_6,0)\leq (t,k)<((2\tau+1)T_6,0)$ and
\begin{align}
    \max_{(t,k)\leq (s,l)< ((2\tau+2)T_6,0)}\|x^{s,l}-x^{t,k}\|> \frac{\delta}{\Ce\rho}
    \label{eq:D-Second-SmallStuckRegion-31}
\end{align}
holds.
Eq. \eqref{eq:D-Second-SmallStuckRegion-31} further leads to
\begin{align}
    T_6K\sum_{t=2\tau T_6}^{2(\tau+1) T_6-1}\sum_{k=1}^K\|x^{t,k}-x^{t,k-1}\|^2
    > \left(\frac{\delta}{\Ce\rho}\right)^2
    \Leftrightarrow
    \sum_{t=2\tau T_6}^{2(\tau+1) T_6-1}\sum_{k=1}^K\|x^{t,k}-x^{t,k-1}\|^2
    > \frac{\delta^2}{T_6K\Ce^2\rho^2}. 
    \label{eq:D-Second-SmallStuckRegion-32}
\end{align}

On the other hand, in \cref{theorem:Dist-First-Main}, we derived that 
\begin{align}
& \sum_{t=1}^T\sum_{k=1}^{K} \|\nabla f(x^{t,k-1})\|^2 \\ & \leq
\frac{2}{\eta}\left(f(x^0)-f(x^{T})\right)\\ & \hspace{5mm} -
\frac{2}{\eta}\sum_{t=1}^T\sum_{k=1}^{K}\left(\frac{1}{2\eta}-\frac{L}{2}-\eta \left(\frac{120C_1^9\zeta^2PK^2}{p^2}+\frac{128C_1^{11}\zeta^2PK}{p^2b}+4\zeta^2K^2+\frac{4C_1^2L^2K}{b}\right)\right)\|x^{t,k}-x^{t,k-1}\|^2
    \\ & \hspace{5mm}+\begin{cases}
    \displaystyle
    \frac{16C_1^2T}{Pb}\left(\sigma^2 + \frac{G^2}{PKb}\right)+\frac{2TKr^2}{\eta^2}+
         192C_1^3\left(\frac{\sigma^2P}{p^2b} + \frac{PG^2}{p^3Kb^2}+\frac{PK\sigma^2_c}{p^2} + \frac{PKG^2_c}{p^3}\right)&\quad (\text{\rm Option I})
    \\\displaystyle
     \frac{16C_1^2T}{Pb}\left(\sigma^2 + \frac{G^2}{PKb}\right)+\frac{2r^2}{\eta^2}&\quad (\text{\rm Option II})
    \end{cases}
\end{align}
with probability $1-8\nu$.
Taking $\eta=\tilde{\Theta}\left(\frac1L\right)$ sufficiently small, applying $p\geq \sqrt{P}$, $K= O\left(\frac{L}{\zeta}\right)$, $K\leq b$ and $f(x^0)-f(x^t)\leq \Delta$, and arranging terms yields
\begin{align}
& \sum_{t=1}^T\sum_{k=1}^K \|\nabla f(x^{t,k-1})\|^2 
+\frac{1}{2\eta^2}\sum_{t=1}^T\sum_{k=1}^K\|x^{t,k}-x^{t,k-1}\|^2
\label{eq:D-Second-SmallStuckRegion-34}
\\ & \leq\frac{2\Delta}{\eta}+
\begin{cases}
    \displaystyle
    \frac{16C_1^2T}{Pb}\left(\sigma^2 + \frac{G^2}{PKb}\right)+\frac{2TKr^2}{\eta^2}+
         192C_1^3\left(\frac{\sigma^2}{b} + \frac{G^2}{pKb^2}+K\sigma^2_c+ \frac{KG^2_c}{p}\right)&\quad (\text{\rm Option I})
    \\\displaystyle
     \frac{16C_1^2T}{Pb}\left(\sigma^2 + \frac{G^2}{PKb}\right)+\frac{2TKr^2}{\eta^2}&\quad (\text{\rm Option II})
    \end{cases}
\label{eq:D-Second-SmallStuckRegion-33}
\end{align}
From the definition of $a^\tau = 1$ and \eqref{eq:D-Second-SmallStuckRegion-32}, We know that \eqref{eq:D-Second-SmallStuckRegion-34} is bounded as
\begin{align}
   \sum_{t=1}^T\sum_{k=1}^K \|\nabla f(x^{t,k-1})\|^2 
+\frac{1}{2\eta^2}\sum_{t=1}^T\sum_{k=1}^K\|x^{t,k}-x^{t,k-1}\|^2
& \geq 
N_1T_6K\eps^2
+
\frac{\delta^2N_2}{2\eta^2T_6K\Ce^2\rho^2}.
\end{align}
Thus, $N_1T_6K \leq \frac{1}{\eps^2}\times (\text{the right-hand side of \eqref{eq:D-Second-SmallStuckRegion-33}})$ and
$N_2T_6K \leq \frac{2\eta^2\Cb^2\rho^2K^2T_6^2}{\delta^2}\times (\text{the right-hand side of \eqref{eq:Second-SmallStuckRegion-33}})$ holds.

Here, $\frac{2\eta^2T_6^2K^2\Ce^2\rho^2}{\delta^2} = \tilde{O}\left(\frac{\rho^2}{\delta^4}+\frac{\eta^2 K^2}{\delta^2}\right)\lesssim \tilde{O}\left(\frac{\rho^2}{\delta^4}\right)$, when $K=O\left(\frac{L}{\zeta}\right)\leq O\left(\frac{L}{\delta}\right)$.
From this, $(N_1 + N_2)T_6\leq \tilde{O}\left(\frac{1}{K\eps^2}+\frac{\rho^2}{K\delta^4}\right)\times (\text{the right-hand side of \eqref{eq:D-Second-SmallStuckRegion-33}})$.
Taking $T\geq 2(N_1+N_2+1)T_6$, there exists $\tau$ such that $a^\tau =3$, which concludes the proof.
\end{proof}

\subsection{Finding Second-Order Stationary Points When Clients are Homogeneous ($\zeta\ll \frac{1}{\delta}$)}

In the previous subsection, we assumed that $\zeta\geq \frac{1}{\delta}$.
Here, we introduce a simple trick to remove this assumption and give its convergence analysis.

Let $T_7=\tilde{\Theta}\left(\frac{L}{\delta}\right)$ with a sufficiently large hidden constant. 
In line $18$-$19$ of \AlgDist, when $k\equiv T_7$, we randomly select $\frac{\Ce^2L^2}{\delta^2}+b$ (not $b$) samples $J_{i_t}^{t,k}$, and update $z^{t,k}$ as
$z^{t,k}\leftarrow z^{t,k-1}+\frac{1}{|J_{i_t}^{t,k}|}\sum_{j\in J_{i_t}^{t,k}} (\nabla f_{i_t,j}(x^{t,k})-\nabla f_{i_t,j}(x^{t,k-1}))$. 
This increases the number of gradient evaluations in each inner-loop by $\tilde{O}(K/(L/\delta))\times \tilde{O}(L^2/\delta^2)=\tilde{O}(KL/\delta)\lesssim \tilde{O}(K^2)\lesssim \tilde{O}(Kb)$.
Thus, this does not affect the inner-loop complexity more than by constant factors.

Then, the following lemma holds, which stands as generalization of \cref{lemma:D-Small-Stuck-Region}.
\begin{lemma}[Small stuck region]\label{lemma:D-Small-Stuck-Region-Relax}
Let $\{x^{t,k}\}$ be a sequence generated by \AlgDist\ and $(\tau_0, \kappa_0)$ be a step where $-\gamma :=\lambda_{\rm min}(\nabla^2 f(x^{\tau_0,\kappa_0}))\leq -\delta$ holds.
We denote the smallest eigenvector direction of $\lambda_{\rm min}(\nabla^2 f(x^{\tau_0,\kappa_0}))$ as $\mathsf{e}$.
Moreover, we define a coupled sequence $\{\tilde{x}^{t,k}\}$ by running \AlgDist\ with $\tilde{x}^{0} = x^0$ and the same choice of all randomness \textit{i.e.}, client samplings, minibatches and noises, but the noise at some step $(\tau,\kappa)> (\tau_0,\kappa_0)$, satisfying $\kappa \equiv T_7$\rm{;} We let $\tilde{\xi}^{\tau,\kappa}
= \xi^{\tau,\kappa} - r_e \mathsf{e}$ with $r_e \geq \frac{r\nu }{TK\sqrt{d}}$.
Let $w^{t,k} = x^{t,k} - \tilde{x}^{t,k}$, $w^{t} = x^{t} - \tilde{x}^{t}$, $v^{t,k} = \frac{1}{P}\sum_{i=1}^P y_i^{t-1}+z^{t,k}$, $\tilde{v}^t = \frac{1}{P}\sum_{i=1}^P\tilde{y}_i^{t-1}+z^{t,k}$, $g^t = \frac{1}{P}\sum_{i=1}^P y_i^{t}-\nabla f(x^t)- \left(\frac{1}{P}\sum_{i=1}^P \tilde{y}_i^{t}-\nabla f(\tilde{x}^t)\right)$, and $h^{t,k} = (z^{t,k} - (\nabla f(x^{t,k})-\nabla f(x^{t,0})))-(\tilde{z}^{t,k} - (\nabla f(\tilde{x}^{t,k})-\nabla f(\tilde{x}^{t,0})))$.
Then, 
$v^{t,k} - \nabla f(x^{t,k}) - (\tilde{v}^{t,k} - \nabla f(\tilde{x}^{t,k}))=g^{t-1}+h^{t,k}$.

There exists a sufficiently large constants $\Ce=\tilde{O}(1),\Cf=O(1)$, with which the following holds:
If we take $p\geq \sqrt{P}+\frac{\Ce^2\zeta^2}{\delta^2}+\frac{\Ce^2L^2}{Kb\delta^2}, b\geq \sqrt{K}$ and, $\eta = \tilde{\Theta}\left(\frac{1}{L}\right)$, with probability $1-\frac{3\nu}{TK}~(\nu \in (0,1))$, we have
\begin{align}
    \max_{(\tau_0,\kappa_0)\leq  (t,k)< (\tau_0,\kappa_0+3T_7)}\{\|x^{\tau,k}-x^{\tau_0,\kappa_0}\|,\|\tilde{x}^{\tau,\kappa}-x^{\tau_0,\kappa_0}\|\} \geq  \frac{\delta}{\Ce\rho}
    .
\end{align}
\end{lemma}
\begin{proof}[Proof of \cref{lemma:D-Small-Stuck-Region-Relax}]
We assume $K$ is at least as large as $3T_7$.
When $K-2T_7 \leq \kappa_0< K-1$, 
taking $T_7\geq T_5$ yields the assertion, considering the two coupled sequence initialized at $(\kappa_0, K)$, according to a slight modification of \cref{lemma:D-Small-Stuck-Region}.

Otherwise, we let $(\tau,\kappa)$ as the first step after $(\tau_0,\kappa_0)$ with 
$\kappa \equiv T_7 $.
Then, it suffice to show that, with probability at least $1-\frac{3\nu}{TK}$, 
\begin{align}
    \max_{(\tau,\kappa)\leq  (t,k)< (\tau,\kappa+T_7)}\{\|x^{\tau,k}-x^{\tau_0,\kappa_0}\|,\|\tilde{x}^{\tau,\kappa}-x^{\tau_0,\kappa_0}\|\} \geq  \frac{\delta}{\Ce\rho}
    .
    \label{eq:D-Small-Stuck-Region-Relax-1}
\end{align}

Since $K\geq 3T_7$ and $\kappa_0<K-2T_7$ imply $g^{t-1}=0$ for all $(\tau,\kappa)\leq (t,k)< (\tau,\kappa+T_7)$, $g^{t-1}+h^{t,k}=h^{t,k}$ holds.
Then, $\|h^{\tau,\kappa}\| = \left\|u_{i_t}^{\tau,\kappa}+\frac{1}{|J^{\tau,\kappa}_{i_t}|}\sum_{j\in J^{\tau,\kappa}_{i_t}}u_{i_t,j}^{\tau,\kappa}\right\|\leq \zeta r_e + \frac{L}{\sqrt{|J^{\tau,\kappa}_{i_t}|}}r_e \leq 2\delta r_e$, using \cref{proposition:Bernstein}.
Moreover, for $(\tau,k)>(\tau,\kappa)$, when we assume $\max_{(\tau,\kappa)\leq  (t,k)< (\tau,\kappa+T_7)}\{\|x^{\tau,k}-x^{\tau_0,\kappa_0}\|,\|\tilde{x}^{\tau,\kappa}-x^{\tau_0,\kappa_0}\|\}<  \frac{\delta}{\Ce\rho}$,
\begin{align}
    &\|h^{\tau,k}\|\\ &= \left\|u_{i_\tau}^{\tau,k}+\sum_{l=\kappa}^k\frac{1}{|J^{\tau,l}_{i_t}|}\sum_{j\in J^{\tau,\kappa}_{i_t}}u_{i_t,j}^{\tau,l}\right\|\\ &\leq \zeta \sum_{l=\tau}^k\|w^{\tau,l}-w^{\tau,l-1}\|+\frac{2\delta}{\Ce}\|w^{\tau,k}\|+\frac{2\delta}{\Ce}\|w^{\tau,0}\|
    +\delta r_e
    +\frac{C_1^2}{\sqrt{b}}\sqrt{
    \sum_{l=1}^k\left(L\|w^{\tau,k}-w^{\tau,k-1}\|+\frac{2\delta}{\Ce}\|w^{\tau,k}\|+\frac{2\delta}{\Ce}\|w^{\tau,k-1}\|\right)^2
    }.
\end{align}
Assuming that ${\rm(a)}\quad \frac12(1+\eta\gamma)^{k-\kappa}r_e\leq \|w^{t,k}\|\leq 2(1+\eta\gamma)^{k-\kappa}r_e$ and $
    {\rm(b)}\quad \|w^{t,k} - w^{t,k-1}\|\leq 
   3\eta\gamma (1+\eta\gamma)^{k-\kappa}r_e$ 
   for $(\tau,\kappa)<(t,k)<(\tau,\kappa+T_7)$, we get $\|h^{t,k}\|\leq \frac{2\Ch\gamma}{\Ce}(1+\eta\gamma)^{k-\kappa}$.
Thus, following the discussion in \cref{lemma:D-Small-Stuck-Region} and taking $T_7$ similarly to $T_5$, we have \eqref{eq:D-Small-Stuck-Region-Relax-1}.
\end{proof}
Previously, we only focused on the noise at the last local step $(\kappa_0, K)$.
Thus, if the number of steps required to escape saddle points $T_5=\tilde{O}(\frac{L}{\delta})$ is smaller than the local steps $K=\tilde{O}(\frac{L}{\zeta})$, the algorithm sometimes have to wait more than $O(T_5)$ steps for the last local step.
Therefore, taking $K\geq T_5$ was useless to reduce the number of communication rounds.
On the other hand, based on \cref{lemma:D-Small-Stuck-Region-Relax}, when FLEDGE comes to a saddle point, FLEDGE does not need to wait next communication, and can escape the stack region within $2T_7$ local steps, even if $T_7\ll K$.
This allows to us to take $K$ larger than $O(\frac{L}{\delta})$, and leads to removal of the assumption $\delta<\frac{1}{\zeta}$ from \cref{theorem:D-Second-Main}'.

\subsection{Convergence under PL condition}

\newtheorem*{theorem:FLEDFEPL}{\rm\bf Theorem~\ref{theorem:D-PL}'}
\begin{theorem:FLEDFEPL}
Under \cref{assumption:GradLipschitz,assumption:GlobalInfimum,assumption:BoundedGradient,assumption:Heterogeneity,assumption:PL}, if we choose
$\eta = \tilde{\Theta}\left(
    \frac{1}{L}\land
    \frac{p\sqrt{b}}{\zeta \sqrt{PK}}\land
    \frac{p}{\mu PK}\land
    \frac{1}{\zeta K}\land
    \frac{\sqrt{b}}{L\sqrt{K}}
    \right),\ PKb \geq \Omega\left(\frac{C_1^2\sigma^2}{\eps^2}+\frac{C_1G}{\eps}\right) \text{ and } r \leq \frac{\eps\sqrt{\eta}}{8}
$,
\cref{alg:dist-Appendix} with Option I finds an $\eps$-first-order stationary points for problem \eqref{eq:Intro-Federated} using
\begin{align}
    \tilde{O}\left(PKb + \left(
    \frac{Lpb}{\mu}\land
    \frac{\zeta \sqrt{P}Kb}{\mu}\land
    PKb\land
    \frac{\zeta pKb}{\mu}\land
    \frac{Lp\sqrt{Kb}}{\mu}
    \right)\log \frac{\Delta+\sigma+G+\sigma_c+G_c}{\eps}\right) \quad \text{\rm stochastic gradients and}\\ \quad
\tilde{O}\left(\frac{P}{p} + \left(
    \frac{L}{\mu K}\land
    \frac{\zeta \sqrt{P}}{\mu p}\land
    \frac{ P}{p}\land
    \frac{\zeta }{\mu}\land
    \frac{L}{\mu\sqrt{Kb}}
    \right)\log \frac{\Delta+\sigma+G+\sigma_c+G_c}{\eps}\right)
    \quad \text{\rm communication rounds}
\end{align}
with probability at least $1-8\nu$.
Moreover, under the same conditions, \cref{alg:dist-Appendix} with Option II finds an $\eps$-first-order stationary points for problem \eqref{eq:Intro-Federated} using
\begin{align}
    \tilde{O}\left(PKb + \left(
    \frac{Lpb}{\mu}\land
    \frac{\zeta \sqrt{P}Kb}{\mu}\land
    PKb\land
    \frac{\zeta pKb}{\mu}\land
    \frac{Lp\sqrt{Kb}}{\mu}
    \right)\log \frac{\Delta}{\eps}\right)
 \quad \text{\rm stochastic gradients and}\\
 \tilde{O}\left(1+ \left(
    \frac{L}{\mu K}\land
    \frac{\zeta \sqrt{P}}{\mu p}\land
    \frac{ P}{p}\land
    \frac{\zeta }{\mu}\land
    \frac{L}{\mu\sqrt{Kb}}
    \right)\log \frac{\Delta}{\eps}\right)
    \quad \text{\rm communication rounds}
\end{align}
with probability at least $1-8\nu$.
Here $\tilde{O}$ hides only at most $\log^{6.5} (P+K+\mu^{-1}+\nu^{-1})$ and ${\rm polyloglog}$ factors.
\end{theorem:FLEDFEPL}
\begin{proof}
According to \cref{eq:Dist-First-FunctionDecrease-1} and PL condition,
\begin{align}
    f(x^{t,k})
    & \leq
    f(x^{t,k-1}) + \eta \|\nabla f(x^{t,k-1}) - v^{t,k-1}\|^2
    - \frac{\eta}{2}\|\nabla f(x^{t,k-1})\|^2
    - \left(\frac{1}{2\eta}-\frac{L}{2}\right)\|x^{t,k}-x^{t,k-1}\|^2
    + \frac{r^2}{\eta}
    \\ & \leq 
    f(x^{t,k-1}) + \eta \|\nabla f(x^{t,k-1}) - v^{t,k-1}\|^2
    \\
    &\hspace{20mm}- \frac{\eta}{4}\|\nabla f(x^{t,k-1})\|^2
    - \frac{\eta\mu}{2}(f(x^{t,k-1})-f^*)
    - \left(\frac{1}{2\eta}-\frac{L}{2}\right)\|x^{t,k-1}-x^{t,k-1}\|^2
    + \frac{r^2}{\eta}.   
\end{align}
Rearranging the above yields that
\begin{align}\label{eq:Dist-PL-Main-1}
    &f(x^{t,k})-f^* + \frac{\eta}{4}\|\nabla f(x^{t,k-1})\|^2
    \\&\leq \left(1-\frac{\eta\mu}{2}\right)(f(x^{t,k-1})-f^*) + \eta \|\nabla f(x^{t,k-1}) - v^{t,k-1}\|^2    - \left(\frac{1}{2\eta}-\frac{L}{2}\right)\|x^{t,k}-x^{t,k-1}\|^2
    + \frac{r^2}{\eta}
    .
\end{align}
holds for all $t,k$ $(1\leq t\leq T, 0\leq k\leq K-1)$ with probability at least $1-8\nu$.

Applying \cref{lemma:Dist-First-VarianceBound} to this, for all $t=1,\cdots,T$ with probability at least $1-8\nu$,
\begin{align}
    &f(x^t)-f(x^*)+\frac{\eta}{4}\sum_{k=1}^{K}(1-\frac{\eta\mu}{2})^{K-k}\|\nabla f(x^{t,k-1})\|^2
    \\ &\leq  (1-\frac{\eta\mu}{2})^K(f(x^{t-1})-f(x^*))     - \sum_{k=1}^K(1-\frac{\eta\mu}{2})^{K-k}\left(\frac{1}{2\eta}-\frac{L}{2}\right)\|x^{t,l}-x^{t,l-1}\|^2
    \\ & \hspace{5mm}+ 
     \eta \sum_{k=1}^K(1-\frac{\eta\mu}{2})^{K-k}\left(\left(\frac{120C_1^8\zeta^2K}{p}+\frac{32C_1^{10}\zeta^2}{pb}\right) \sum_{s=\max\{1,t-T_3\}}^{t-1}\sum_{l=1}^K\|x^{s,l}-x^{s,l-1}\|^2
     \right.
     \\ & \hspace{90mm}\left.+ \left(4\zeta^2K+\frac{4C_1^2L^2}{b}\right)\sum_{l=1}^{K-1}\|x^{t,l}-x^{t,l-1}\|^2
     \right)
    \\ & \hspace{5mm}+
    \sum_{k=1}^K(1-\frac{\eta\mu}{2})^{K-k}
    \left(\frac{r^2}{\eta}+\frac{8C_1^2\eta }{PKb}\left(\sigma^2 + \frac{G^2}{PKb}\right)\right)
    \\ & \hspace{5mm}+
    \begin{cases}
    \displaystyle
         \eta\sum_{k=1}^K(1-\frac{\eta\mu}{2})^{K-k} 96C_1^2\mathbbm{1}[t\leq T_3]\left(\frac{\sigma^2}{pKb} + \frac{G^2}{p^2K^2b^2}+\frac{\sigma^2_c}{p} + \frac{G^2_c}{p^2}\right)
    &\quad (\text{Option I})\\\displaystyle
    0
&\quad (\text{Option II})
    \end{cases}
\end{align}
By using this bound repeatedly, we get
\begin{align}
    &f(x^T)-f(x^*)+\frac{\eta}{4}\sum_{t=1}^T\sum_{k=1}^{K}(1-\frac{\eta\mu}{2})^{(T-t+1)-k}\|\nabla f(x^{t,k-1})\|^2
    \\ &\leq  (1-\frac{\eta\mu}{2})^{TK}(f(x^{0})-f(x^*))   
\\ & \hspace{5mm} - \sum_{t=1}^T\sum_{k=1}^K(1-\frac{\eta\mu}{2})^{(T-t+1)K-k}\left(\frac{1}{2\eta}-\frac{L}{2}
    - \sum_{s=1}^{T_3}(1-\frac{\eta\mu}{2})^{-(s+1)K}\left(\frac{120C_1^8\zeta^2\eta K^2}{p}+\frac{32C_1^{10}\zeta^2\eta K}{pb}\right)
    \right.\\&\hspace{5mm}\left.-\eta\sum_{l=1}^K(1-\frac{\eta\mu}{2})^{l-K}\left(4\zeta^2K+\frac{4C_1^2L^2}{b}\right)
    \right)\|x^{t,k}-x^{t,k-1}\|^2
    \\ & \hspace{5mm}+ \sum_{t=1}^T\sum_{k=1}^K(1-\frac{\eta\mu}{2})^{(T-t+1)K-k}
    \left(\frac{r^2}{\eta}+\frac{8C_1^2\eta }{PKb}\left(\sigma^2 + \frac{G^2}{PKb}\right)\right)
    \\ & \hspace{5mm}+
    \begin{cases}
    \displaystyle
         \eta\sum_{t=1}^T\sum_{k=1}^K(1-\frac{\eta\mu}{2})^{(T-t+1)K-k}96C_1^2\mathbbm{1}[t\leq T_3]\left(\frac{\sigma^2}{pKb} + \frac{G^2}{p^2K^2b^2}+\frac{\sigma^2_c}{p} + \frac{G^2_c}{p^2}\right)
    &\quad (\text{Option I})\\\displaystyle
    0
&\quad (\text{Option II}).
    \end{cases}
\end{align}

We take $\eta$ as
\begin{align}
    \eta = \Theta\left(
    \frac{1}{L}
    \land
    \frac{p}{C_1^{4.5}\zeta \sqrt{P}K}
    \land
    \frac{p\sqrt{b}}{C_1^{5.5}\zeta \sqrt{PK}}\land
    \frac{p}{\mu C_1 PK}\land
    \frac{1}{\zeta K}\land
    \frac{\sqrt{b}}{C_1L\sqrt{K}}
    \right)
\end{align}
so that $\frac{1}{2\eta}-\frac{L}{2}
    - \sum_{s=1}^{T_3}(1-\frac{\eta\mu}{2})^{-(s+1)K}\left(\frac{120C_1^8\zeta^2\eta K^2}{p}+\frac{32C_1^{10}\zeta^2\eta K}{pb}\right)
    -\eta\sum_{k=1}^K(1-\frac{\eta\mu}{2})^{k-K}\left(4\zeta^2K+\frac{4C_1^2L^2}{b}\right)\geq 0$ holds. 
We also take $r\leq \frac{\eps\sqrt{\eta}}{8}$ and $PKb \geq \frac{512C_1^2\sigma^2}{\eps^2}+\frac{64C_1G}{\eps}$, then 
$\sum_{t=1}^T\sum_{k=1}^K(1-\frac{\eta\mu}{2})^{(T-t+1)K-k}
    \left(\frac{r^2}{\eta}+\frac{8C_1^2\eta }{PKb}\left(\sigma^2 + \frac{G^2}{PKb}\right)\right)\leq \frac{\eps^2}{8\mu}$ holds.

Then, we have that
\begin{align}
    &f(x^T)-f(x^*)+\frac{\eta}{4}\sum_{t=1}^T\sum_{k=1}^{K}(1-\frac{\eta\mu}{2})^{(T-t+1)-k}\|\nabla f(x^{t,k-1})\|^2 \\&\leq \frac{\eps^2}{8}+
    (1-\frac{\eta\mu}{2})^{TK}(f(x^0)-f^*)\\ & \hspace{5mm}+\begin{cases}
    \displaystyle
         (1-\frac{\eta\mu}{2})^{(T-t+1-T_3)K-k}96C_1^3\left(\frac{\sigma^2P}{p^2b} + \frac{G^2P}{p^3Kb^2}+\frac{\sigma^2_cPK}{p^2} + \frac{G^2_cPK}{p^3}\right)
    &\quad (\text{Option I})\\\displaystyle
    0
&\quad (\text{Option II})
    \end{cases}
\end{align}

For Option I, the first term $(1-\frac{\eta\mu}{2})^{TK}(f(x^0)-f(x^*))$ is smaller than $\frac{\eps^2}{32}$ if we take $TK = O \left(\frac{1}{\eta\mu}\log \frac{\Delta}{\eps}\right)$.
The third term is bounded by $\frac{\eps^2}{32}$, if we take $T = T_3 + O \left(\frac{1}{\eta\mu K}\log \frac{ C_1^3\left(\frac{\sigma^2P}{p^2b} + \frac{G^2P}{p^3Kb^2}+\frac{\sigma^2_cPK}{p^2} + \frac{G^2_cPK}{p^3}\right)}{\eps}\right)=O\left(\frac{P}{p}C_1+ \frac{C_1}{\eta\mu K}\log \frac{\sigma+G+\sigma_c+G_c}{\eps}\right)$. 
Moreover, note that $f(x^T)-f(x^*)+\frac{\eta}{4}\sum_{t=1}^T\sum_{k=1}^{K}(1-\frac{\eta\mu}{2})^{(T-t+1)-k}\|\nabla f(x^{t,k-1})\|^2\leq \frac{6}{32\mu}\min_{t,k}\|\nabla f(x^{t,k-1})\|^2$ holds when we take $T = O\left(\frac{1}{\eta\mu K}\right)$.

Thus, for Option I, if we take 
 \begin{align}T=O\left(\frac{P}{p}C_1 + C_1\left(
    \frac{L}{\mu K}\land
    \frac{C_1^{4.5}\zeta \sqrt{P}}{\mu p}\land
    \frac{C_1^{5.5}\zeta \sqrt{P}}{\mu p\sqrt{bK}}\land
    \frac{C_1 P}{p}\land
    \frac{\zeta }{\mu}\land
    \frac{C_1L}{\mu\sqrt{Kb}}
    \right)\log \frac{\Delta+\sigma+G+\sigma_c+G_c}{\eps}\right),
\end{align}we obtain the desired bound with probability at least $1-8\nu$.

For Option II, taking
\begin{align}T=O\left(\left(
    \frac{L}{\mu K}\land
    \frac{C_1^{4.5}\zeta \sqrt{P}}{\mu p}\land
    \frac{C_1^{5.5}\zeta \sqrt{P}}{\mu p\sqrt{bK}}\land
    \frac{C_1 P}{p}\land
    \frac{\zeta }{\mu}\land
    \frac{C_1L}{\mu\sqrt{ K b}}
    \right)\log \frac{\Delta}{\eps}\right),
\end{align}
yields the desired bound.

Note that $T$ depends on $\eps^{-1}$ only logarithmically, which means that $C_1$ depends on $\eps^{-1}$ in only $\log\log$ order and $C_1 = O^*(\log(P+K+\mu^{-1}+\nu^{-1}))$ (where $O^*$ suppresses $\log\log$ factors).
\end{proof}

\begin{remark}
In order to find $\eps$-solutions (i.e., $f(x^{t,k})-f^*\leq \eps$), the same statement holds, except for  slight change on
the assumptions on $PKb$ and $r$:
$PKb \geq \Omega\left(\frac{C_1^2\sigma^2}{\mu\eps}+\frac{C_1G}{\eps\sqrt{\eps\mu}}\right) \text{ and } r \leq \frac{\eta\sqrt{\eps\mu}}{2}
$.

In fact, we can derive
\begin{align}
    f(x^{t,k})-f^* 
    \leq \left(1-\eta\mu\right)(f(x^{t,k-1})-f^*) + \eta \|\nabla f(x^{t,k-1}) - v^{t,k-1}\|^2    - \left(\frac{1}{2\eta}-\frac{L}{2}\right)\|x^{t,k}-x^{t,k-1}\|^2
    + \frac{r^2}{\eta}
\end{align}
similarly to \eqref{eq:Dist-PL-Main-1}, and using this, we have
\begin{align}
    &f(x^t)-f(x^*)
    \\ &\leq  (1-\eta\mu)^{TK}(f(x^{0})-f(x^*))   \\ & \hspace{5mm}
 - \sum_{t=1}^T\sum_{k=1}^K(1-\eta\mu)^{(T-t+1)K-k}\left(\frac{1}{2\eta}-\frac{L}{2}
    - \sum_{s=1}^{T_3}(1-\eta\mu)^{-(s+1)K}\left(\frac{120C_1^8\zeta^2\eta K^2}{p}+\frac{32C_1^{10}\zeta^2\eta K}{pb}\right)
    \right.\\&\hspace{5mm}\left.-\eta\sum_{l=1}^K(1-\eta\mu)^{l-K}\left(4\zeta^2K+\frac{4C_1^2L^2}{b}\right)
    \right)\|x^{t,k}-x^{t,k-1}\|^2
    \\ & \hspace{5mm}+ \sum_{t=1}^T\sum_{k=1}^K(1-\eta\mu)^{(T-t+1)K-k}
    \left(\frac{r^2}{\eta}+\frac{8C_1^2\eta }{PKb}\left(\sigma^2 + \frac{G^2}{PKb}\right)\right)
    \\ & \hspace{5mm}+
    \begin{cases}
    \displaystyle
         \eta\sum_{t=1}^T\sum_{k=1}^K(1-\eta\mu)^{(T-t+1)K-k}96C_1^2\mathbbm{1}[t\leq T_3]\left(\frac{\sigma^2}{pKb} + \frac{G^2}{p^2K^2b^2}+\frac{\sigma^2_c}{p} + \frac{G^2_c}{p^2}\right)
    &\quad (\text{Option I})\\\displaystyle
    0
&\quad (\text{Option II})
    \end{cases}
\end{align}
Taking $r\leq \frac{\eta\sqrt{\eps\mu}}{2}$ and $PKb \geq \Omega\left(\frac{C_1^2\sigma^2}{\mu\eps}+\frac{C_1G}{\sqrt{\mu\eps}}\right)$ yields
$\sum_{t=1}^T\sum_{k=1}^K(1-\eta\mu)^{(T-t+1)K-k}
    \left(\frac{r^2}{\eta}+\frac{8C_1^2\eta }{PKb}\left(\sigma^2 + \frac{G^2}{PKb}\right)\right)\leq \frac{\eps}{2}$.
Thus, we finally have the following:
\begin{align}
    &f(x^t)-f(x^*) \\&\leq \frac{\eps}{2}+
    (1-\eta \mu)^{TK}(f(x^0)-f^*)+\begin{cases}
    \displaystyle
         (1-\eta\mu)^{(T-t+1-T_3)K-k}96C_1^3\left(\frac{\sigma^2P}{p^2b} + \frac{G^2P}{p^3Kb^2}+\frac{\sigma^2_cPK}{p^2} + \frac{G^2_cPK}{p^3}\right)
    &\quad (\text{Option I})\\\displaystyle
    0
&\quad (\text{Option II}).
    \end{cases}
\end{align}
Now it is trivial to see that the desired bound holds.
\end{remark}

\section{Lower bound}\label{section:lowerbound}
\Cref{proposition:LowerBound} can be derived by using the bounds of \citet{carmon2020lower,fang2018spider,li2021page}.
First, we give a definition of a linear-span first-order algorithm.
\begin{definition}[Linear-span first-order algorithm]
Fix some $x^0$. Let $\mathcal{A}$ be a (randomized) algorithm with the initial point $x^0$, and $x^t$ be the point at the $t$-th iteration.
We assume $\mathcal{A}$ select one individual function $i_t$ at each iteration $t$ and computes $\nabla f_{i_t}(x^t)$.
Then $\mathcal{A}$ is called a linear-span first-order algorithm if
\begin{align}
    x^t \in {\rm span}\{x^0, x^1, \cdots, x^{t-1}, \nabla f_{i_0}(x^0),\nabla f_{i_1}(x^1),\cdots,\nabla f_{i_{t-1}}(x^t)\}
\end{align}
holds for all $t$ with probability one.
\end{definition}
Note that this definition includes minibatch updade, by letting $x^{sb}=x^{sb+1}=,\cdots,=x^{(s+1)b-1}$ with the minibatch size $b$.

We also define problem classes $\mathcal{F}_{n,\Delta}^{L}$ and $\mathcal{F}_{n,\Delta}^{L,\zeta}$ for \eqref{eq:Intro-FiniteSum}, as follows.
\begin{definition}[A class of finite-sum optimization problems]
Fix some $x^0$. For an integer $n$, $L>0$, we define a problem class $\mathcal{F}_n^{L}$ as
\begin{align}
    \mathcal{F}_{n,\Delta}^{L} = \left\{ \left.
        f=\frac1n \sum_{i=1}f_i\colon \R^d\to \R        \right|
              \text{$d\in \mathbb{N}$. Each $f_i\colon \R^d\to \R$ is $L$-gradient Lipschitz and $f(x^0)-\inf_x f(x)=\Delta$.}
    \right\}
\end{align}
Moreover, for an integer $n$, $L>0$, and $\zeta>0$, a problem class $\mathcal{F}_n^{L,\zeta}$ is defined as
\begin{align}
    \mathcal{F}_n^{L,\zeta} = \left\{ 
        f=\frac1n \sum_{i=1}f_i\colon \R^d\to \R        \left|
            \begin{array}{l}
                  \text{$d\in \mathbb{N}$. Each $f_i\colon \R^d\to \R$ is $L$-gradient Lipschitz, }\\
                \text{$\{f_i\}_{i=1}^n$ is Hessian-heterogeneous with $\zeta$, and $f(x^0)-\inf_x f(x)=\Delta$.}  
    \end{array}\right.
    \right\}.
\end{align}
\end{definition}

\citet{carmon2020lower} proved the following lower bound.
\begin{proposition}[\citet{carmon2020lower}]\label{proposition:Carmon}
    Fix $x^0$. 
    For any $L>0$, $\Delta>0$, and $\eps>0$, there exists a function $f\in \mathcal{F}_{1,\Delta}^{L}$ such that any linear-span first-order algorithm requires 
    $\Omega\left(\frac{\Delta L}{\eps^2}\right)$
    stochastic gradient accesses in order to find $\eps$-first-order stationary points.
\end{proposition}
\citet{fang2018spider,li2021page} extended this to the lower bound on the finite-sum optimization problem.
\begin{proposition}[\citet{fang2018spider,li2021page}]\label{proposition:FangandLi}
    Fix $x^0$. 
    For $n>0$, $L>0$, $\Delta>0$, and $\eps>0$, there exists a function $f\in \mathcal{F}_{n,\Delta}^{L}$ such that any linear-span first-order algorithm requires 
    $\Omega\left(n+\frac{\Delta L \sqrt{n}}{\eps^2}\right)$
    stochastic gradient accesses in order to find $\eps$-first-order stationary points.
\end{proposition}
Based on these, we give the lower bound under the additional assumption of $\zeta$-Hessian-heterogeneity.
\setcounter{theorem}{2}
\begin{proposition}
    Assume \cref{assumption:GradLipschitz,assumption:GlobalInfimum,assumption:Heterogeneity}.
For any $L>0$, $\Delta>0$, and $\eps>0$, there exists a function $f\in \mathcal{F}_{n,\Delta}^{L,\zeta}$ such that any linear-span first-order algorithm requires 
    \begin{align}
        \Omega\left(n+\frac{\Delta(\zeta\sqrt{n} +L)}{\eps^2}\right)
    \end{align}
    stochastic gradient accesses in order to find $\eps$-first-order stationary points.
\end{proposition}
\begin{proof}
It is easy to see that the lower bound of \cref{proposition:Carmon} also applies to $\mathcal{F}_{n,\Delta}^L$, by letting $f_1=f_2=\cdots=f_n = f^*$ where $f^*$ is the function that gives the bound of \cref{proposition:Carmon}.
On the other hand, we have $\mathcal{F}_{n,\Delta}^{\frac{\zeta}{2}}\subseteq\mathcal{F}_{n,\Delta}^{L,\zeta}$.
Thus, \cref{proposition:FangandLi} yields that there exists a function $f\in \mathcal{F}_{n,\Delta}^{\frac{\zeta}{2}}\subseteq\mathcal{F}_{n,\Delta}^{L,\zeta}$ that requires $\Omega\left(n+\frac{\Delta\zeta\sqrt{n}}{\eps^2}\right)$ stochastic gradients to find $\eps$-first-order stationary points.
Therefore, by combining these two bounds, we have the desired lower bound.
\end{proof}

\end{document}